\documentclass[pdflatex,sn-nature]{sn-jnl}

\usepackage{graphicx}%
\usepackage{multirow}%
\usepackage{amsmath,amssymb,amsfonts}%
\usepackage{amsthm}%
\usepackage[title]{appendix}%
\usepackage{xcolor}%
\usepackage{textcomp}%
\usepackage{manyfoot}%
\usepackage{booktabs}%
\usepackage{algorithm}%
\usepackage{algorithmicx}%
\usepackage{algpseudocode}%
\usepackage{listings}%

\theoremstyle{thmstyleone}%
\theoremstyle{thmstyletwo}%

\theoremstyle{thmstylethree}%

\usepackage{xurl}

\usepackage[dvipsnames]{xcolor}
\usepackage{placeins} 
\usepackage[final]{changes}
\definechangesauthor[name={FW},color=orange]{FW}
\definechangesauthor[name={USGS},color=ForestGreen]{USGS}
\definechangesauthor[name={CM},color=Purple]{CM}
\setdeletedmarkup{[\textsuperscript{del}\,#1]} 
 \usepackage{array}
 \newcommand{\PreserveBackslash}[1]{\let\temp=\\#1\let\\=\temp}
 \newcolumntype{C}[1]{>{\PreserveBackslash\centering}p{#1}}
 \newcolumntype{R}[1]{>{\PreserveBackslash\raggedleft}p{#1}}
 \newcolumntype{L}[1]{>{\PreserveBackslash\raggedright}p{#1}}

\usepackage{adjustbox} 
\usepackage{rotating} 

\makeatletter
\def\printabstract{\ifx\@abstract\empty\else\clearpage\@abstract\fi\par}
\makeatother

\begin{document}

\title[Large trees of the Sierra Nevada, California: individual delineation and health status over the period 2020--2025 from remotely sensed data]{Large trees of the Sierra Nevada, California: individual delineation and health status over the period 2020--2025 from remotely sensed data} 








\author*[1]{\fnm{Fabien H.} \sur{Wagner}}\email{fwagner@ctrees.org }

\author[2]{\fnm{Dan J.} \sur{Dixon}}

\author[1]{\fnm{Christopher W.} \sur{Woodall}}

\author[1,3,4]{\fnm{Mayumi C. M.} \sur{Hirye}}

\author[5]{\fnm{Felipe} \sur{Saad}}

\author[6]{\fnm{Adrian J.} \sur{Das}}

\author[1]{\fnm{Griffin} \sur{Carter}}

\author[1]{\fnm{Ricardo} \sur{Dalagnol}}

\author[1,3]{\fnm{Lorena} \sur{Alves}}

\author[1]{\fnm{Javiera} \sur{Canales}}

\author[1,3]{\fnm{Cynthia} \sur{Creze}}

\author[1]{\fnm{Stephen C.} \sur{Hagen}}

\author[1]{\fnm{Aijing} \sur{Li}}

\author[1]{\fnm{Zhihua} \sur{Liu}}

\author[1]{\fnm{Christopher} \sur{ Mihiar}}

\author[1,3]{\fnm{Adugna} \sur{Mullissa}}

\author[1]{\fnm{Naomi} \sur{Provost}}

\author[7]{\fnm{Alison L.} \sur{Ritz}}

\author[1]{\fnm{Ignacio} \sur{San Miguel}}

\author[1,3]{\fnm{Bryan} \sur{Shaddy}}

\author[1,3]{\fnm{Le Bienfaiteur Sagang} \sur{Takougoum}}

\author[1]{\fnm{Yan} \sur{Yang}}

\author[1,3]{\fnm{Dafeng} \sur{Zhang}}

\author[1,3,8]{\fnm{Sassan} \sur{Saatchi}}

\affil*[1]{\orgname{CTrees}, \orgaddress{\city{Pasadena}, \state{CA}, \postcode{91105}, \country{USA}}}

\affil[2]{\orgname{Department of Land, Air and Water Resources, University of California, Davis}, \orgaddress{\city{Davis}, \state{CA}, \postcode{95616}, \country{USA}}}


\affil[3]{\orgname{Institute of Environment and Sustainability, University of California, Los Angeles}, \orgaddress{\city{Los Angeles}, \state{CA}, \country{USA}}}

\affil[4]{\orgname{Quapá Lab, Faculty of Architecture and Urbanism, University of São Paulo}, \orgaddress{\postcode{05508080}, \city{São Paulo}, \state{SP}, \country{Brazil}}}

\affil[5]{\orgname{Department of Geography and Environmental Systems, University of Maryland Baltimore County}, \orgaddress{\country{USA}}}

\affil[6]{\orgname{U.S. Geological Survey, Western Ecological Research Center}, \orgaddress{\city{Three Rivers}, \state{CA}, \country{USA}}}

\affil[7]{\orgdiv{Forest Resources and Environmental Conservation}, \orgname{Virginia Polytechnic Institute and State University}, \orgaddress{\city{Blacksburg}, \state{VA}, \postcode{24061}, \country{USA}}}

\affil[8]{\orgname{Jet Propulsion Laboratory, California Institute of Technology}, \orgaddress{\street{4800 Oak Grove Avenue}, \city{Pasadena}, \state{CA}, \postcode{91109}, \country{USA}}}


\abstract{

Large trees store a disproportionate share of forest biomass and take centuries to replace, yet their distribution and health status remain unknown at regional scale. Here, we mapped the large trees (crown area $\geq$ 100 m$^2$) of the Sierra Nevada and characterized their health status from 2020 to 2025. Crowns were delineated in a canopy height model derived from 2020 USDA-NAIP aerial images, using a deep learning model trained on 3.3 million synthetic trees. The delineation, validated with 20,273 giant sequoia crowns and hand-drawn crowns without species identification, reached a median IoU of 0.60 after centroid alignment. In total, 6,515,705 large trees were mapped, and their distribution was mainly associated with temperature and precipitation. Colour changes in the visible and near infrared were detected in the Sentinel-2 time series of each crown and used to determine its health status, validated against field observations of 1,015 giant sequoias. Inside the 2020--2024 wildfire perimeters, 61.4\% of the large trees (946,786) were disturbed or dead, against 2.2\% outside. We estimate that $\sim$650,000 large trees, one in ten in the Sierra Nevada, very likely died within these perimeters, mostly in 2020 and 2021. Individual trees can now be monitored over entire regions.

}

\keywords{Large trees; Individual tree delineation; Deep learning; Sentinel-2 MSI; Wildfire; Tree mortality; Giant sequoia}



\maketitle

\section{Introduction}\label{sec1}

Large trees are few, but they dominate the forests in which they grow \citep{Lutz2018,Piponiot2022}. They store and accumulate a disproportionate share of the aboveground biomass, shape the structure of the canopy, and provide habitats that smaller trees cannot offer \citep{enquist2020,Lutz2018,Stephenson2014}. They are also slow to replace, as a crown that took centuries to grow will not come back within a human lifetime. The Sierra Nevada of California is one of the places on Earth where these trees are the most impressive. It is the home of the giant sequoia (\textit{Sequoiadendron giganteum}), endemic to its western slope \citep{Griffin1976,Stephenson1996}, whose individuals can live for more than 3,200 years. The largest of them, the General Sherman Tree, has a height of 83.8 m, a maximum base diameter of 11.1 m, an average crown spread of 32.5 m \citep{NPS1997} and a volume of 1,486.6 m$^3$ \citep{Stephenson1996, Shive2026}. The giant sequoias are, however, only a small part of the large trees of the Sierra Nevada. Pines such as ponderosa pine (\textit{Pinus ponderosa}) and sugar pine (\textit{Pinus lambertiana}), incense cedar (\textit{Calocedrus decurrens}), firs (\textit{Abies magnifica} and \textit{Abies lowiana}) and Douglas-fir (\textit{Pseudotsuga menziesii}) commonly exceed 30--40 m in height \citep{Griffin1976,Barbour1988}, and contribute to the high aboveground biomass of these forests, which reaches 2,683 Mg ha$^{-1}$ in giant sequoia stands \citep{Winsemius2024,Sillett2019}.

However, these trees are under threat. Over the twentieth century, the forests of California became denser while their large trees became fewer \citep{mcintyre2015}. More recently, the drought of 2012--2015 caused a massive die-off of trees \citep{goulden2019}, and the losses of tree cover driven by fire and climate stress have kept increasing \citep{wang2022,Williams2016}. Then came the fire seasons of 2020 and 2021, with wildfires whose size and severity exceeded historic conditions \citep{Safford2022,Keeley2021,Williams2023,Hung2026}. Fire is a normal component of these ecosystems, and large conifers such as the giant sequoia are adapted to it, but not to fires of this intensity. An estimated 17.6\% of all large giant sequoias have been killed by fire since 1984, and $\sim$80\% of this mortality occurred during the 2020 and 2021 fires alone \citep{Shive2026}. Mortality was lower where prescribed burns had been conducted before the fires \citep{Dixon2026} and higher where giant sequoias were surrounded by dense stands of other trees \citep{Hardlund2026}. Furthermore, mortality does not stop when the fire is out, as trees with a scorched crown can die several years later, or flush again after appearing dead \citep{Hanson2009,Jeronimo2020,Reilly2023}. These estimates exist because the giant sequoias grow in fewer than a hundred groves, which are delimited and well known, and where more than 20,000 of their large individuals have already been located \citep{Dixon2026}. For the millions of other large trees of the Sierra Nevada, it is still unknown how many there are, where they are, and how many of them survived these fires.

Field inventories remain the reference to assess the health of individual trees, but they cannot cover a region of 65,582 km$^2$. Remote sensing has therefore been increasingly used, in two different ways. On the one hand, very high resolution images or airborne LiDAR can resolve individual trees at one date. Using the aerial images of 2020, Cheng et al.~\cite{Cheng2024} mapped 91.4 million dead trees across California and showed that 60\% of this mortality occurred in groups of three trees or fewer, a scattered pattern often missed by conventional monitoring. Using airborne LiDAR, Kane et al.~\cite{Kane2023} counted more than 2.7 million very large trees in three landscapes of the Sierra Nevada. On the other hand, satellite time series follow the changes of the forest through time, but at the scale of the stand rather than of the tree. They showed that wildfire accounted for more than 60\% of the large forest disturbances of California between 2019 and 2021 \citep{Favrichon2025}, and that 21.4\% of the forest area of the Sierra Nevada was disturbed between 2020 and 2021 \citep{Carter2024}. At 3 m spatial resolution, PlanetScope time series revealed the heterogeneous patterns of tree survival and mortality left by large wildfires \citep{Dixon2023}, including in 19 giant sequoia groves burned by the 2020 Castle Fire and the 2021 KNP Complex Fire, which combined the Paradise and Colony fires in Sequoia and Kings Canyon National Parks \citep{Dixon2026}. However, to date, there is no method to follow each large tree of an entire region through time: either the trees are mapped once, or the forest is monitored without knowing its trees.

Here, we aim to bridge this gap in two steps. First, we delineate the crown of every tree in a canopy height model derived from the aerial images of the National Agriculture Imagery Program (NAIP) acquired in 2020 at 0.6 m spatial resolution \citep{Wagner2024}. A canopy height model shows the trees from above, as a LiDAR would, and in this model the crowns appear as rounded shapes that are easier to separate than in the images themselves. The delineation is made with a deep learning model, the U-Net-ID \citep{Wagner2020UnetID}, trained only with synthetic canopy height models in which the position and shape of each simulated tree are known, so that no crown has to be drawn by hand. Second, we keep the trees with a crown area $\geq$ 100 m$^2$, hereafter the large trees. This threshold is the area of one pixel of the Sentinel-2 satellites (10 $\times$ 10 m), which means that each large tree can be followed in the Sentinel-2 images acquired every five days. In the time series of each crown, an abrupt change is searched for with the BFAST algorithm \citep{Verbesselt2010,Verbesselt2010a}, commonly used to detect forest disturbances \citep{Chen2021,Schiller2026}, and the evolution of the colour and greenness of the crown after this change indicates whether the tree remained alive, recovered or died.

The objective of this study is to map all the large trees of the Sierra Nevada Floristic Province and to describe their health status from 2020 to 2025. Specifically, we (i) delineate all the tree crowns of the province and validate this delineation with 20,273 giant sequoia crowns obtained from airborne LiDAR and with crowns without species identification drawn by hand on the aerial images; (ii) describe the distribution of the large trees in relation to elevation, topography and climate; (iii) classify the health status of each large tree from its Sentinel-2 time series, and validate these classes with the status of 1,015 giant sequoias assessed in the field after the 2020 and 2021 fires and with aerial images of 2020 and 2024; and (iv) compare the health status of the large trees inside and outside the perimeters of the 2020--2024 wildfires. The study concerns all the large trees of the Sierra Nevada, whatever their species. The giant sequoias, for which exceptional reference data exist, are used to validate the crown delineation, to build the reference colour range that filters the Sentinel-2 time series, and to validate the health classes.

\FloatBarrier 
\section{Results}\label{sec2}

\FloatBarrier 
\subsection{Validation of the individual tree delineation}

\begin{figure}[!ht]
\centering
\includegraphics[width=0.65\linewidth]{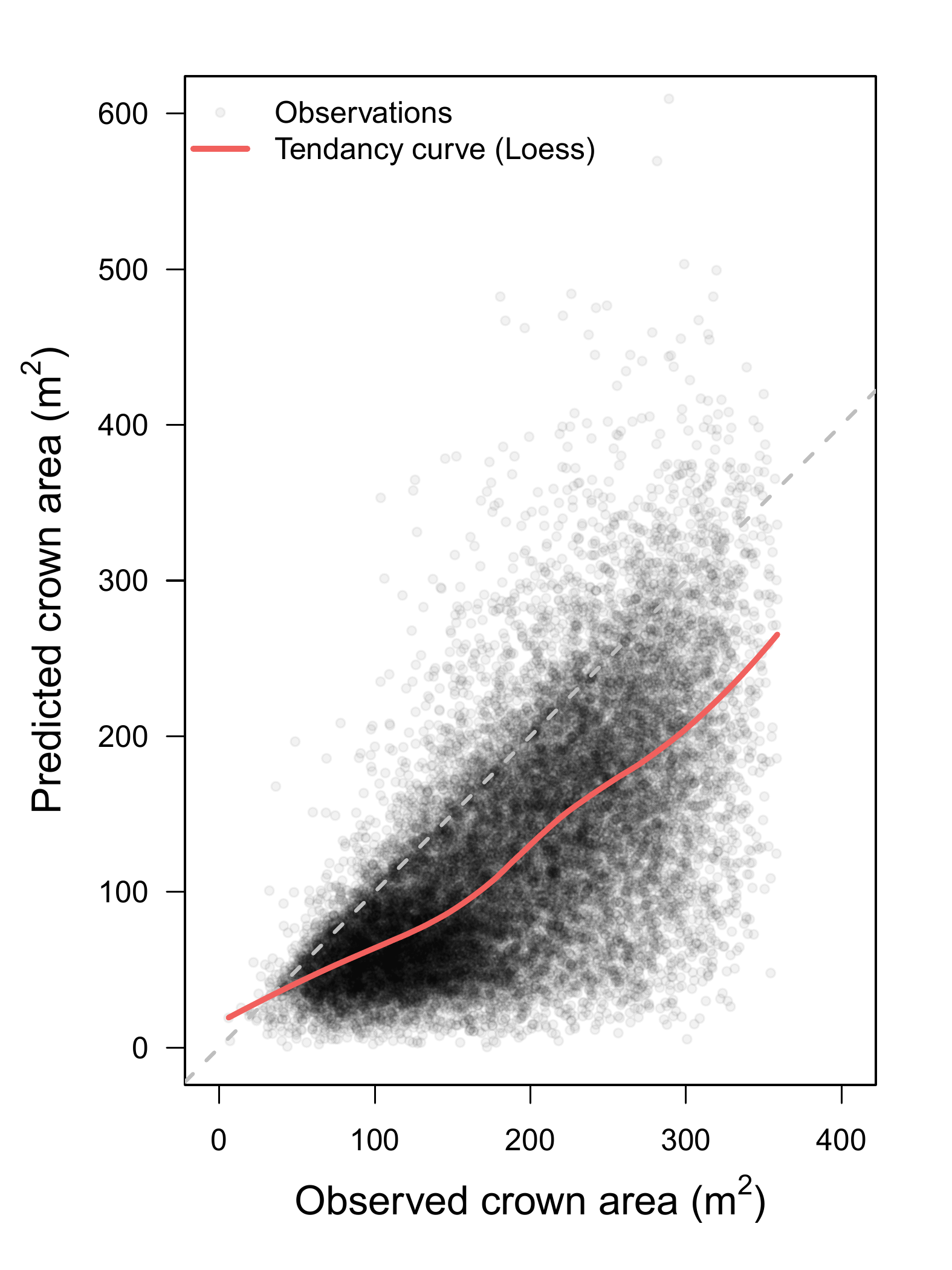}
\caption{Predicted versus observed crown area (m$^2$) for the 20,273 giant sequoias of the reference dataset, whose crowns were delineated from airborne LiDAR \citep{Dixon2026}. The dashed grey line is the 1:1 line and the curve is a LOESS smoothing.}
 
 \label{Figpredobs1} 
 \end{figure}




The crowns delineated by our model were smaller than the reference crowns obtained from airborne LiDAR for the 20,273 giant sequoias (Fig. \ref{Figpredobs1}). The comparison of the crown areas gave an RMSE of 85.5 m$^2$, a Pearson correlation coefficient of 0.66 and a mean bias of $-$60.3 m$^2$ (predicted minus observed area). This underestimation has a consequence for the count of large trees, because a crown that is slightly above 100 m$^2$ in the reference can fall below this threshold in our dataset. Of the 17,228 reference crowns $\geq$ 100 m$^2$, 40.8\% were predicted with an area $<$ 100 m$^2$. This loss was concentrated near the threshold, that is, 62.0\% of the reference crowns of 100--200 m$^2$ were predicted below 100 m$^2$, but only 18.5\% of those of 200--300 m$^2$ and 10.2\% of those above 300 m$^2$ (109 of 1,068 crowns).

When a reference crown above 300 m$^2$ was predicted below 100 m$^2$, the crown was generally split by the model into several smaller polygons. Conversely, the predicted crowns larger than 400 m$^2$ typically resulted from the merging of two neighbouring crowns, but rarely more than two.

\begin{figure}[!ht]
\centering
\includegraphics[width=0.48\linewidth]{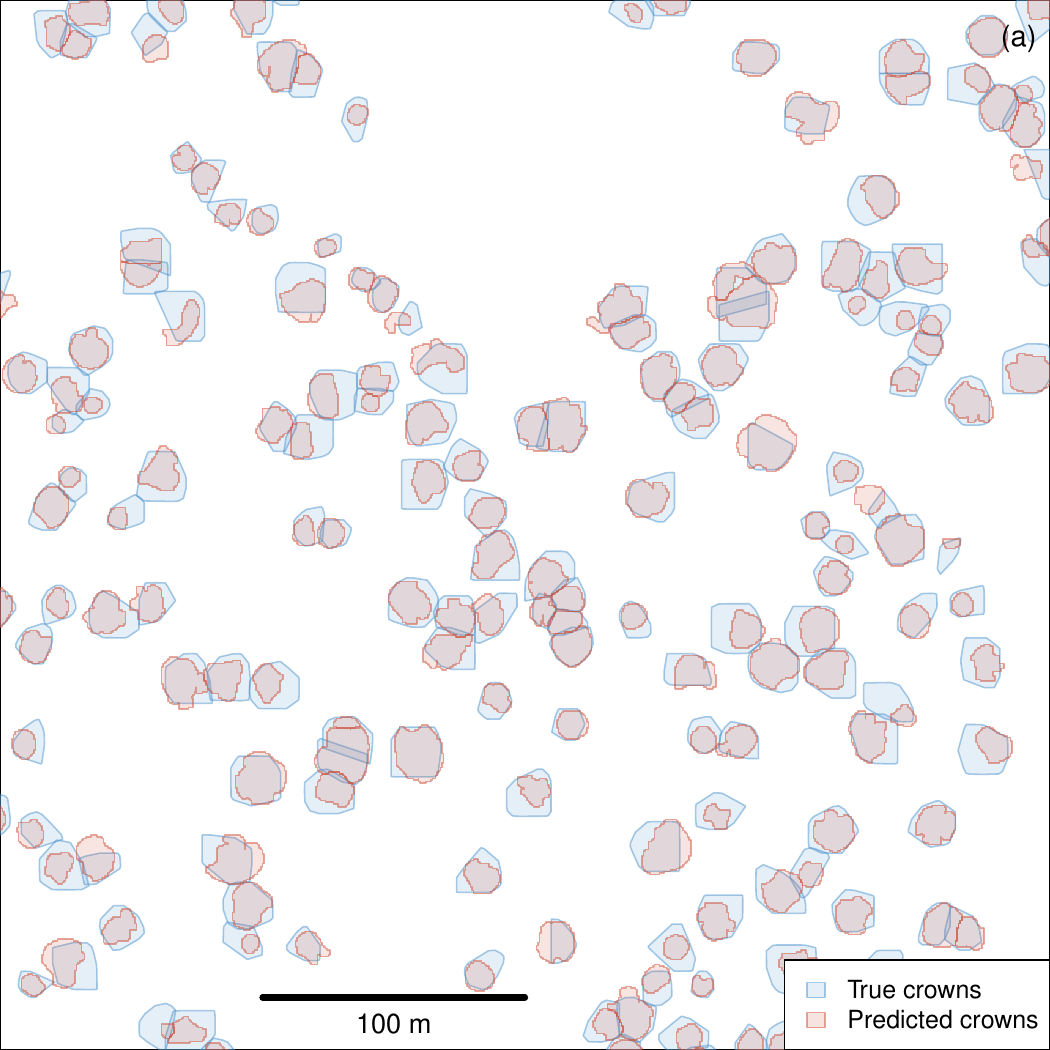}
\includegraphics[width=0.48\linewidth]{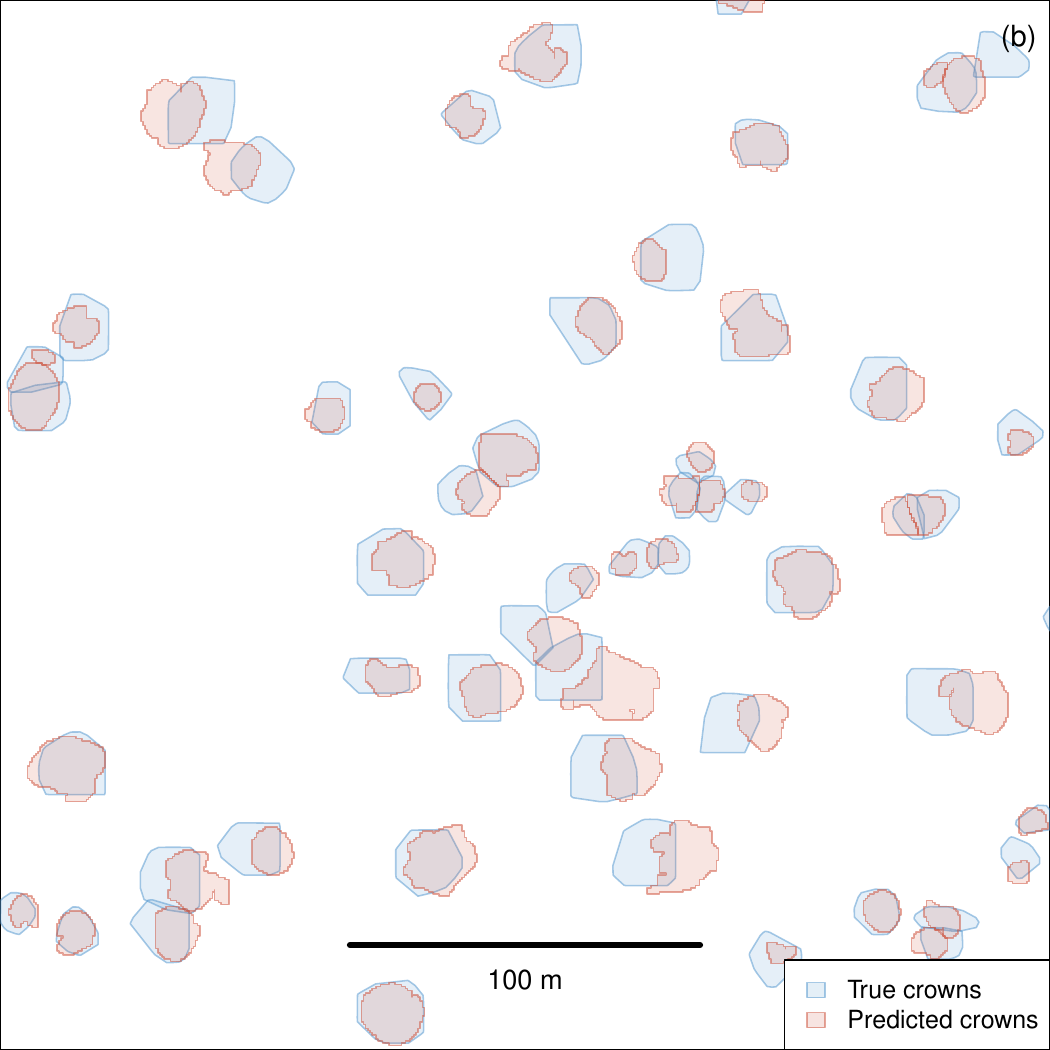}
\caption{Examples of giant sequoia crowns from the LiDAR-based reference dataset (blue) and delineated by the U-Net-ID model (red).}
 \label{Figpredobs2} 
 \end{figure}

Examples of the delineation of giant sequoias are presented in Fig. \ref{Figpredobs2}. Despite the systematic bias in size and a slight displacement, the crowns delineated by the U-Net-ID model agreed well with the LiDAR-based crowns. Note that the reference polygon can be larger than the real crown (Fig. \ref{Figpredobs2}a), and that the predicted crown can be slightly displaced from the reference crown, particularly on slopes (Fig. \ref{Figpredobs2}b).

\begin{figure}[!ht]
\centering
\includegraphics[width=0.48\linewidth]{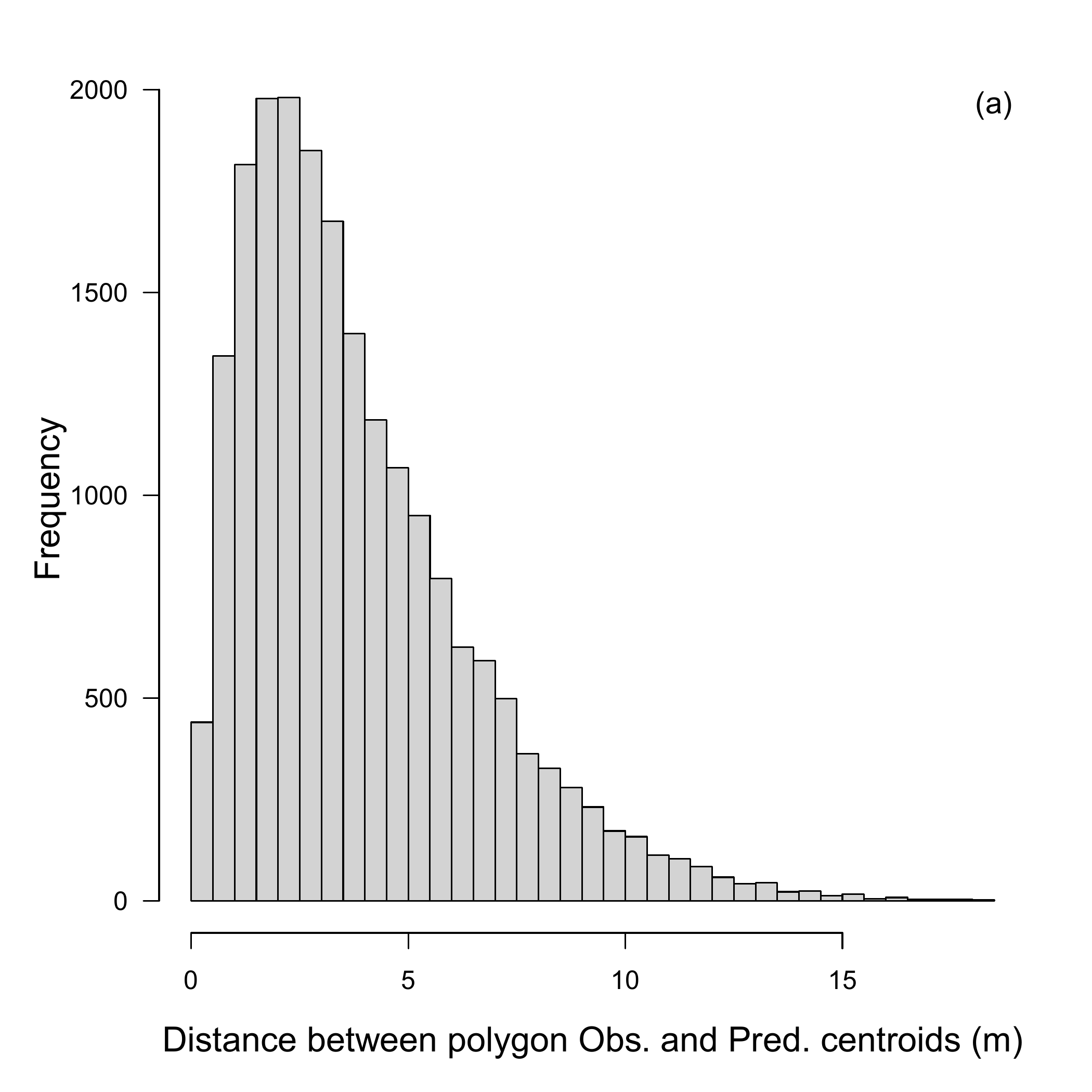}
\includegraphics[width=0.48\linewidth]{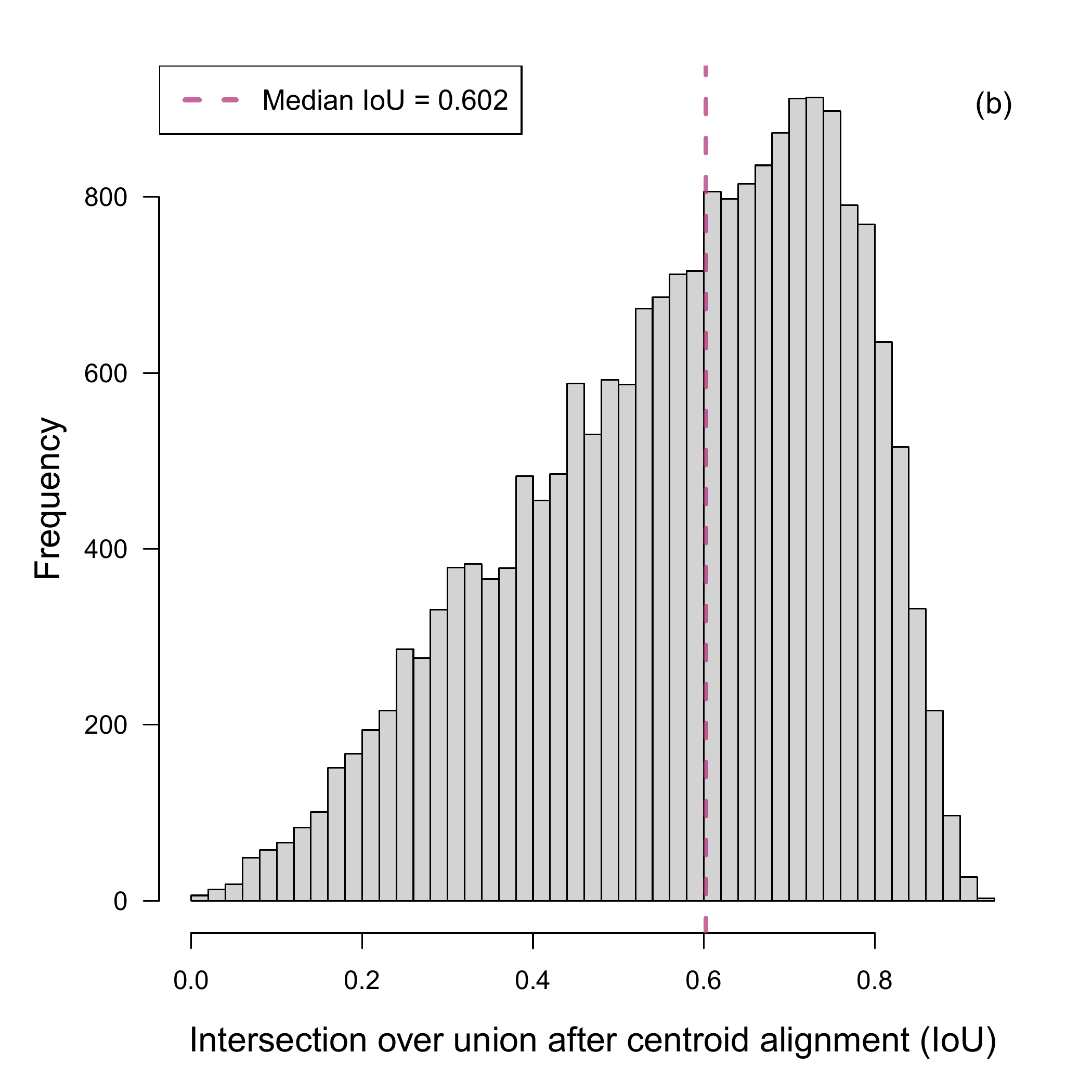}
 \caption{Distance between the centroids of the predicted and reference crowns for the 20,272 giant sequoias with a predicted crown (a), and distribution of their Intersection over Union (IoU) after centroid alignment (b).}
 \label{distance_centroid_and_iou} 
 \end{figure}

%

The median distance between the centroids of the reference and predicted crowns was 3.2 m (2.5--97.5 percentiles: 0.54--10.64 m; Fig. \ref{distance_centroid_and_iou}a). This displacement strongly affects the Intersection over Union (IoU). The median IoU was 0.399 without alignment (mean of 0.405), and 0.602 when the predicted crown was translated to the centroid of the reference crown (mean of 0.572; Fig. \ref{distance_centroid_and_iou}b). In other words, the size and shape of the crowns were better reproduced than their exact position.

The agreement was similar for the crowns without species identification drawn by hand on the NAIP images (Table \ref{tab_valid_all_species}). For the 331 reference crowns $\geq$ 100 m$^2$, the median IoU was 0.335 without alignment and 0.605 after alignment, very close to the value obtained for the giant sequoias (0.602). The median distance between centroids was 4.1 m and the crown area was underestimated by 45.7 m$^2$ on average, so that 56.1\% of these crowns were predicted with an area $<$ 100 m$^2$. Note that these crowns were drawn on images that are not always acquired at nadir, that is, the trees can appear distorted, which increases the size of the drawn crowns. When all the reference crowns were considered, whatever their size, the crown area was not biased (+4.7 m$^2$) and the median IoU after alignment was 0.530. For the large crowns (279 predicted and 331 reference crowns), the precision, recall and F1-score at IoU $\geq$ 0.5 were low without alignment (0.12, 0.10 and 0.11) and reached 0.48, 0.41 and 0.44 after alignment, which confirms that the main source of disagreement is the displacement of the crowns and not their shape.

\begin{table}[ht]
\caption{Agreement between predicted and reference crowns for the giant sequoias (lidar reference crowns) and for the crowns without species identification drawn by hand on the NAIP 2020 images of 11 tiles, with the same one-to-one matching. The IoU after alignment is computed after translating the predicted crown to the centroid of the reference crown. Precision and recall were computed only for the hand-drawn crowns $\geq$ 100 m$^2$.}
\centering
\footnotesize
\begin{tabular}{lrrr}
\hline
 & Giant sequoias & \multicolumn{2}{c}{Hand-drawn, no species identification} \\
 & (lidar) & crowns $\geq$ 100 m$^2$ & all crowns \\
\hline
Reference crowns & 20,273 & 331 & 20,393 \\
Matched one-to-one & 20,272 & 314 & 13,895 \\
Median IoU & 0.399 & 0.335 & 0.252 \\
Median IoU after alignment & 0.602 & 0.605 & 0.530 \\
Median centroid distance (m) & 3.2 & 4.1 & 1.8 \\
Bias of crown area (m$^2$) & $-$60.3 & $-$45.7 & $+$4.7 \\
RMSE of crown area (m$^2$) & 85.5 & 76.9 & 22.4 \\
Pearson r of crown area & 0.66 & 0.42 & 0.66 \\
Precision / recall at IoU $\geq$ 0.5 & -- & 0.12 / 0.10 & -- \\
Precision / recall at IoU $\geq$ 0.5 after alignment & -- & 0.48 / 0.41 & -- \\
\hline
\end{tabular}
\label{tab_valid_all_species}
\end{table}






\FloatBarrier 
\subsection{Distribution of large trees across the Sierra Nevada}

\begin{figure}[!ht]
\centering
\includegraphics[width=0.78\linewidth]{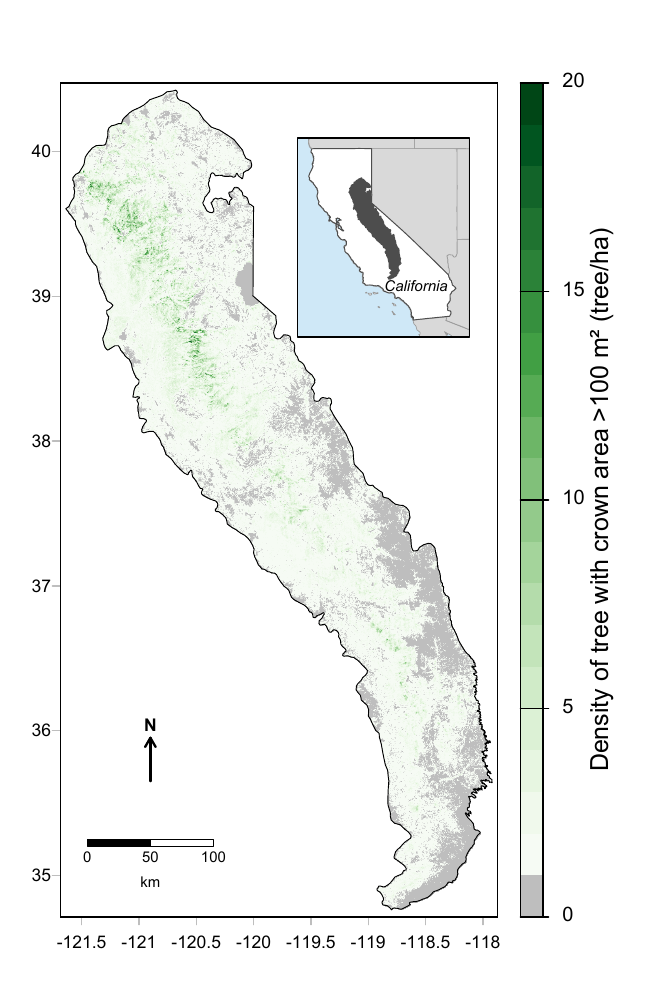}
\caption{Distribution of the 6,515,705 tree crowns $\geq$ 100 m$^2$ delineated with the U-Net-ID model in the Sierra Nevada. The inset locates the Sierra Nevada Floristic Province (dark grey) in California.} 
 \label{tree_distrib} 
 \end{figure}




We identified 6,515,705 large trees in the Sierra Nevada Floristic Province, distributed across 78.7\% of its area (Fig. \ref{tree_distrib}). Their density was low in most areas ($\sim$98\% with 0--5 trees ha$^{-1}$) and reached $\sim$20 trees ha$^{-1}$ in localized clusters. The highest densities were found in the mid-elevation forests of the western slope, while large trees were sparse in the south and east of the province and almost absent at the highest elevations, mainly along the eastern border, and at the lowest elevations, mainly in the west. Most of the giant sequoia groves are located in the southern Sierra Nevada, and their mean density of large trees ranged from 0.08 to 10.60 trees ha$^{-1}$.


 \begin{figure}[ht]
 \centering
 \includegraphics[width=0.95\linewidth]{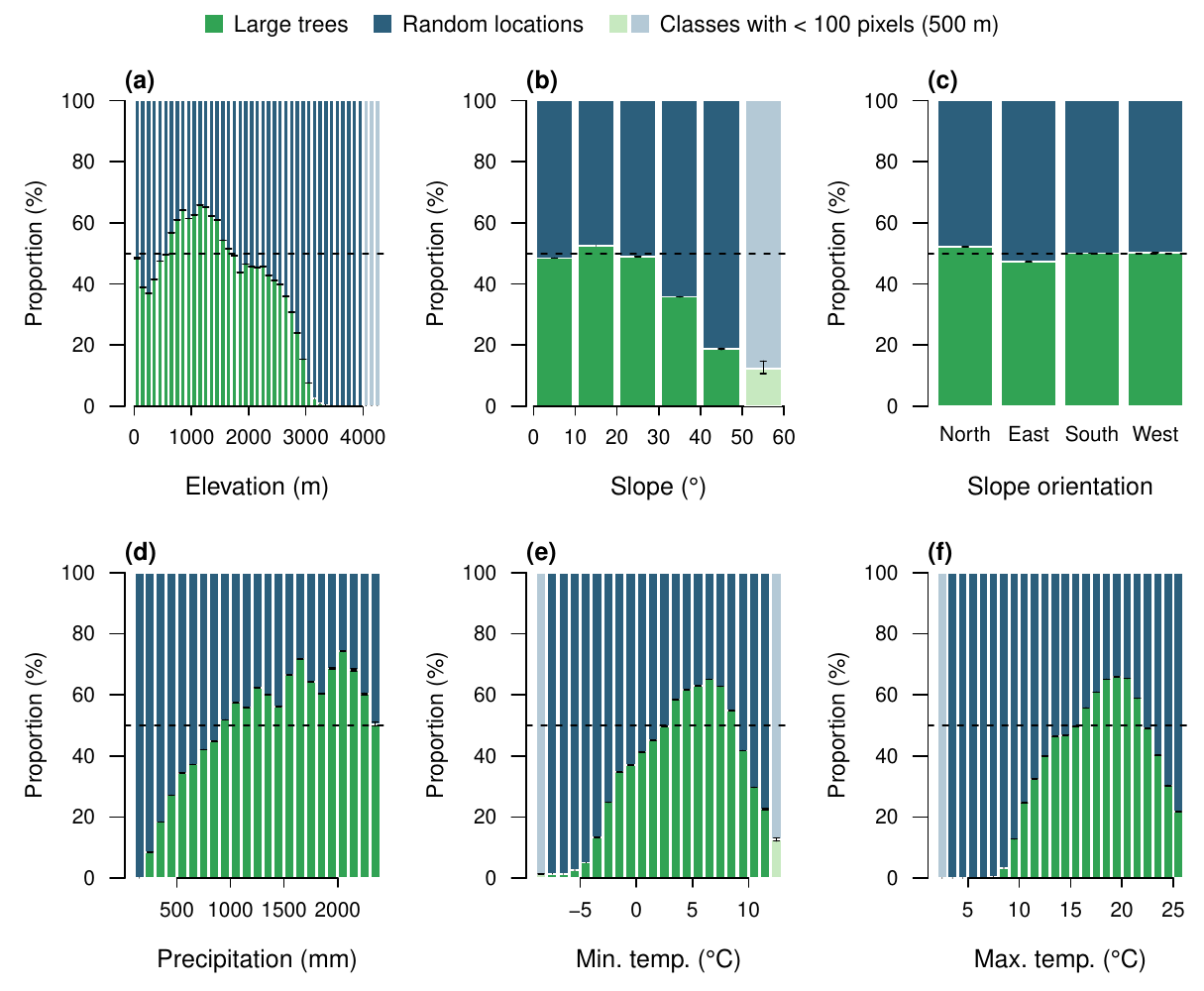}
  \caption{Proportion of large trees in relation to the same number of random locations over the Sierra Nevada Floristic Province, for classes of (a) elevation (m), (b) slope ($^\circ$), (c) slope orientation (North, East, South, West), (d) mean annual precipitation (mm~yr$^{-1}$), (e) annual mean minimum temperature ($^\circ$C), and (f) annual mean maximum temperature ($^\circ$C). The x-axis gives the class limits (classes of 100 m for elevation, 10$^\circ$ for slope, 100 mm for precipitation and 1 $^\circ$C for temperatures). The random locations are bootstrapped 100 times to create the 95\% confidence interval (black error bars). A proportion above 50\% (black dashed line) indicates an over-representation of large trees in the class, and a proportion below 50\% an under-representation. Classes with less than 100 pixels of 500 m are shown in light colours.}
  \label{EnvVarRes}
  \end{figure}


The large trees were not distributed at random with respect to the environment (Fig. \ref{EnvVarRes}). Elevation showed the clearest pattern. Large trees were over-represented between 600 and 1,700 m, with a maximum between 1,100 and 1,200 m, where they were $\sim$1.9 times more frequent than random locations (proportion of 65.9\%; Fig. \ref{EnvVarRes}a). Above 1,700 m, their proportion decreased gradually, and they were almost absent above 3,000 m. The maximum temperature showed a very similar pattern, as the two variables are correlated, with an over-representation of large trees between 16 and 22 $^\circ$C and a maximum between 19 and 20 $^\circ$C (65.8\%; Fig. \ref{EnvVarRes}f). For the minimum temperature, large trees were over-represented between 3 and 9 $^\circ$C (Fig. \ref{EnvVarRes}e). Large trees were also associated with wet areas, and were over-represented in all the classes of precipitation between 900 and 2,300 mm yr$^{-1}$, with a maximum of 74.3\% between 2,000 and 2,100 mm yr$^{-1}$ (Fig. \ref{EnvVarRes}d). Topography had less influence. Large trees were slightly over-represented on slopes between 10 and 20$^\circ$ (52.5\%) and under-represented on steep slopes (35.8\% between 30 and 40$^\circ$ and 18.7\% between 40 and 50$^\circ$; Fig. \ref{EnvVarRes}b). The differences between orientations were small, with slightly more large trees on north-facing slopes (52.2\%) and slightly fewer on east-facing slopes (47.3\%; Fig. \ref{EnvVarRes}c).

%

\FloatBarrier 
\subsection{Validation of the health classes with field observations}

Our classes of health status agreed well with the status of the 1,015 giant sequoias observed in the field (Table \ref{tab_valid_classes}). The trees of the alive (1) and alive with breakpoint (2) classes were alive on the ground in 96.7\% and 95.3\% of the cases, respectively. At the other end, 72.0\% of the trees of the breakpoint with $<$80\% NDVI recovery (4) class and 68.9\% of the trees of the dead (6) class were dead on the ground. The permanent alteration (5) class was mostly made up of trees alive on the ground (88.9\%), and the already dead or outside the reference colour range (0) class mostly of dead trees (75.0\%).

\begin{table}[ht]
\centering
\begin{tabular}{lcc}
  \hline
 & \multicolumn{2}{c}{\textbf{Ground based observations}} \\
 & Alive & Dead \\ 
  \hline
    \textbf{Satellite based observation} &  &  \\ 
  alive (1) & 202 (96.7\%, 35.7\%) & 7 (3.3\%, 1.6\%) \\ 
  alive with breakpoint (2) & 142 (95.3\%, 25.1\%) & 7 (4.7\%, 1.6\%) \\ 
  breakpoint with $\geq$80\% NDVI recovery (3) & 57 (62.0\%, 10.1\%) & 35 (38.0\%, 7.8\%) \\ 
  \quad under a closed canopy (3a) & 28 (80.0\%, 4.9\%) & 7 (20.0\%, 1.6\%) \\ 
  \quad with canopy opening (3b) & 29 (50.9\%, 5.1\%) & 28 (49.1\%, 6.2\%) \\ 
  breakpoint with $<$80\% NDVI recovery (4) & 142 (28.0\%, 25.1\%) & 365 (72.0\%, 81.3\%) \\ 
  dead (6) & 14 (31.1\%, 2.5\%) & 31 (68.9\%, 6.9\%) \\ 
  permanent alteration (5) & 8 (88.9\%, 1.4\%) & 1 (11.1\%, 0.2\%) \\ 
  already dead or outside the reference colour range (0) & 1 (25.0\%, 0.2\%) & 3 (75.0\%, 0.7\%) \\ 
  \hline
  \textbf{Total} & 566 (55.8\%, 100\%) & 449 (44.2\%, 100\%) \\ 
   \hline
\end{tabular}
\caption{Status observed in the field for the 1,015 giant sequoias that intersect our large crowns, by health class. Values are the number of trees, followed by the row percentage (share of the trees of the class that are alive or dead on the ground) and the column percentage (share of the alive or dead trees on the ground that fall in the class). The subclasses 3a (closed canopy) and 3b (canopy opening) split class 3 (Section \ref{classes_def}) and are not counted in the total.}

\label{tab_valid_classes}
\end{table}

The breakpoint with $\geq$80\% NDVI recovery (3) class was more mixed, with 62.0\% of its trees alive and 38.0\% dead on the ground. The change of summer brightness separated these two cases: 49.1\% (28 of 57) of the crowns with canopy opening (3b) were dead on the ground, against 20.0\% (7 of 35) of the crowns under a closed canopy (3a). The same signal was found in the other disturbed classes. Above the brightness threshold, 78.3\% of the trees of the breakpoint with $<$80\% NDVI recovery (4) class and 77.1\% of the trees of the dead (6) class were dead on the ground, against 39.0\% and 40.0\% below it (n = 425 and 82 for class 4, and 35 and 10 for class 6). The visual interpretation of the NAIP 2020 and 2024 images of class 3 crowns burned in 2020 and 2021, which include all species, gave higher proportions of dead trees than the field data: 19 of the 20 crowns with a change of summer brightness between 0.025 and 0.030 were dead, and 39\% (95\% CI: 29--51\%, n = 76) of the crowns below 0.025.

The changes of NDVI and reflectance were also consistent with the status observed in the field. For the crowns with a significant breakpoint, the mean NDVI decreased more after the breakpoint for the trees dead on the ground than for the trees alive on the ground ($-$0.413 $\pm$ 0.096 against $-$0.182 $\pm$ 0.114, mean $\pm$ SD, n = 439 and 363), and the visible reflectance increased more (red: $+$0.039 $\pm$ 0.010 against $+$0.023 $\pm$ 0.011; green: $+$0.015 $\pm$ 0.008 against $+$0.010 $\pm$ 0.006; blue: $+$0.020 $\pm$ 0.007 against $+$0.012 $\pm$ 0.006). This difference was found in each class (Table \ref{tab_bands_field_status}). Examples of three large crowns of each class, as seen in the NAIP images of 2020 and 2024, are shown in Fig. \ref{fig_naip_compact}, and 18 other crowns per class are available in Appendix \ref{secA_naip} (Figs. \ref{fig_naip_class1} to \ref{fig_naip_class0}). In these images, the crowns of the alive and alive with breakpoint classes remained green in 2024 (Fig. \ref{fig_naip_compact}a-b). In the breakpoint with $\geq$80\% NDVI recovery class, the crowns under a closed canopy (3a) were still green (Fig. \ref{fig_naip_compact}c), while the crowns with canopy opening (3b) were lost and green vegetation covered the ground around them (Fig. \ref{fig_naip_compact}d). The crowns of the breakpoint with $<$80\% NDVI recovery and dead classes were replaced by grey dead crowns or bare ground (Fig. \ref{fig_naip_compact}e-f). In the permanent alteration class, the trees were still present in 2024, but their crown, their direct neighbours or the ground around them had changed (Fig. \ref{fig_naip_compact}g). Finally, the three crowns of the already dead or outside the reference colour range class show two trees that were already dead or dying in 2020 and one tree with a very different crown that could be an oak (Fig. \ref{fig_naip_compact}h).

\begin{figure}[ht]
\centering
\includegraphics[width=0.95\linewidth]{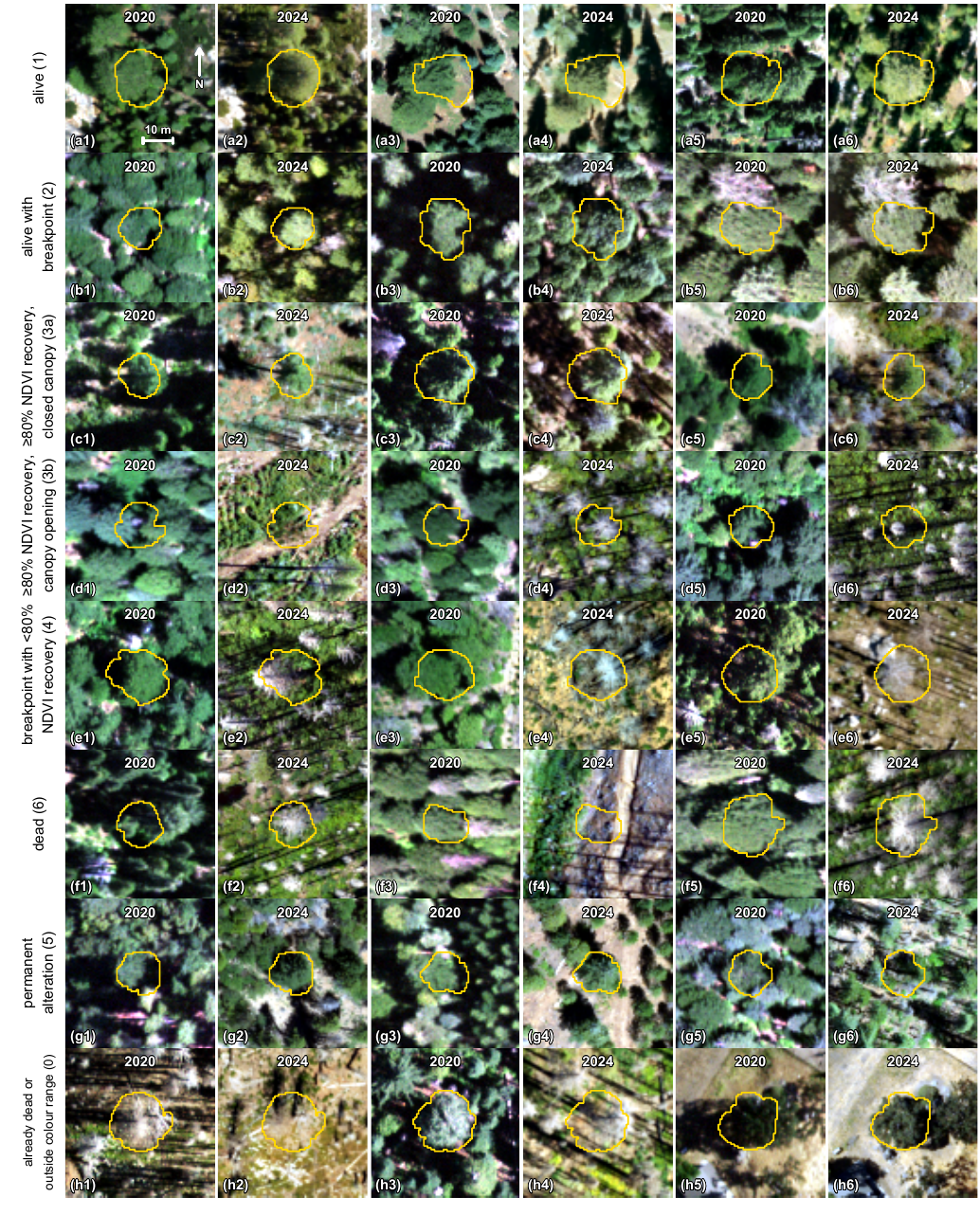}
\caption{NAIP images of three large crowns of each health class, before (2020) and after the fires (2024). Each image covers 50 $\times$ 50 m around the crown and the crown polygon is shown in gold. Rows: (a) alive (1), (b) alive with breakpoint (2), (c) breakpoint with $\geq$80\% NDVI recovery under a closed canopy (3a), (d) breakpoint with $\geq$80\% NDVI recovery and canopy opening (3b), (e) breakpoint with $<$80\% NDVI recovery (4), (f) dead (6), (g) permanent alteration (5), and (h) already dead or outside the reference colour range (0). Note that the crowns were delineated in the canopy height model, a nadir representation, and do not necessarily align with the images, where trees can be displaced or distorted by the view angle.}
\label{fig_naip_compact}
\end{figure}

\FloatBarrier 
\subsection{Health status of the large trees and wildfires}
\label{section_main_results_fire}

\begin{figure}[!ht]
\centering
\includegraphics[width=0.49\linewidth, trim={1.15cm 0.49cm 1.0cm 1.0cm},clip]{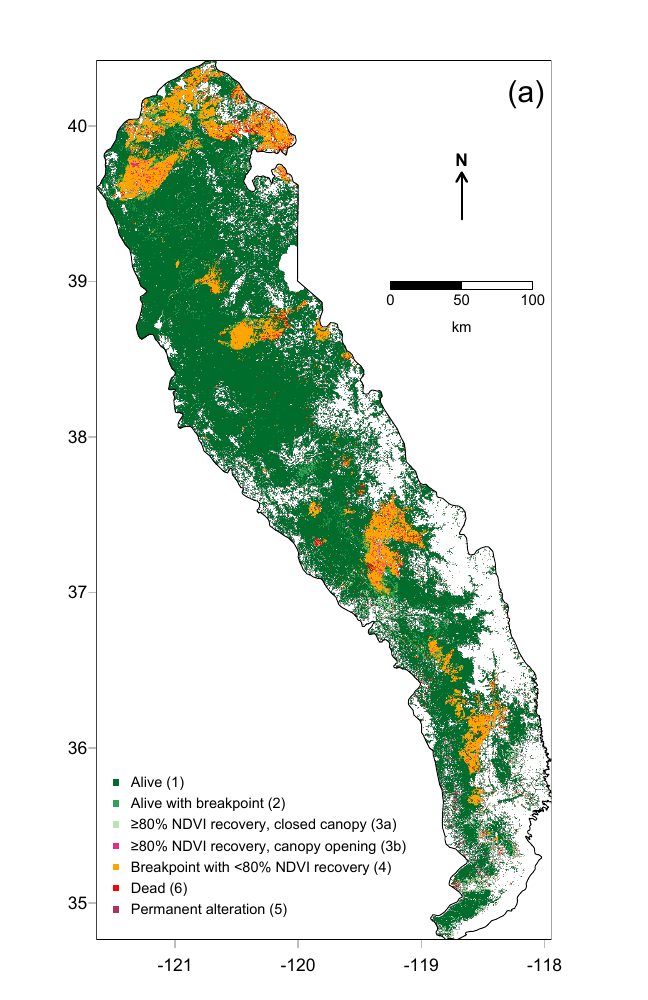}
\includegraphics[width=0.49\linewidth, trim={1.15cm 0.49cm 1.0cm 1.0cm},clip]{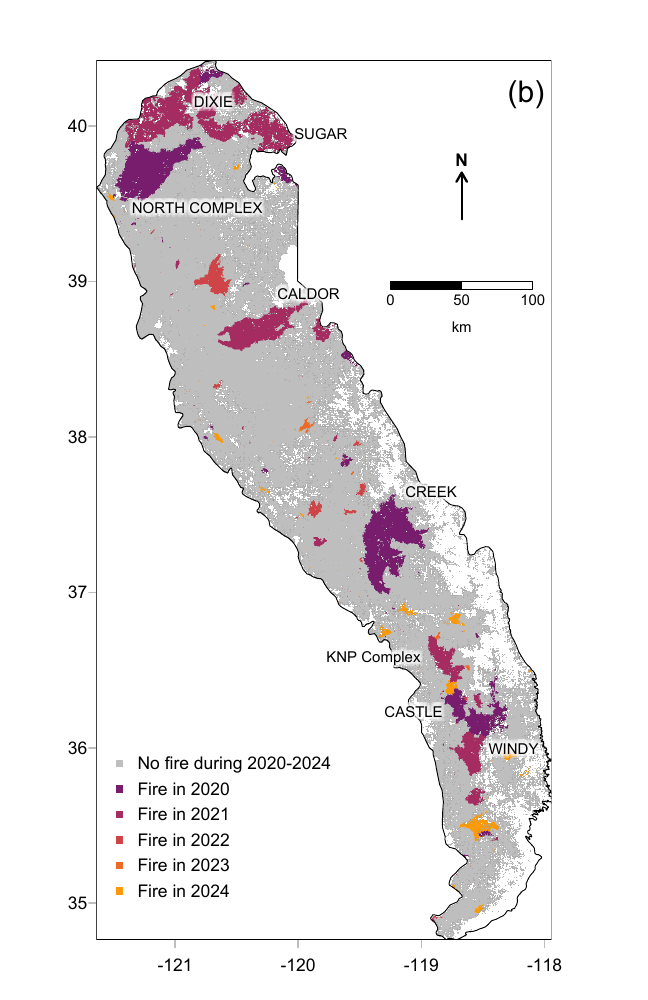}
 \caption{Health classes of the large trees of the Sierra Nevada (a) and year of fire during the 2020--2024 period (b), at 500 m spatial resolution. In (a), each pixel shows its most represented class. White pixels contain no crown $\geq$ 100 m$^2$. The black line is the limit of the Sierra Nevada Floristic Province. The names of the eight largest fires, which all occurred in 2020 and 2021, are displayed.}
 \label{tree_fire_distrib} 
 \end{figure}

The health classes of the large trees showed clear spatial patterns (Fig. \ref{tree_fire_distrib}a). The large trees classified as alive formed a nearly continuous distribution along the mountain range, while the disturbed classes were concentrated in a few large clusters. These clusters matched the perimeters of the 2020--2024 wildfires (Fig. \ref{tree_fire_distrib}b), in the northern Sierra Nevada and in several large fires of the central and southern parts of the range. Outside the fire perimeters, large trees were mostly classified as alive, and a substantial number of large trees also remained classified as alive inside them.


Over the period 2020--2024, the area within the fire perimeters was 9,978 km$^2$, of which 86.8\% burned in 2020 (39.9\%) and 2021 (46.9\%). The eight largest fires occurred during these two years (Fig. \ref{tree_fire_distrib}b), and 87.9\% of the large trees located within fire perimeters (1,355,019 of 1,541,024) were within the perimeters of the 2020 and 2021 fires.

\newsavebox{\tabfirebox}
\begin{lrbox}{\tabfirebox}
\begin{tabular}{l ll|r}
  \hline
Class & Burned (N, \%row, \%col) & Unburned (N, \%row, \%col) & Total (N) \\ 
  \hline
  alive (1) & 271,755 (7.23\%, 17.63\%) & 3,486,934 (92.77\%, 70.09\%) & 3,758,689 \\ 
  alive with breakpoint (2) & 152,017 (18.71\%, 9.86\%) & 660,432 (81.29\%, 13.28\%) & 812,449 \\ 
  breakpoint with $\geq$80\% NDVI recovery (3) & 152,543 (59.61\%, 9.9\%) & 103,346 (40.39\%, 2.08\%) & 255,889 \\ 
  \quad under a closed canopy (3a) & 45,524 (34.48\%, 2.95\%) & 86,517 (65.52\%, 1.74\%) & 132,041 \\ 
  \quad with canopy opening (3b) & 107,019 (86.41\%, 6.94\%) & 16,829 (13.59\%, 0.34\%) & 123,848 \\ 
  breakpoint with $<$80\% NDVI recovery (4) & 692,992 (92.38\%, 44.97\%) & 57,156 (7.62\%, 1.15\%) & 750,148 \\ 
  dead (6) & 146,775 (79.61\%, 9.52\%) & 37,591 (20.39\%, 0.76\%) & 184,366 \\ 
  permanent alteration (5) & 20,576 (37\%, 1.34\%) & 35,029 (63\%, 0.7\%) & 55,605 \\ 
  already dead or outside the reference colour range (0) & 103,376 (14.96\%, 6.71\%) & 587,426 (85.04\%, 11.81\%) & 690,802 \\ 
  not enough data (255) & 990 (12.76\%, 0.06\%) & 6,767 (87.24\%, 0.14\%) & 7,757 \\ 
    \hline
  TOTAL & 1,541,024 & 4,974,681 & 6,515,705 \\ 
   \hline
\end{tabular}
\end{lrbox}
\begin{tableorg}[ht]
\caption{Health classes of the large crowns of the Sierra Nevada inside and outside the 2020--2024 fire perimeters. N is the number of large crowns; \%row is the share of the crowns of the class located inside (burned) or outside (unburned) the fire perimeters, and \%col is the share of the crowns inside or outside the fire perimeters that belong to the class. The subclasses 3a (closed canopy) and 3b (canopy opening) split class 3 and are not counted in the total.}
\centering
\resizebox{\textwidth}{!}{\usebox{\tabfirebox}}
\label{tab_fire}
\end{tableorg}


Of the 6.52 million large trees of the Sierra Nevada, 1.54 million (23.6\%) were located within the fire perimeters (Table \ref{tab_fire}). To assess the role of these fires, we compared the proportion of disturbed large trees inside and outside the perimeters, a tree being considered as disturbed when it was classified as breakpoint with $<$80\% NDVI recovery (4), dead (6) or breakpoint with $\geq$80\% NDVI recovery and canopy opening (3b). Within the fire perimeters, 61.4\% of the large trees were disturbed (946,786 of 1,541,024), against 2.2\% outside (111,576 of 4,974,681), that is, 27.4 times more ($\chi^2$ = 3,030,421, df = 1, p $<$ 0.001).

This contrast did not depend on the size of the crowns or on the fire considered. Inside the fire perimeters, 61.6\%, 58.0\% and 47.1\% of the crowns of 100--200, 200--400 and $\geq$ 400 m$^2$ were disturbed, against 2.3\%, 2.0\% and 3.3\% outside (Table \ref{tab_crown_size}). For each of the eight largest fires, the proportion of disturbed crowns dropped from 46.5--74.9\% inside the perimeter to 1.5--4.3\% in a band of 2 to 5 km around it (Table \ref{tab_largest_fires}). The proportion of disturbed crowns was 65.5\%, 64.2\% and 52.0\% for the fires of 2020, 2021 and 2022, but only 9.1\% and 2.2\% for the fires of 2023 and 2024 (Table \ref{tab_fire_year}). These last two years burned fewer large trees (10,215 and 49,950) and have less than two years of Sentinel-2 observations after the fire. Finally, for the crowns with a significant breakpoint, the changes after the breakpoint were larger inside than outside the fire perimeters, with a decrease of the mean NDVI of 0.32 $\pm$ 0.15 against 0.04 $\pm$ 0.08, and an increase of the red reflectance of 0.032 $\pm$ 0.015 against 0.008 $\pm$ 0.012 (mean $\pm$ SD; green: +0.015 $\pm$ 0.010 against +0.006 $\pm$ 0.009; blue: +0.016 $\pm$ 0.009 against +0.004 $\pm$ 0.008).





The detail of the classes gives the same picture (Table \ref{tab_fire}). Most large trees of the Sierra Nevada were classified as alive (3.76 million, 57.7\%) or alive with breakpoint (0.81 million, 12.5\%), and the vast majority of them were outside the fire perimeters (92.8\% and 81.3\%, respectively). These two classes nevertheless remained common inside the fire perimeters, where they represented 27.5\% of the large trees (271,755 and 152,017 crowns). In contrast, 92.4\% of the 750,148 crowns of the breakpoint with $<$80\% NDVI recovery class and 79.6\% of the crowns of the dead class were inside the fire perimeters (692,992 and 146,775 crowns, respectively). The breakpoint with $\geq$80\% NDVI recovery class was intermediate, with 59.6\% of its crowns inside the fire perimeters. However, when this class was separated by canopy opening, 86.4\% of the crowns with canopy opening (3b) were inside the fire perimeters (107,019 of 123,848), against only 34.5\% of the crowns under a closed canopy (3a). Most class 3 crowns within the fire perimeters showed canopy opening (70.2\%), against 16.3\% outside (Table \ref{tab_fire}). By year of fire, this share was 78.8\%, 58.3\% and 58.1\% for the fires of 2020, 2021 and 2022, against 8.3\% and 5.7\% for the fires of 2023 and 2024, which had at most one summer of Sentinel-2 observations after the fire. Overall, the most frequent class inside the fire perimeters was breakpoint with $<$80\% NDVI recovery (45.0\% of the crowns), followed by alive (17.6\%), breakpoint with $\geq$80\% NDVI recovery (9.9\%), alive with breakpoint (9.9\%) and dead (9.5\%), whereas outside the perimeters the alive and alive with breakpoint classes accounted for 70.1\% and 13.3\% of the crowns.



The number of dead trees was then estimated from the field validation, which showed that not all the crowns of the disturbed classes were dead on the ground (Table \ref{tab_valid_classes}). We applied the shares of dead trees measured in the field for the dead (68.9\%), breakpoint with $<$80\% NDVI recovery (72.0\%) and breakpoint with $\geq$80\% NDVI recovery with canopy opening (3b, 49.1\%) classes to the crowns located within the fire perimeters. The crowns under a closed canopy (3a) were considered alive. This gives approximately 650,000 large trees very likely dead (95\% CI: 615,000--689,000), that is, 42.3\% (39.9--44.7\%) of the large trees within the fire perimeters and $\sim$10.0\% (9.4--10.6\%) of all the large trees of the Sierra Nevada. With the same calculation, 1.5\% (1.4--1.6\%) of the large trees located outside the fire perimeters are very likely dead, a value that should be taken as an upper bound.

\FloatBarrier 
\subsection{Large trees of the giant sequoia groves}

\newsavebox{\tabgrovebox}
\begin{lrbox}{\tabgrovebox}
\begin{tabular}{l ll|r}
  \hline
Class & Burned (N, \%row, \%col) & Unburned (N, \%row, \%col) & Total (N) \\ 
  \hline
  alive (1) & 18,346 (50.92\%, 41.57\%) & 17,680 (49.08\%, 76.47\%) & 36,026 \\ 
  alive with breakpoint (2) & 7,096 (65.67\%, 16.08\%) & 3,710 (34.33\%, 16.05\%) & 10,806 \\ 
  breakpoint with $\geq$80\% NDVI recovery (3) & 2,682 (87.05\%, 6.08\%) & 399 (12.95\%, 1.73\%) & 3,081 \\ 
  \quad under a closed canopy (3a) & 916 (77.43\%, 2.08\%) & 267 (22.57\%, 1.15\%) & 1,183 \\ 
  \quad with canopy opening (3b) & 1,766 (93.05\%, 4.00\%) & 132 (6.95\%, 0.57\%) & 1,898 \\ 
  breakpoint with $<$80\% NDVI recovery (4) & 12,223 (98.38\%, 27.69\%) & 201 (1.62\%, 0.87\%) & 12,424 \\ 
  dead (6) & 2,330 (96.24\%, 5.28\%) & 91 (3.76\%, 0.39\%) & 2,421 \\ 
  permanent alteration (5) & 831 (86.47\%, 1.88\%) & 130 (13.53\%, 0.56\%) & 961 \\ 
  already dead or outside the reference colour range (0) & 626 (40.83\%, 1.42\%) & 907 (59.17\%, 3.92\%) & 1,533 \\ 
  not enough data (255) & 3 (75\%, 0.01\%) & 1 (25\%, 0\%) &   4 \\ 
     \hline
  TOTAL & 44,137 & 23,119 & 67,256 \\ 
   \hline
\end{tabular}
\end{lrbox}
\begin{tableorg}[ht]
\caption{Large tree crowns health status estimated from remote sensing within the Sequoias groves and their distribution in CAL FIRE fire polygons.}
\centering
\resizebox{\textwidth}{!}{\usebox{\tabgrovebox}}
\label{tab_fire_seq_only}
\end{tableorg}


The 97 giant sequoia groves contained 67,256 large trees, of which 44,137 (65.6\%) were located within the 2020--2024 fire perimeters (Table \ref{tab_fire_seq_only}). Note that these large trees are giant sequoias, but also the other large trees that grow in the groves. The alive class was the most common (36,026 crowns), and represented 41.6\% of the large trees inside the fire perimeters of the groves, against 76.5\% outside.

As in the rest of the Sierra Nevada, the disturbed classes were concentrated within the fire perimeters, which contained 98.4\% of the crowns of the breakpoint with $<$80\% NDVI recovery class, 96.2\% of the dead crowns and 87.1\% of the crowns of the breakpoint with $\geq$80\% NDVI recovery class. Among the latter, 65.8\% showed canopy opening (3b) inside the fire perimeters, against 33.1\% outside. Within the fire perimeters of the groves, 37.0\% of the large trees were disturbed (16,319 of 44,137), against 1.8\% outside (424 of 23,119). When the breakpoint with $\geq$80\% NDVI recovery and permanent alteration classes are also counted, these proportions were 40.9\% and 3.6\%, respectively.

With the shares of dead trees measured in the field, approximately 11,300 large trees (95\% CI: 10,600--11,900) are very likely dead within the fire perimeters of the groves, that is, 25.5\% (24.1--27.0\%) of the large trees within the fire perimeters and $\sim$16.8\% (15.8--17.7\%) of all the large trees of the groves. Outside the fire perimeters, this proportion was 1.2\% (1.1--1.3\%).


\FloatBarrier 
\subsection{Disturbance at the scale of individual trees: the North Complex Fire}


 \begin{figure}[!ht]
 \centering
 \includegraphics[width=0.98\linewidth]{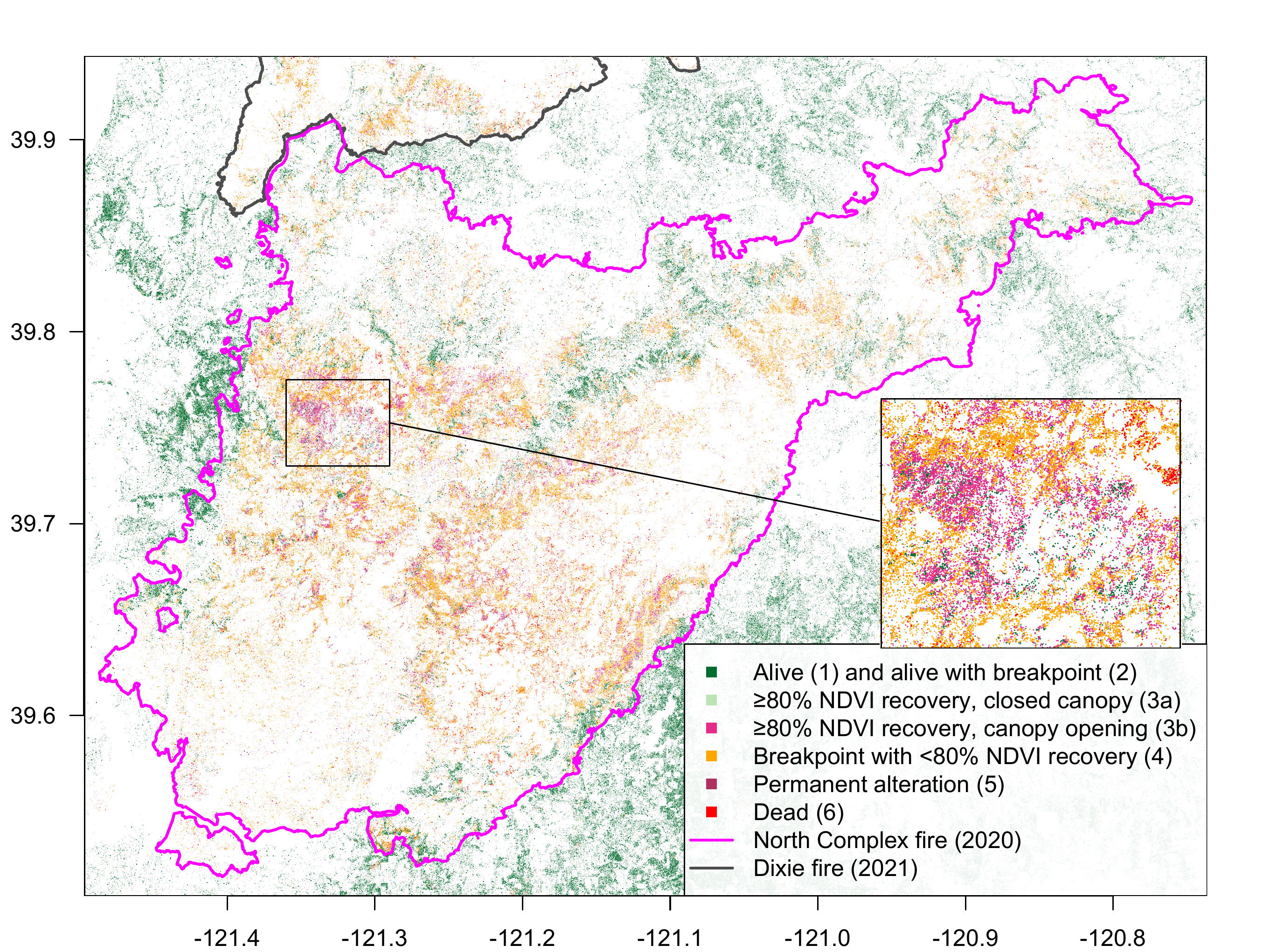}
\caption{Spatial pattern of large tree disturbance in the North Complex Fire. Only the main live/disturbed health classes are shown to ease visualisation. Each dot is a crown $\geq$ 100 m$^2$ (smaller trees are not mapped); white areas have no crown $\geq$ 100 m$^2$ or a crown of the classes already dead or outside the reference colour range (0) and not enough data (255). The perimeters of the North Complex Fire (2020) and of the Dixie Fire (2021) are from CAL FIRE; the inset is a zoom on the black rectangle.}
  \label{northcomplex} 
  \end{figure}

The North Complex Fire (2020) is the fire of the period with the most large trees within its perimeter (394,005), and is used here to show the pattern of disturbance at the scale of individual trees (Fig. \ref{northcomplex} and Table \ref{northcomplextab}). Inside the fire perimeter, 48.0\% of the large trees were classified as breakpoint with $<$80\% NDVI recovery, 17.6\% as breakpoint with $\geq$80\% NDVI recovery (14.7\% with canopy opening (3b) and 2.9\% under a closed canopy (3a)) and 5.6\% as dead. The dead trees formed localized hotspots, the largest one in the southeastern part of the fire, which were frequently embedded in larger areas dominated by the breakpoint with $<$80\% NDVI recovery class. Inside these impacted areas, pockets of alive trees remained, often mixed with trees of the breakpoint with $\geq$80\% NDVI recovery under a closed canopy (3a) class. Living crowns (alive and alive with breakpoint classes) remained common throughout the fire perimeter ($\sim$27\%).

\begin{table}[ht]
\caption{Health classes of the large crowns within the perimeter of the North Complex Fire (2020). N is the number of large crowns and \%col the share of the crowns of the fire that belong to the class. Subclasses 3a and 3b as in Table \ref{tab_fire}.}
\centering
\begin{tabular}{lrl}
  \hline
Class & Class ID & Burned (N, \%col) \\ 
  \hline

  alive & 1 & 51,687 (13.12\%) \\ 
  alive with breakpoint & 2 & 54,878 (13.93\%) \\ 
  breakpoint with $\geq$80\% NDVI recovery & 3 & 69,227 (17.57\%) \\ 
  \quad under a closed canopy & 3a & 11,269 (2.86\%) \\
  \quad with canopy opening & 3b & 57,958 (14.71\%) \\
  breakpoint with $<$80\% NDVI recovery & 4 & 188,927 (47.95\%) \\ 
  dead & 6 & 21,883 (5.55\%) \\ 
  permanent alteration & 5 & 4,228 (1.07\%) \\ 
  already dead or outside the reference colour range & 0 & 3,157 (0.80\%) \\
  not enough data & 255 & 18 (0\%) \\ 
  TOTAL &  & 394,005 \\ 
   \hline
\end{tabular}
\label{northcomplextab}
\end{table}

\FloatBarrier 
\section{Discussion}\label{sec12}

Here, we show that all the large trees of a mountain range can be mapped individually from a canopy height model, and that the health of each of them can then be followed with freely available satellite images. We delineated 6,515,705 trees with a crown area $\geq$ 100 m$^2$ in the Sierra Nevada Floristic Province, and found that 61.4\% of those located within the perimeters of the 2020--2024 wildfires were disturbed or dead, against 2.2\% outside. To our knowledge, this is the first time that the health status of the large trees of an entire region is described tree by tree.

The delineation of the crowns relied on three choices that differ from previous works. First, the crowns were delineated in a canopy height model and not in the images. In NAIP images, tall trees are displaced and deformed when the image is not acquired at nadir, whereas the canopy height model represents the trees as seen from above, as a LiDAR would \citep{Wagner2024}, and the crowns appear as smooth rounded shapes, which are easier to separate from their neighbours than in the images. Second, the model was trained exclusively with synthetic data. Previous studies relied on thousands of crowns delineated by hand \citep{Cheng2024,Ritz2026} or on a mix of manual and automatically generated labels \citep{Weinstein2019,Braga2020}. Here, 3,356,877 trees were simulated in a few hours, with labels that are always consistent, which cannot be obtained by manual annotation. Third, the U-Net-ID model reduces the delineation of individual trees to a segmentation followed by a simple polygonization and, unlike instance segmentation methods such as Mask R-CNN \citep{he2017mask,Braga2020,Weinstein2019}, the number of trees expected in an image does not have to be defined in advance. With these choices, the median IoU after centroid alignment was 0.602 for the giant sequoias and 0.605 for the large crowns without species identification drawn by hand. For comparison, an IoU of 0.53 was reported for the delineation of dead tree crowns in the same NAIP images of California \citep{Cheng2024}. The similar values obtained for the giant sequoias and for the other species were expected, as all large crowns, of conifers as of broadleaf trees, have the same rounded shape in the canopy height model.

The delineation has two systematic errors that must be kept in mind. The first is a displacement of a few metres between our crowns and the reference crowns (median of 3.2 m for the giant sequoias and 4.1 m for the hand-drawn crowns). It explains the low IoU, precision and recall obtained without alignment, and a part of it comes from the reference data, as crowns drawn on NAIP images are displaced when the image is not at nadir. The second is the underestimation of the crown area, by 60 m$^2$ on average for the giant sequoias. Because only the crowns $\geq$ 100 m$^2$ are kept, this underestimation removes from our dataset a part of the trees whose crown is slightly above this threshold: $\sim$40\% of the reference sequoia crowns $\geq$ 100 m$^2$ and $\sim$56\% of the hand-drawn ones were predicted below 100 m$^2$, mostly those between 100 and 200 m$^2$. Our count of 6.5 million large trees is therefore conservative, and should be read as the number of trees whose delineated crown reaches 100 m$^2$ rather than as an exhaustive census. Two elements moderate this bias. The reference crowns obtained from LiDAR are likely larger than the real crowns, as the segmentation allocates all the canopy space around a treetop to its crown, and the hand-drawn crowns of all sizes showed no bias (+4.7 m$^2$). Furthermore, the proportion of disturbed trees inside and outside the fire perimeters was similar for all crown sizes (Table \ref{tab_crown_size}), so the trees missed near the threshold are unlikely to change our conclusions on the health of large trees.

The 6.5 million large trees were found across 78.7\% of the province, which remains a major reservoir of large trees despite the drought, bark beetle outbreaks and high-severity fires of the last decades \citep{Cheng2024,Kane2023,Hung2026}. The only comparable census, based on airborne LiDAR, found more than 2.7 million very large trees (diameter $\geq$ 101.6 cm) in three landscapes covering $\sim$396,000 ha of the Sierra Nevada \citep{Kane2023}, and our map extends this type of census to the entire province. Large trees were mostly found between 600 and 1,700 m of elevation and in the wettest areas, and were almost absent above 3,000 m, while slope and orientation had little influence. This is consistent with the known distribution of the giant sequoias, on the western slope between $\sim$900 and 2,700 m \citep{Stephenson1996,Shive2026}, and of the other large conifers \citep{Griffin1976,Kane2023}. Note that wind, which may also influence the distribution of large trees, was not included in this analysis.

The Sentinel-2 classes were consistent with the status of the trees observed in the field, but not all classes have the same reliability. The alive classes were highly reliable, with fewer than 5\% of their trees dead on the ground. In the dead and breakpoint with $<$80\% NDVI recovery classes, $\sim$30\% of the trees were still alive on the ground. The crown of these trees was probably heavily scorched, and several years can pass before such a tree flushes again or dies \citep{Hanson2009,Jeronimo2020,Reilly2023}. Our classes describe what happened to the crown during the five years of observation, and a longer time series will be needed to know whether these trees survive or die. The opposite error was found in the breakpoint with $\geq$80\% NDVI recovery class, where 38\% of the trees were dead on the ground although their NDVI had recovered. In this case, the green signal comes from the shrubs and herbs that grow under the lost crown (Fig. \ref{fig_naip_compact}d), and the NDVI alone cannot separate them from a recovering crown. The change of summer brightness gives a part of the answer, as a living conifer crown is dark and the vegetation exposed after its loss is bright: 49.1\% of the crowns with canopy opening were dead on the ground, against 20.0\% of the crowns under a closed canopy. While the NDVI can be restored by a few years of shrub growth, the shade of a crown cannot come back without the crown. Note that the reference colour envelope built from giant sequoia crowns did not limit the analysis of the other species, as 89.4\% of all the large crowns fell within it.

Fire effects on vegetation are commonly mapped with the difference of a spectral index before and after the fire, such as the dNBR and its relativized versions \citep{Miller2007,Miller2009,Parks2018,Shive2026}. These indices require the date and the perimeter of the fire. Our approach differs in that the time series of each tree is analysed without any information on the fires: the date of the change is found by the algorithm, the classes describe the trajectory of the crown after the change, and the fire perimeters are used only afterwards, to compare the trees inside and outside them.

The contrast between burned and unburned areas leaves little doubt about the role of fire in the recent disturbance of large trees. The proportion of disturbed large trees was 27 times higher inside the fire perimeters than outside, whatever the crown size, and for each of the eight largest fires it dropped from 46.5--74.9\% inside the perimeter to 1.5--4.3\% a few kilometres away. After correction with the field validation, we estimate that $\sim$650,000 large trees (615,000--689,000), that is, one in ten of the large trees of the Sierra Nevada, very likely died within the fire perimeters in five years. Most of them (88\% of the burned large trees) were in the perimeters of the 2020 and 2021 fires, two fire seasons associated with severe drought, high temperatures and exceptionally dry fuels \citep{Safford2022,Keeley2021,Kennedy2021,Williams2023,Hung2026}. These results are in line with stand-scale studies, which reported wildfire as the main cause of large forest disturbances in California \citep{Favrichon2025} and the highest rates of forest loss in the Sierra Nevada during the same years \citep{Carter2024}. The low proportions of disturbed trees for the fires of 2023 and 2024 (9.1\% and 2.2\%) should not be interpreted as a lower impact of these fires, as less than two years of observations were available after them and delayed mortality is likely not yet detected. Outside the fire perimeters, the proportion of large trees very likely dead (1.5\%) is an upper bound, and reflects other causes of mortality such as drought, insects, disease or logging, which were already important before the 2020 fire season \citep{Cheng2024}.

Inside the fire perimeters, the effects of fire were far from uniform. In the North Complex Fire, patches where nearly all the large trees were classified as dead were found next to pockets of surviving trees, and living crowns remained common over large areas (Fig. \ref{northcomplex}). Such fire refugia, unburned or lightly burned, are known to persist within fire perimeters and to support the recovery of the forest \citep{Blomdahl2019,Paudel2023,Dixon2023}. A part of the large trees classified as alive inside the perimeters are therefore probably trees that did not burn, as the CAL FIRE perimeters give the extent of the fires and not their severity. In the giant sequoia groves, we estimate that $\sim$17\% of the large trees very likely died within the fire perimeters. This value is close to the 17.6\% of large giant sequoias killed by fire since 1984 (13.9\% during the 2020 and 2021 fire seasons) \citep{Shive2026}, but the two estimates are not directly comparable, as ours includes the large trees of all species of the groves over 2020--2024. The proportion of disturbed large trees inside the fire perimeters was lower in the groves than in the entire province (37.0\% against 61.4\%). Our data do not allow us to attribute this difference to a cause, and the prescribed burns conducted in several groves, which reduced the mortality of giant sequoias in 2020 and 2021 \citep{Dixon2026}, are one possible explanation.

Our estimate of the number of dead trees has several limitations. First, the shares of dead trees per class were measured on giant sequoias, a fire-resistant species, and applied to the large trees of all species. The visual interpretation of NAIP images, which includes all species, gave higher shares of dead trees for the class 3 crowns than the field data, so our estimate is likely conservative for the other species. Second, these shares depend on the proportion of dead trees in the field sample (44\%). They overestimate mortality where dead trees are rare, which is why the value obtained outside the fire perimeters is only an upper bound. Third, the 95\% intervals reflect only the size of the field sample. They do not include the errors of the crown delineation or the uncertainty of the brightness threshold, which was set with a limited number of crowns. Fourth, the signal of a crown of 100 m$^2$ is measured in a single Sentinel-2 pixel of 10 m, which could also contain a part of the neighbouring trees and of the understory. The classes might therefore describe, to some extent, the tree and its immediate surroundings, although the results were similar for all crown sizes. Fifth, the PERMANOVA treats the observations before and after the breakpoint as independent, whereas they are temporally autocorrelated. The p-values are therefore optimistic, and the classes rely mainly on the magnitude of the changes in colour and NDVI after the breakpoint. Finally, we compared the trees inside and outside the fire perimeters without accounting for the severity of the fire, which could be explored in future work.

In this study, we present the first map of all the large trees of the Sierra Nevada and of their health status after the fires of 2020--2024. The delineation model was trained without any manual label and the health classes were derived from freely available Sentinel-2 images. This approach is low-cost, scalable and can be repeated through time. The resulting dataset shows where the large trees are, which ones were lost and which ones survived inside the burned areas, and could be used to locate the surviving large trees of primary importance for conservation. The next steps are to (i) extend the individual tree mapping to all trees across the United States, (ii) apply the framework to the redwood forests and other large coniferous forests of California, (iii) integrate additional structural and spectral information, such as fire severity, to improve the characterization of disturbances, and (iv) assess whether similar approaches can be used to monitor tree condition, mortality and recovery globally with Sentinel-2.

\FloatBarrier 
\section{Materials and Methods}

\begin{figure}[ht]
\centering
\includegraphics[width=1.0\linewidth]{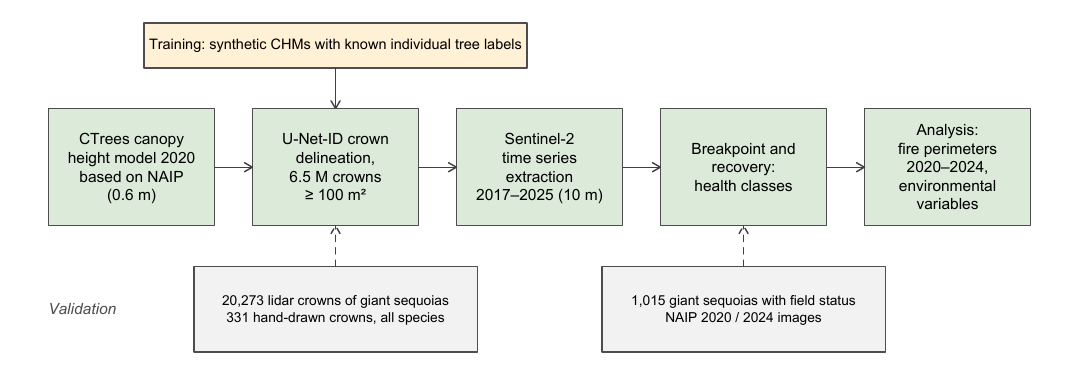}
\caption{Workflow of the study: delineation of the large tree crowns, extraction of their Sentinel-2 time series, health classes and analysis, with the datasets used for validation.}
\label{fig_workflow}
\end{figure}

The workflow of the study is summarized in Fig. \ref{fig_workflow}.

\FloatBarrier 
\subsection{Study site}

\begin{figure}[ht]
 \centering
\includegraphics[width=0.97\linewidth]{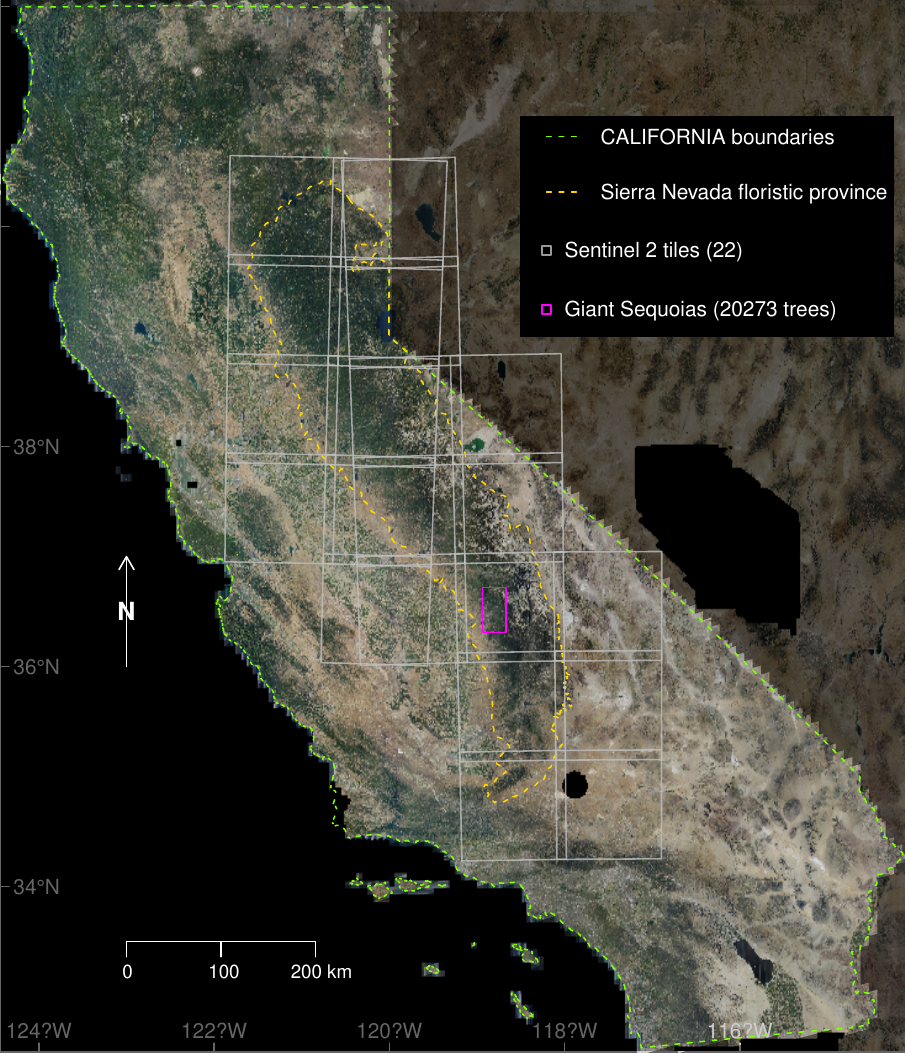}
\caption{Mosaic of NAIP aerial imagery acquired over California (USA) during the 2020 campaign, with the Sierra Nevada Floristic Province outlined in dashed yellow ($\sim$65,582 km$^2$). The map also shows the extent of the 22 Sentinel-2 MSI tiles used for time-series analysis (grey) and the extent of the 20,273 \replaced[id=USGS]{sequoia}{Sequoia} polygons used for individual tree crown model validation (magenta).}
  \label{Fig1}
  \end{figure}

The study covered the Sierra Nevada Floristic Province (Fig. \ref{Fig1}), located in eastern California between 34.7$^o$ and 40.5$^o$ N \citep{Burge2021}. The province spans approximately 620 km from the southern Cascades to the Tehachapi Mountains and covers 65,582 km$^2$, that is, $\sim$16\% of California. Elevations range from 30 to 4,421 m at Mount Whitney, the highest peak of the contiguous United States \citep{farr2007shuttle,NASA2026}, and the vegetation is mainly composed of mixed-conifer forests \citep{Barbour1988,VanWagtendonk2018}. Annual precipitation ranges from $\sim$170 to 2,390 mm, and the annual means of minimum and maximum temperatures range from $-$8.8 to 12.4$^o$C and from 2.6 to 25.8$^o$C, respectively \citep{Daly2008,PRISM2026}.

\FloatBarrier 
\subsection{Individual tree delineation}

To map all the large trees of the Sierra Nevada, we delineated the crown of each tree in a canopy height model (CHM) and kept the crowns with an area $\geq$ 100 m$^2$. This threshold was chosen to match the area of one Sentinel-2 pixel (10 $\times$ 10 m), used to extract the time series of each crown. The CHM is the model produced in our previous work for the entire state of California from the NAIP images of 2020 at 0.6 m spatial resolution \citep{Wagner2024}, in which individual tree crowns are visible and identifiable (see Fig. 7 and Supplementary Figures in \cite{Wagner2024}). The NAIP 2020 images of the Sierra Nevada were acquired between 1 May and 5 August 2020, before the large fires of 2020. To delineate the trees in the CHM, we treated crown delineation as a blob segmentation problem. First, we generated realistic synthetic CHMs from simulated tree crowns. Second, we trained a U-Net-ID deep learning model with this synthetic dataset to identify the individual crowns in a CHM. Finally, the model, trained entirely on synthetic data, was applied to the real CHM. The main advantage of this approach is that a wide range of tree sizes, heights, densities and spatial configurations can be simulated, so that the model learns to delineate tree crowns without any manually labelled training data or additional LiDAR processing.

\subsubsection{Generation of synthetic canopy height models}
\label{simmetm}

We generated synthetic canopy height models (CHMs) by simulating 3D individual tree crowns and merging them into a LiDAR-like CHM, so that the location, size and geometry of each tree are known. Each simulated forest is an image of 128 $\times$ 128 pixels at 0.6 m resolution, the resolution of our CHM of California, with heights stored as 8-bit integers scaled by 2.5 \citep{Wagner2024}. Trees of eleven functional types (conifers, deciduous trees, saplings, shrubs, giant conifers, giant deciduous trees and giant sequoias), each with its own range of height and crown radius, were placed at random with a minimum distance between them, following forest scenarios of different composition and density, from shrublands to giant sequoia forests. Conifers and giant trees have conical crowns, and broadleaf trees and shrubs rounded crowns. Only the highest point per pixel is kept, as in a LiDAR CHM, and a random blur was added as data augmentation. The generated dataset consists of 59,134 synthetic CHM images containing 3,356,877 simulated trees. The scenarios, functional types and crown shapes, and examples of simulated images, are given in Appendix \ref{secA_synthetic}.

\FloatBarrier 
\subsection{Neural Network Architecture}

 \begin{figure}[ht]
 \centering
 \includegraphics[width=0.90\linewidth]{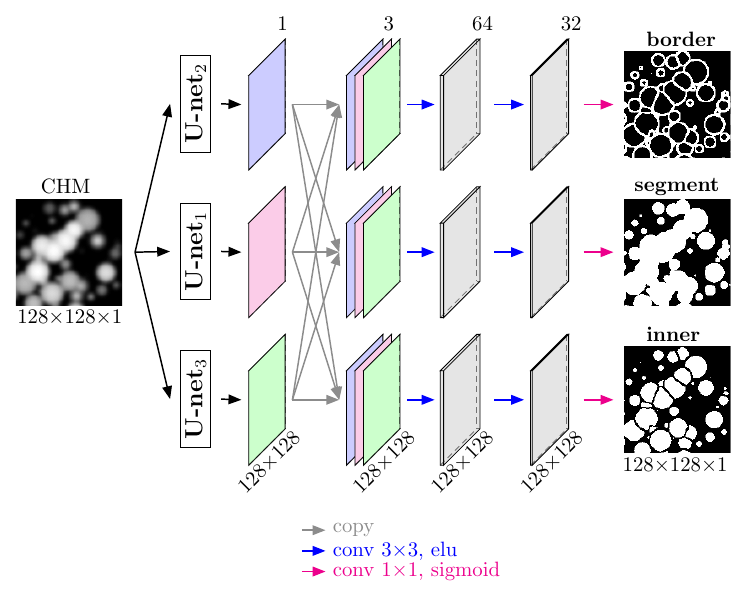}
  \caption{U-Net-Id model architecture used for individual tree segmentation from Canopy height model, adapted from Wagner et al.~\cite{Wagner2020UnetID}. The three U-net units used in our model have the same layers as in Wagner et al.~\cite{Wagner2020UnetID} only that the input and output of the U-net units here are 128 $\times$ 128. The outputs of the U-net units are represented with three different colors: magenta (U-net$_1$), blue (U-net$_2$), and green (U-net$_3$). The number of channels is indicated above the cuboids, and the vertical numbers indicate the row and column size in pixels. The operations (skip connections and convolutions) performed in each layer and their sizes are indicated by the colored arrows.}
  \label{FigUnetID}
  \end{figure}

The individual tree segmentation from the Sierra Nevada NAIP-derived CHM images in California was performed using the U-Net-Id model, Fig. \ref{FigUnetID} \citep{Wagner2020UnetID, Wagner2024}. The U-Net-Id model is designed to convert an instance segmentation problem into a semantic segmentation problem of binary masks, which is well resolved using U-Net-like architectures \citep{Ronneberger2015}.

The inputs of the U-Net-ID model are synthetic canopy height model images of 128 $\times$ 128 pixels at a spatial resolution of 0.6 m, Fig. \ref{FigUnetID}. The image is processed through three independent paths, each beginning with a separate U-Net model \citep{Wagner2020UnetID}. The three U-Nets predict the binary masks corresponding to the object border, the object segment, and the object inner segment. These training masks are generated directly from the simulated individual tree polygons that were used to produce the simulated CHM image  (see section \ref{simmetm}). The object segment corresponds to the original polygon, the border is defined as the area between a buffer of 0.6 m and -0.6 m around the polygon boundary, and the inner segment is obtained by applying a negative buffer of 0.6 m to the original polygon. The resulting masks therefore overlap along the object boundary, creating transition zones equivalent to $\sim$ 1 to 3 pixels in width. The inner segment is individual to each tree. The three 128 $\times$ 128 activation maps produced by the U-Nets are then concatenated. This is followed by two convolutional layers with 64 and 32 filters, respectively. Finally, separate predictions for the segment, border, and inner segment are produced using a final convolutional layer with a sigmoid activation function. 

The instance individualization is performed in post-processing by extracting the inner segments, which are unique and do not touch each other, and expanding them with a 1-pixel (0.6 m) buffer. The resulting objects are then combined with predicted segments that do not intersect any inner segment. These isolated segments correspond to the smallest trees, with crown sizes of only 1–4 pixels, which do not have an inner segment. Finally, a unique ID is added.

\subsubsection{Training and testing}


The training sample comprised a total of 59,134 patches of 128 $\times$ 128 \replaced[id=USGS]{pixel}{pixels} one-band CHM images and their associated three-band mask \replaced[id=USGS]{patches}{patchs}. 90\% (53,372) of the samples were used for training and 10\% (5,762) for internal model testing. Before being inputted into the U-Net model, image patches underwent random vertical and horizontal flips as data augmentation.


The network was trained for 2,048 epochs with a batch size of 128 images. The loss function was designed as a sum of two terms: binary cross-entropy and Dice coefficient-related loss of the three predicted masks \citep{Dice1945,AllaireChollet2019, chollet2015keras}.
During network training, we used standard stochastic gradient descent (SGD) optimization with the RMSprop optimizer (learning rate of 0.0001). The SGD procedure minimizes the loss function using random mini-batches of data, enabling faster convergence and improved generalization \citep{chollet2018deep}. The model with the best validation loss (i.e., Dice coefficient-related loss of 0.02324226 and a Dice coefficient of 0.97362) was kept for prediction. The training of the model took around 12 hours using an Nvidia RTX4090 Graphics Processing Unit (GPU) with 24 GB of memory.

 \subsubsection{Prediction}
 
For prediction, the 1,885 NAIP-based CHM tiles covering the Sierra Nevada Floristic Province were resized by adding columns and rows, resulting in images of 11,392 $\times$ 13,440 pixels, that is, a multiple of 1,024 pixels plus a 128-pixel border. This adjustment was made to meet the input size requirements for prediction. The standardized-size NAIP tiles were subsequently divided into sub-images of 1,152 $\times$ 1,152 pixels with a 64-pixel overlap between them. Predictions were made on these 1,152 $\times$ 1,152-pixel images, and the 64-pixel border on each sub-image prediction was removed to mitigate border artifacts \citep{Ronneberger2015}. The resulting 1024 $\times$ 1024 images were merged and cropped to the original NAIP extent, and, finally, a 60-pixel border (36 m) was removed to mitigate any remaining border artifacts. This last step did not affect the final crown map since NAIP images have an overlap of $\sim$ 400 meters between them. The prediction for the 1,885 tiles took around 13 hours using the RTX4090 GPU.

\subsubsection{Instance polygonization and tree crown statistics}

Polygonisation of individual tree crowns was performed in post-processing using the \texttt{terra} and \texttt{sf} R packages \citep{Hijmans2026,Pebesma2018,Pebesma2023}. Individual crowns were defined as either (i) inner segments buffered by 0.6 m to reconstruct full crowns size, or (ii) isolated predicted segments that do not intersect any inner segment, i.e. very small trees.

As NAIP tiles overlap, the crowns touching the left and bottom edges of each tile of the NAIP tiling grid were removed, so that each tree is kept in only one tile and counted once.

For each individual crown, a unique identifier was assigned. Centroids were extracted for each polygon, crown area computed, and crown radius was approximated assuming circular crowns ($radius = \sqrt{area/\pi}$). Height statistics were then computed from the extracted CHM raster values for each polygon, including mean, median, minimum, and maximum height values.

\subsubsection{Instance validation}

The delineation was first validated with a reference dataset of 20,273 giant sequoia crowns \citep{Dixon2026}. These crowns were delineated in a 1 m CHM derived from airborne LiDAR data collected in August 2015 and July 2016, with an unsupervised region-growing algorithm of the lidR package applied from treetops detected as local maxima within a 6 m window \citep{Dalponte2016, Roussel2020}. Only the trees with a one-to-one correspondence between the crown polygon and a field-validated stem location were retained (see Dixon et al.~\cite{Dixon2026} for a full description of the dataset). The reference crowns predate our 2020 CHM by four to five years. As the dataset only includes mature giant sequoias that were alive before the 2020--2021 fires \citep{Dixon2026}, whose crown dimensions are expected to change little over this period, we considered this time gap negligible. Note that the region-growing segmentation allocates the canopy space around each treetop to its crown (64.3\% of the reference crowns touch a neighbouring crown), so the reference crowns are likely larger than the real crowns.

Because our dataset contains many more trees than the reference, giant sequoias as well as other species, each reference crown was matched to the intersecting predicted crown with the closest crown area, each predicted crown being used only once. The agreement between the reference and predicted crowns was evaluated with the root mean square error (RMSE), the Pearson correlation coefficient ($r$) and the bias of the crown area (predicted minus observed), with the distance between the crown centroids, and with the intersection over union (IoU). As the crowns can be displaced by a few metres, the IoU was computed without and with centroid alignment, that is, after translating the predicted crown to the centroid of the reference crown. The alignment was only used to measure the agreement of crown shapes; the crowns of our product were not shifted. Finally, to quantify how the underestimation of the crown area affects the 100 m$^2$ threshold, we computed the proportion of the reference crowns $\geq$ 100 m$^2$ whose matched predicted crown was $<$ 100 m$^2$, for all the crowns and by class of reference crown area (100--200, 200--300 and $>$ 300 m$^2$).

To assess the delineation on other species, we also used crowns without species identification drawn by hand on the NAIP 2020 images of 11 NAIP tiles of the Sierra Nevada (20,393 crowns, of which 331 had a crown area $\geq$ 100 m$^2$). As these crowns were drawn on the NAIP images, they are displaced and deformed where the image is not at nadir, whereas our crowns are delineated on the CHM, a nadir representation. The same matching and metrics as for the giant sequoias were applied. In addition, we computed the precision, recall and F1-score of the large crowns, without and with centroid alignment. A predicted crown $\geq$ 100 m$^2$ was counted as a true positive when its matched reference crown was $\geq$ 100 m$^2$ and their IoU was $\geq$ 0.5. Precision is the number of true positives divided by the number of predicted crowns $\geq$ 100 m$^2$, and recall the number of true positives divided by the number of reference crowns $\geq$ 100 m$^2$.

\FloatBarrier 
\subsection{Sentinel-2 data cubes preparation and analysis}
We prepared Sentinel-2 L2A surface reflectance data cubes for the 22 Sentinel-2 tiles that intersect the Sierra Nevada with the \texttt{sits} package \citep{Simoes2021}, from the collection hosted on Amazon Web Services. We used the blue (B02), green (B03), red (B04), near-infrared (B08), shortwave infrared (B11 and B12) and cloud mask (CLOUD) bands from December 2017 to February 2025, and kept only the images with a cloud cover $\leq$ 5\% over the tile. The images of each tile were then regularized to a time step of 16 days at 10 m spatial resolution. The download procedure is described in Appendix \ref{secA_sits}.

\subsubsection{Extraction and filtering of the time series}

The Sentinel-2 time series of each large crown was extracted at the 10 m pixel containing the crown centroid, for the bands B02, B03, B04, B08 and B11. The CHM (0.6 m), the crown polygons and the regularized Sentinel-2 cubes (10 m) were only reprojected to a common coordinate system, without further co-registration.

\begin{figure}[ht]
 \includegraphics[width=0.45\linewidth, trim={2.0cm 2.5cm 2.5cm 2.5cm},clip]{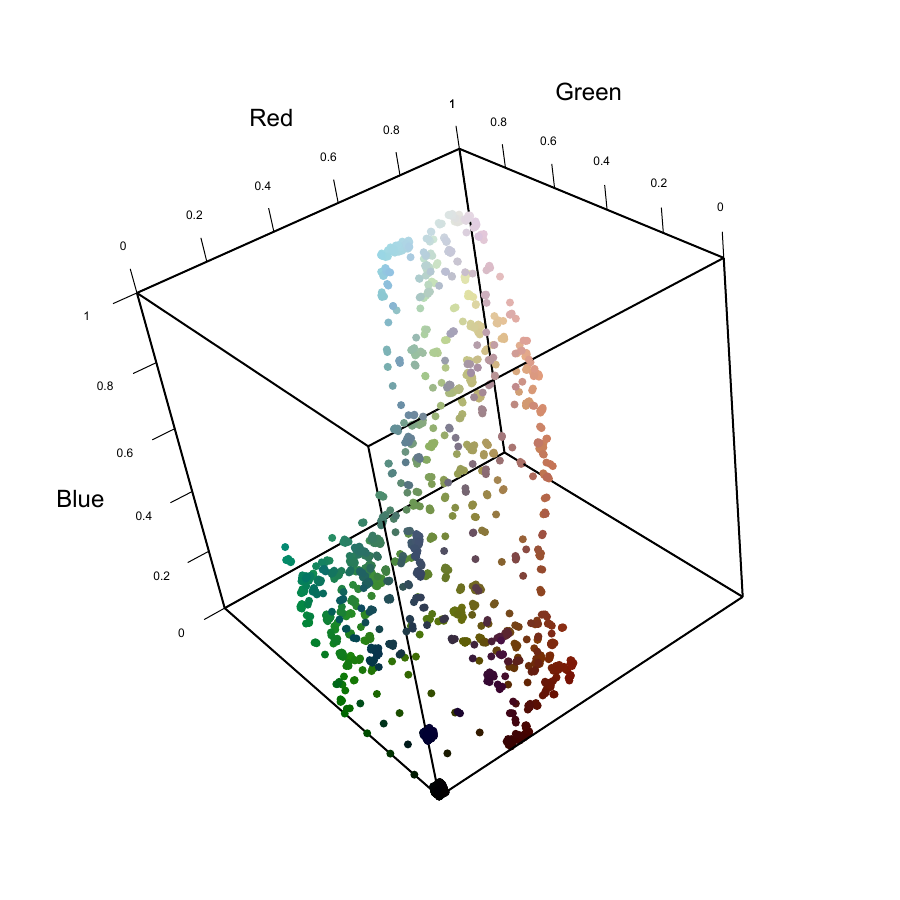}%
 \begin{picture}(0,0)
\put(-22,162){\textbf{(a)}}
\end{picture}
 \includegraphics[width=0.45\linewidth, trim={2.0cm 2.5cm 2.5cm 2.5cm},clip]{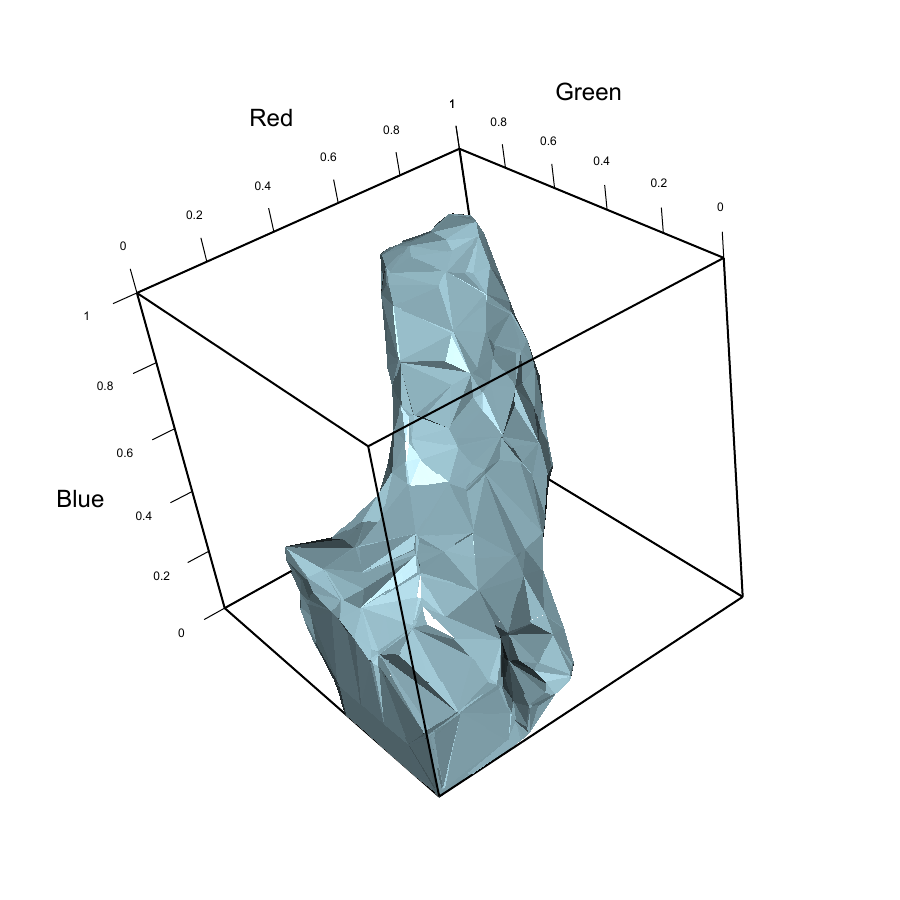}%
 \begin{picture}(0,0)
\put(-22,162){\textbf{(b)}}
\end{picture}
\caption{Colours of the giant sequoia crowns observed in the Sentinel-2 images and used to define the reference colour envelope (a), and 3D mesh of this envelope, used to filter the Sentinel-2 time series of all the large crowns (b).}
  \label{Mesh}
  \end{figure}

These time series still contain spurious values caused by clouds, shadows, snow and atmospheric artefacts, which were removed in two steps. First, the exact zeros and the very low reflectance values (digital numbers of $\sim$1--10 in the Sentinel-2 product), which correspond to invalid observations, were replaced in each band by the last valid value of the time series. Second, we kept only the observations whose colour is consistent with the colour of a tree crown. To define this colour range, we used the giant sequoia crowns of the reference dataset that matched one of our crowns, for which we know that the pixel contains a real tree. From their time series, we selected by visual inspection the observations free of clouds, shadows and artefacts, and converted their red, green and blue reflectances (B04, B03 and B02) to 8-bit values (Fig. \ref{Mesh}a). As some of these trees burned during the study period, the colours range from the green of living crowns to the brown of dead crowns. A closed three-dimensional mesh enclosing these points in the RGB space was then built (Fig. \ref{Mesh}b), hereafter the reference colour envelope. For all the large crowns, the observations whose RGB values fell outside this envelope were discarded.

This reference colour envelope, built from giant sequoia crowns, is not a species filter. It does not separate the giant sequoias from the other large conifers, and 89.4\% of all the large crowns fall within it. It acts as a filter of observations with a plausible canopy colour. Only the crowns that were already dead, or that have a different crown colour such as the broadleaf trees of the lower slopes, fall outside it and form the class already dead or outside the reference colour range (0).

\hspace{1cm}

\subsubsection{Breakpoint detection and health classes}
\label{classes_def}


For each crown, the Normalized Difference Vegetation Index (NDVI) was computed from the filtered red (B04) and near-infrared (B08) reflectances as:

\begin{eqnarray}
NDVI&=& \frac{B08-B04}{B08+B04}   
\end{eqnarray}

Abrupt changes in the NDVI time series were detected with the \texttt{bfast} algorithm \citep{Verbesselt2010, Verbesselt2010a}. The method decomposes the time series into seasonal and trend components with a harmonic seasonal model and identifies the most likely breakpoint in the trend component. We used a single breakpoint, one iteration and a seasonal-trend decomposition with \texttt{stlplus} (\texttt{season = harmonic}, \texttt{breaks = 1}, \texttt{max.iter = 1}, \texttt{decomp = stlplus}).

When a breakpoint was detected, the time series was divided into the observations before and after the breakpoint, and the mean red, green and blue reflectances and the mean NDVI were computed for each period. The difference in colour between the two periods was tested with a permutational multivariate analysis of variance (PERMANOVA) based on Euclidean distances, with 999 permutations. A breakpoint was considered significant when the PERMANOVA gave p $<$ 0.05 and when it occurred after 1 January 2020, as the study focuses on the period 2020--2025.

The recovery after a significant breakpoint was characterized with a linear regression of the NDVI observations following the breakpoint against time. The slope of this regression gives the rate of NDVI change after the disturbance, and its significance was assessed with a t-test. We also computed the percentage of NDVI recovery, that is, the share of the NDVI lost at the breakpoint that was regained at the end of the time series.


A breakpoint can also be caused by residual noise or by a short disturbance without consequence for the crown. To identify these cases, we examined the mean red and green reflectances after the breakpoint. The crowns whose colour remained green, that is, with a green reflectance still higher than the red reflectance, were considered alive, whereas the crowns that turned brown were considered disturbed.


Each large crown was then assigned to one of eight classes according to the trajectory of its time series (Fig. \ref{classes}). Two classes have no significant breakpoint. The alive (1) class corresponds to healthy living trees of any species, with a mean NDVI close to 0.8 and an annual seasonality (Fig. \ref{classes}a and Fig. \ref{fig_naip_class1}). The already dead or outside the reference colour range (0) class includes the crowns whose colour is not green and whose mean NDVI is below 0.55 during the entire period (Fig. \ref{classes}d and Fig. \ref{fig_naip_class0}). These crowns, with NDVI values close to 0.4, are trees that were already dead before 2020 or trees with a naturally different crown reflectance, such as the large oaks of the lower slopes of the Sierra Nevada.

The other classes have a significant breakpoint. In the alive with breakpoint (2) class, the colour of the crown remained green after the breakpoint, which indicates that the crown is alive despite a short disturbance signal (Fig. \ref{classes}b and Fig. \ref{fig_naip_class2}). For the crowns that turned brown, the class depends on the evolution of the NDVI after the breakpoint. When the NDVI increased significantly, the crown was assigned to the breakpoint with $\geq$80\% NDVI recovery (3) class if the percentage of NDVI recovery was $\geq$ 80\% (Fig. \ref{classes}c and Figs. \ref{fig_naip_class3a} and \ref{fig_naip_class3b}), and to the breakpoint with $<$80\% NDVI recovery (4) class otherwise (Fig. \ref{classes}e and Fig. \ref{fig_naip_class4}). The crowns of class 3 recovered rapidly to values close to their original NDVI and were first interpreted as recovered. As this green signal can also come from the understory seen through a lost crown, this class was split into two subclasses (see below). When the NDVI did not increase significantly, the crown was assigned to the dead (6) class if its mean NDVI after the breakpoint was $\leq$ 0.55, which corresponds to a persistent brown colour (Fig. \ref{classes}f and Fig. \ref{fig_naip_class6}), and to the permanent alteration (5) class if it remained above 0.55 (Fig. \ref{classes}g and Fig. \ref{fig_naip_class5}). In this last class, the crown is altered and does not recover, but keeps a part of its green signal, which may indicate a partial loss of the crown, decline, stress or disease. Finally, the crowns whose time series did not contain enough data for the BFAST algorithm form the not enough data (255) class.

The NDVI recovery of the breakpoint with $\geq$80\% NDVI recovery (3) class can come either from the crown itself or from the shrubs and herbs that grow under a crown lost to fire and become visible in the 10 m Sentinel-2 pixel. To separate these two cases, we used the change of summer brightness of the crown pixel. A living conifer crown is dark in all bands, because of the absorption by the needles and of the shade within the crown, whereas the shrubs, herbs and soil exposed after the loss of the crown are brighter, mainly in the NIR and SWIR. Summer brightness was computed as the median, over the June to September observations, of the mean reflectance of the B02, B03, B04, B08 and B11 bands (B11 being resampled from 20 m to 10 m in the data cube), and its change as the value of summer 2024 minus the value of the summers 2018 and 2019. The crowns of class 3 with a change of summer brightness $>$ 0.025 were assigned to the subclass breakpoint with $\geq$80\% NDVI recovery and canopy opening (3b), and the others to the subclass breakpoint with $\geq$80\% NDVI recovery under a closed canopy (3a). The threshold was determined in two steps. It was first set to 0.030 by maximizing the sum of sensitivity and specificity (AUC of 0.80, 95\% CI: 0.76--0.83) on 998 class 3 crowns burned in 2020 and 2021, considered as lost when their mean height in the CTrees canopy height model of 2022 was below half of that of 2020. It was then lowered to 0.025 after the visual interpretation of NAIP images (see below), in which 19 of the 20 crowns between 0.025 and 0.030 were dead. As the change of summer brightness requires at least one summer of observations after the fire, the subclasses are not informative for the fires of late 2024.

NAIP aerial images of 2020 and 2024 (0.6 m spatial resolution; red, green, blue and NIR bands) were used to check the interpretation of the Sentinel-2 classes, as they show the crowns before and after the fires. A random sample of 100 class 3 crowns burned in 2020 and 2021, with a change of summer brightness $\leq$ 0.030, was interpreted visually as dead, alive or unsure, without any Sentinel-2 information. Examples of crowns of each class in the NAIP images of 2020 and 2024 are given in Fig. \ref{fig_naip_compact} and in Figs. \ref{fig_naip_class1}--\ref{fig_naip_class0} (Appendix \ref{secA_naip}). Note that tall trees lean in NAIP images that are not taken at nadir, so the crown can be displaced from its polygon; the visual interpretation was therefore made on the crown itself rather than strictly within the polygon.

\begin{figure}[ht]
\centering
\includegraphics[width=0.90\linewidth]{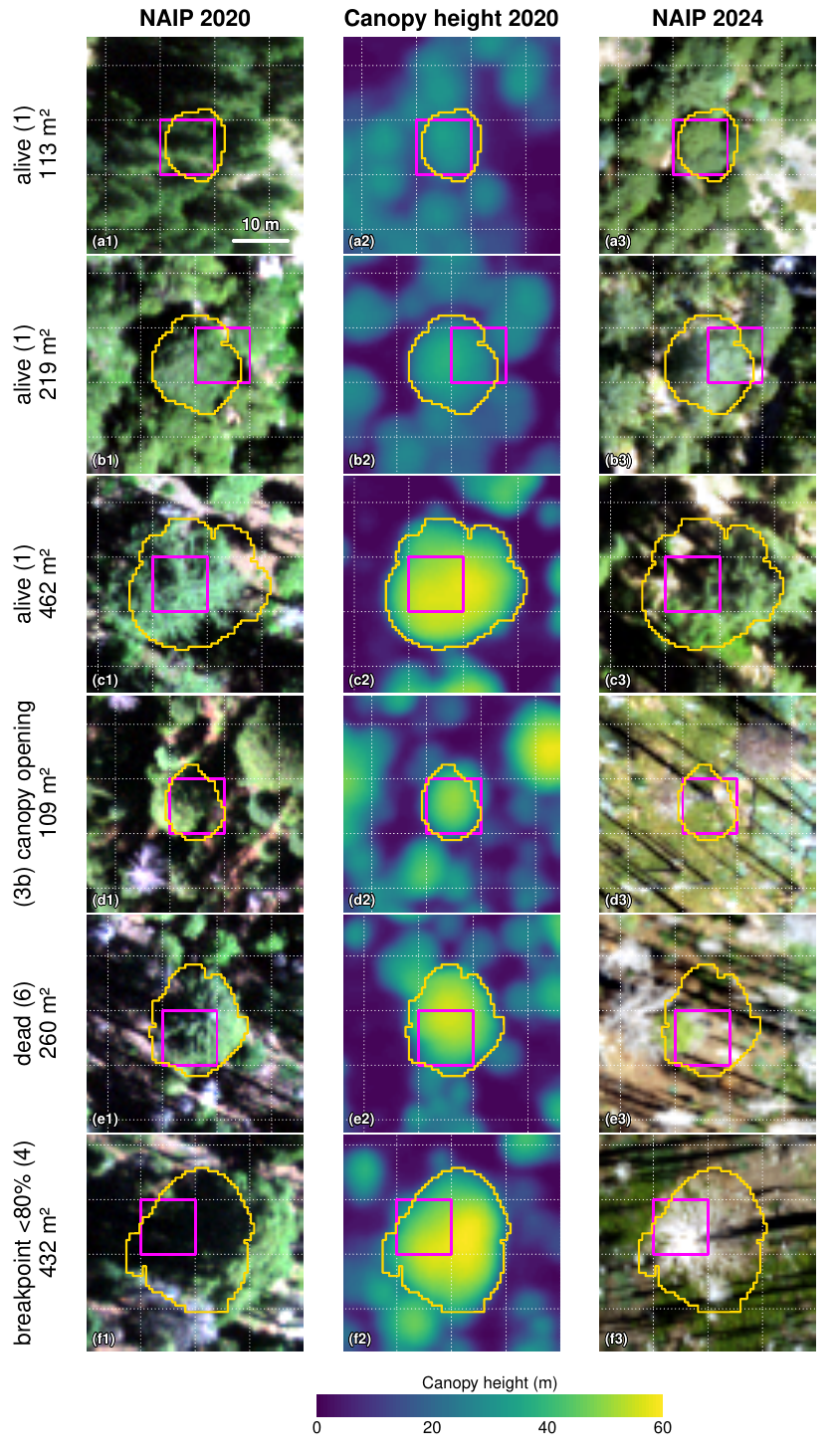}
\caption{Alignment of the datasets for six large crowns: NAIP 2020 (left), canopy height model of 2020 derived from NAIP 2020 (centre) and NAIP 2024 (right). Gold: crown polygon delineated from the canopy height model; white dotted lines: 10 m Sentinel-2 grid; magenta: Sentinel-2 pixel containing the crown centroid, used for the time series of the crown. Rows (a)--(c): alive (1) crowns outside the fire perimeters; rows (d)--(f): crowns burned in 2020 or 2021. Each box is 40 $\times$ 40 m.}
\label{fig_alignment}
\end{figure}

\begin{figure}[ht]
 \centering
\includegraphics[width=0.49\linewidth]{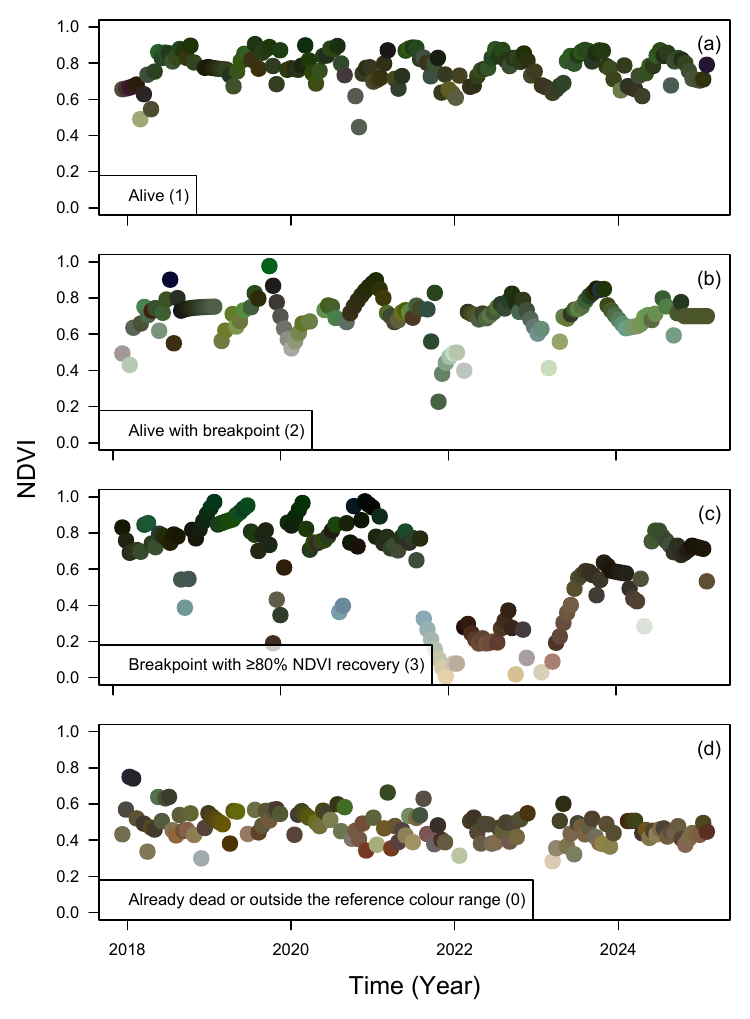}
\includegraphics[width=0.49\linewidth]{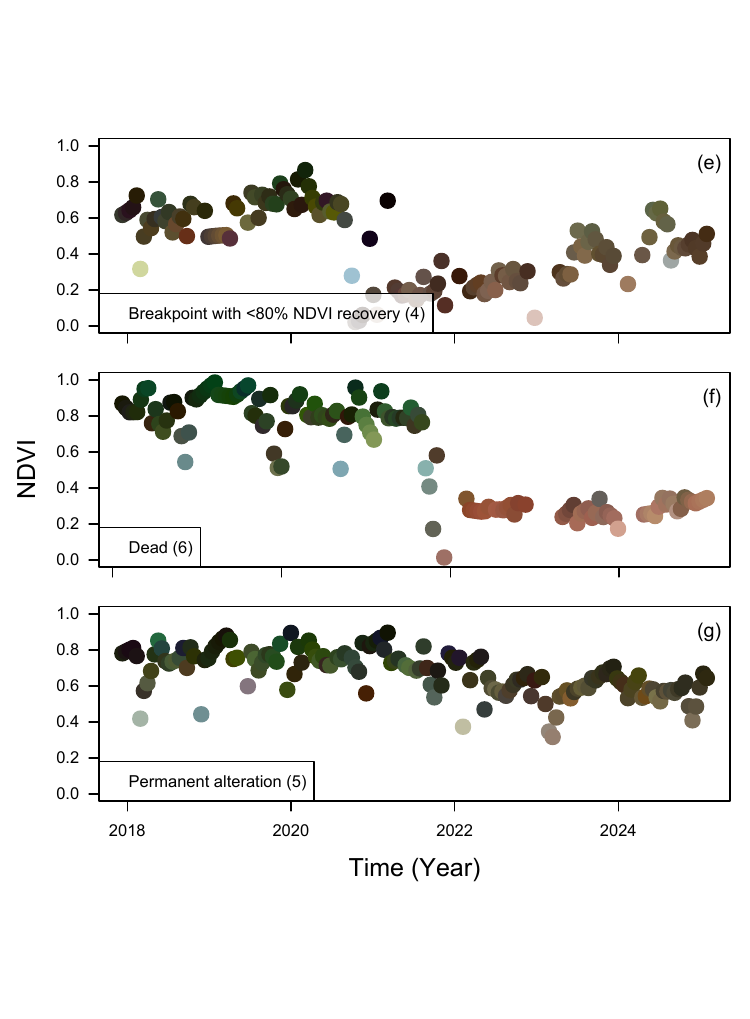}
\caption{Examples of the seven time series trajectories identified in the dataset and used to classify the large crowns into alive or disturbed classes. The class breakpoint with $\geq$80\% NDVI recovery (3) is further split by the change of summer Sentinel-2 brightness into closed canopy (3a) and canopy opening (3b) (Section \ref{classes_def}).}
  \label{classes}
  \end{figure}

The alignment between the NAIP images, the CHM in which the crowns were delineated and the Sentinel-2 pixel used for the time series is illustrated for six large crowns in Fig. \ref{fig_alignment}. As no independent measurement of the co-registration between these datasets is available, this figure shows how the crown polygons match the crowns visible in the NAIP images and in the CHM, and which part of the crown is covered by the 10 m Sentinel-2 pixel. Only one Sentinel-2 pixel was extracted per crown, for three reasons. First, the volume of Sentinel-2 data to process is already very large, several terabytes for the 22 tiles. Second, our aim was to test whether the health status of a tree can be followed with the pixel of its crown, and it makes little sense to add pixels that are further from the tree and whose signal could be independent of the crown. Finally, the NAIP and Sentinel-2 images are both well georeferenced \citep{USGSNAIP2026,ESAS2L2A2026}, so that, on average, the pixel containing the crown centroid should fall on the right tree, as can be seen in Fig. \ref{fig_alignment}.

\newpage

To assess how our classes correspond to the health of the trees observed in the field, we used the assessment of giant sequoia (\textit{Sequoiadendron giganteum}) mortality made within the perimeters of the 2020 Castle Fire and of the 2021 Windy and KNP Complex fires \citep{Soderberg2023} (\url{https://www.sciencebase.gov/catalog/item/63d40479d34e06fef150e7b4}). Of the 3,036 giant sequoias of this dataset, 1,015 intersect our large crowns. Each of these crowns received the last known status, alive or dead, of the tree that it intersects. When several trees with different statuses intersected the same crown, the crown was considered dead, as this is the status that has the most influence on the signal seen by the satellite.
The shares of dead trees measured in the field in the dead, breakpoint with $<$80\% NDVI recovery, and breakpoint with $\geq$80\% NDVI recovery and canopy opening (3b) classes (Table \ref{tab_valid_classes}) were then used to estimate the number of large trees very likely dead inside and outside the fire perimeters. The crowns of the subclass under a closed canopy (3a), mostly alive in the field, were considered alive. The uncertainty of these estimates was obtained by resampling the 1,015 field trees with replacement (10,000 bootstrap samples) and is reported as the 2.5 and 97.5 percentiles. Note that these shares were measured on giant sequoias burned in 2020 and 2021, of which 44\% were dead, and that they are applied to the large trees of all species. They are most reliable within fire perimeters, where mortality is high, and likely overestimate mortality outside fire perimeters, where dead trees are rare.

The health classes of all the large trees were then compared with the perimeters of the 2020--2024 wildfires, obtained from the Fire Perimeters database (Firep24\_1) of the CAL FIRE Fire and Resource Assessment Program (FRAP), a statewide compilation developed jointly by CAL FIRE, the U.S. Forest Service, the Bureau of Land Management, the National Park Service and the U.S. Fish and Wildlife Service \citep{CDFFPF2025}. A crown was considered inside a fire perimeter when its polygon intersected one of these perimeters. The analysis was made for the entire Sierra Nevada and for the 97 giant sequoia groves. To test whether disturbed large trees were more frequent within fire perimeters, we compared the proportion of disturbed large trees inside and outside the perimeters with a chi-square test. A tree was considered as disturbed when it was classified as breakpoint with $<$80\% NDVI recovery (4), dead (6), or breakpoint with $\geq$80\% NDVI recovery and canopy opening (3b), the latter because most of its crowns are lost. To verify that the result did not depend on the crown size or on the fire, the same proportion was computed by class of crown area (100--200, 200--400 and $\geq$ 400 m$^2$), the smallest crowns being close to the size of a Sentinel-2 pixel, and for each of the eight largest fires, inside the fire perimeter and in a band of 2 to 5 km around it that excludes all the 2020--2024 fire perimeters. The class proportions were also computed by year of fire, each crown being assigned the year of the first fire that burned it. Finally, for the crowns with a significant breakpoint, we compared the changes of the mean NDVI and of the mean red, green and blue reflectances after the breakpoint, inside and outside the fire perimeters.

\FloatBarrier 
\subsection{Environmental data}

\begin{figure}[ht]
 \centering
 \includegraphics[width=0.95\linewidth]{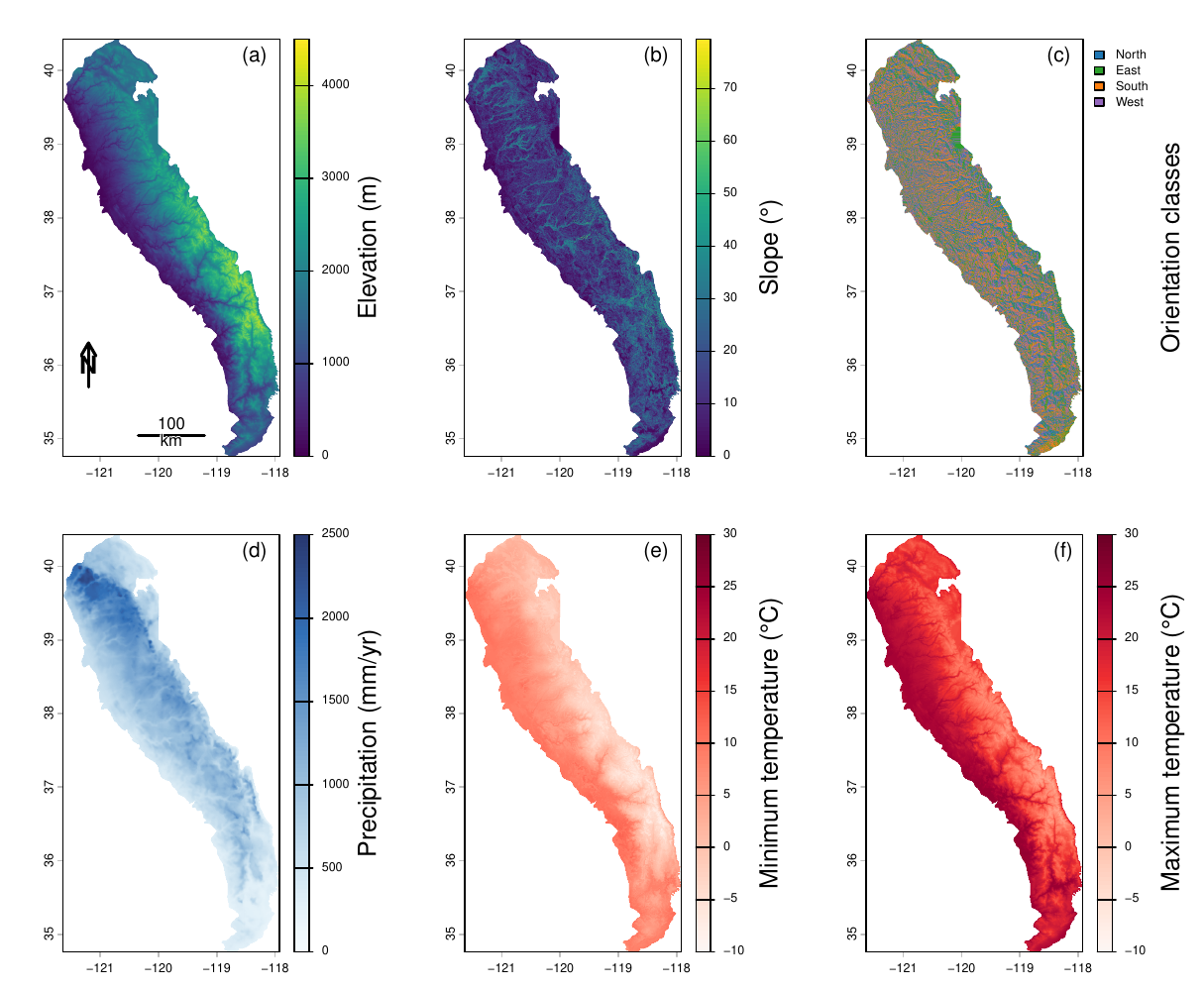}
  \caption{Environmental and climatic variables used in the study to analyse large trees density distribution (a) elevation (m), (b) slope ($^o$), (c) Slope orientation in classes (North, East, South and West), (d) mean annual precipitation (mm.yr$^{-1}$), (e) annual mean of minimum temperatures ($^o$C), and (f) annual mean of maximal temperatures ($^o$C).}
  \label{EnvVar}
  \end{figure}

To describe the association of large trees with elevation, slope and orientation, we used the elevation data of the Shuttle Radar Topography Mission (SRTM) at 1 arc-second ($\sim$30 m) spatial resolution, distributed by OpenTopography \citep{farr2007shuttle,NASA2026}. Slope ($^o$) and orientation were computed from the elevation of the 8 neighbouring pixels (Fig. \ref{EnvVar}a-c).

For climate, we used the annual means of precipitation and of minimum and maximum air temperatures over the period 1981--2025 from the PRISM climate dataset at 800 m spatial resolution (Fig. \ref{EnvVar}d-f). PRISM provides gridded climate data derived from station observations and spatial interpolation across the United States \citep{Daly2008, PRISM2026}.

All environmental variables were resampled to 500 m spatial resolution by averaging, to match the resolution of the large-tree density map.


Each environmental variable was then divided into classes (100 m for elevation, 10$^\circ$ for slope, four orientations, 100 mm for precipitation and 1 $^\circ$C for temperatures). In each class, we compared the number of large trees with the number of random locations, obtained by drawing the same number of locations (6.5 million) uniformly over the Sierra Nevada Floristic Province \citep{wagner2021}. The random locations were bootstrapped 100 times to compute the 95\% confidence interval of the proportion of large trees among large trees and random locations. A proportion of 50\% indicates that large trees occur as expected from a random distribution, a proportion above 50\% that they are over-represented, and a proportion below 50\% that they are under-represented. Because of the very large number of trees, all classes differ statistically from a random distribution, and we therefore interpret the magnitude of the deviation from 50\%.

\FloatBarrier 
\section*{Acknowledgements}
The authors wish to thank the Grantham Foundation and High Tide Foundation for their generous gift to UCLA and support to \url{CTrees.org}. Part of this work was carried out at the Jet Propulsion Laboratory, California Institute of Technology, under a contract with the National Aeronautics and Space Administration (NASA).\added[id=USGS]{ Any use of trade, firm, or product names is for descriptive purposes only and does not imply endorsement by the U.S. Government.}


\section*{Declarations}


\begin{itemize}

\item Conflict of interest/Competing interests: 
The authors declare no competing financial interests.

\item Data availability:  
The Sentinel-2 MSI and NAIP data that support the findings of this study are publicly available from their sources. Processed data, products and codes produced in this study are available from the corresponding author upon request. The crowns drawn by hand on the NAIP images are available from the corresponding author upon request.

\item Author contribution: 
Conceptualization, F.H.W., M.C.M.H., R.D., G.C., A.L., S.S. and S.C.H.; methodology, F.H.W., D.J.D., A.J.D., C.W.W., M.C.M.H., F.S., R.D., G.C., C.M. and S.S.; software, F.H.W., R.D., G.C., F.S. and M.C.M.H.; validation, F.H.W., D.J.D., A.J.D. and A.L.R.; formal analysis, F.H.W.; investigation, F.H.W., R.D., G.C., F.S., M.C.M.H. and S.S.; resources, J.C., N.P., I.S.M. and S.S.; data curation, F.H.W., M.C.M.H., D.J.D., A.J.D., A.L.R., J.C., N.P., I.S.M. and F.S.; writing—original draft preparation, F.H.W.; writing—review and editing, F.H.W., R.D., G.C., M.C.M.H., L.B.S.T., F.S., L.A., C.C., S.C.H., Z.L., C.M., A.M., A.L.R., Y.Y., D.J.D., A.J.D., C.W.W., B.S., D.Z. and S.S.; visualization, F.H.W. and M.C.M.H.; supervision, S.S.; project administration, C.W.W., S.C.H. and S.S.; funding acquisition, C.W.W., S.C.H. and S.S. All authors reviewed and approved the final manuscript.

\end{itemize}


===============

\clearpage
\bibliography{references_sequoias}

@Article{Ronneberger2015,
  author        = {Olaf Ronneberger and Philipp Fischer and Thomas Brox},
  title         = {U-Net: Convolutional Networks for Biomedical Image Segmentation},
  journal       = {CoRR},
  year          = {2015},
  volume        = {abs/1505.04597},
  archiveprefix = {arXiv},
  bibsource     = {dblp computer science bibliography, https://dblp.org},
  eprint        = {1505.04597},
  url           = {http://arxiv.org/abs/1505.04597},
}

@Misc{chollet2015keras,
  author       = {Chollet, Fran\c{c}ois and others},
  title        = {Keras},
  howpublished = {\url{https://keras.io}},
  year         = {2015},
}

@Article{Dice1945,
  author    = {Dice, Lee R},
  title     = {Measures of the amount of ecologic association between species},
  journal   = {Ecology},
  year      = {1945},
  volume    = {26},
  number    = {3},
  pages     = {297--302},
  publisher = {Wiley Online Library},
}

@Article{farr2007shuttle,
  author    = {Farr, Tom G and Rosen, Paul A and Caro, Edward and Crippen, Robert and Duren, Riley and Hensley, Scott and Kobrick, Michael and Paller, Mimi and Rodriguez, Ernesto and Roth, Ladislav and others},
  title     = {The shuttle radar topography mission},
  journal   = {Reviews of geophysics},
  year      = {2007},
  volume    = {45},
  number    = {2},
  publisher = {Wiley Online Library},
}

@Misc{he2017mask,
  author    = {He, Kaiming and Gkioxari, Georgia and Doll{\'a}r, Piotr and Girshick, Ross},
  title     = {Mask r-cnn},
  year      = {2017},
  booktitle = {Proceedings of the IEEE international conference on computer vision},
  pages     = {2961--2969},
}

@Article{Pebesma2018,
  author  = {Edzer Pebesma},
  title   = {{Simple Features for R: Standardized Support for Spatial Vector Data}},
  journal = {{The R Journal}},
  year    = {2018},
  volume  = {10},
  number  = {1},
  pages   = {439--446},
  doi     = {10.32614/RJ-2018-009},
  url     = {https://doi.org/10.32614/RJ-2018-009},
}

@Manual{AllaireChollet2019,
  title  = {keras: R Interface to 'Keras'},
  author = {JJ Allaire and François Chollet},
  year   = {2019},
  note   = {R package version 2.2.4.1.9001},
  url    = {https://keras.rstudio.com},
}

@Book{chollet2018deep,
  title     = {Deep Learning with R},
  publisher = {Manning Publications Co.},
  year      = {2018},
  author    = {Chollet, Francois and Allaire, JJ},
}

@Article{wagner2021,
  author    = {Wagner, Fabien H},
  title     = {The flowering of Atlantic Forest Pleroma trees},
  journal   = {Scientific reports},
  year      = {2021},
  volume    = {11},
  number    = {1},
  pages     = {1--20},
  publisher = {Nature Publishing Group},
}

@Article{Roussel2020,
  author  = {Jean-Romain Roussel and David Auty and Nicholas C. Coops and Piotr Tompalski and Tristan R.H. Goodbody and Andrew Sánchez Meador and Jean-François Bourdon and Florian {de Boissieu} and Alexis Achim},
  title   = {lidR: An R package for analysis of Airborne Laser Scanning (ALS) data},
  journal = {Remote Sensing of Environment},
  year    = {2020},
  volume  = {251},
  pages   = {112061},
  issn    = {0034-4257},
  doi     = {10.1016/j.rse.2020.112061},
  url     = {https://www.sciencedirect.com/science/article/pii/S0034425720304314},
}

@Article{mcintyre2015,
  author    = {McIntyre, Patrick J and Thorne, James H and Dolanc, Christopher R and Flint, Alan L and Flint, Lorraine E and Kelly, Maggi and Ackerly, David D},
  title     = {Twentieth-century shifts in forest structure in California: Denser forests, smaller trees, and increased dominance of oaks},
  journal   = {Proceedings of the National Academy of Sciences},
  year      = {2015},
  volume    = {112},
  number    = {5},
  pages     = {1458--1463},
  publisher = {National Acad Sciences},
}

@Article{lutz2018,
  author    = {Lutz, James A and Furniss, Tucker J and Johnson, Daniel J and Davies, Stuart J and Allen, David and Alonso, Alfonso and Anderson-Teixeira, Kristina J and Andrade, Ana and Baltzer, Jennifer and Becker, Kendall ML and others},
  title     = {Global importance of large-diameter trees},
  journal   = {Global Ecology and Biogeography},
  year      = {2018},
  volume    = {27},
  number    = {7},
  pages     = {849--864},
  publisher = {Wiley Online Library},
}

@Article{enquist2020,
  author    = {Enquist, Brian J and Abraham, Andrew J and Harfoot, Michael BJ and Malhi, Yadvinder and Doughty, Christopher E},
  title     = {The megabiota are disproportionately important for biosphere functioning},
  journal   = {Nature Communications},
  year      = {2020},
  volume    = {11},
  number    = {1},
  pages     = {699},
  publisher = {Nature Publishing Group UK London},
}

@Article{Braga2020,
  author         = {G. Braga, José R. and Peripato, Vinícius and Dalagnol, Ricardo and P. Ferreira, Matheus and Tarabalka, Yuliya and O. C. Aragão, Luiz E. and F. de Campos Velho, Haroldo and Shiguemori, Elcio H. and Wagner, Fabien H.},
  title          = {Tree Crown Delineation Algorithm Based on a Convolutional Neural Network},
  journal        = {Remote Sensing},
  year           = {2020},
  volume         = {12},
  number         = {8},
  issn           = {2072-4292},
  article-number = {1288},
  doi            = {10.3390/rs12081288},
  url            = {https://www.mdpi.com/2072-4292/12/8/1288},
}

@Article{Williams2016,
  author    = {Williams, Christopher A and Gu, Huan and MacLean, Richard and Masek, Jeffrey G and Collatz, G James},
  title     = {Disturbance and the carbon balance of US forests: A quantitative review of impacts from harvests, fires, insects, and droughts},
  journal   = {Global and Planetary Change},
  year      = {2016},
  volume    = {143},
  pages     = {66--80},
  publisher = {Elsevier},
}

@Article{wang2022,
  author    = {Wang, Jonathan A and Randerson, James T and Goulden, Michael L and Knight, Clarke A and Battles, John J},
  title     = {Losses of tree cover in California driven by increasing fire disturbance and climate stress},
  journal   = {AGU Advances},
  year      = {2022},
  volume    = {3},
  number    = {4},
  pages     = {e2021AV000654},
  publisher = {Wiley Online Library},
}

@Article{goulden2019,
  author    = {Goulden, Michael L and Bales, Roger C},
  title     = {California forest die-off linked to multi-year deep soil drying in 2012--2015 drought},
  journal   = {Nature Geoscience},
  year      = {2019},
  volume    = {12},
  number    = {8},
  pages     = {632--637},
  publisher = {Nature Publishing Group UK London},
}

@Article{Barbour1988,
  author    = {Barbour, Michael G and Minnich, Richard A},
  journal   = {North American terrestrial vegetation},
  title     = {California upland forests},
  year      = {1988},
  pages     = {132--164},
  publisher = {Cambridge University Press},
}

@Article{VanWagtendonk2018,
  author  = {Van Wagtendonk, Jan W and Fites-Kaufman, Jo Ann and Safford, Hugh D and North, Malcolm P and Collins, Brandon},
  journal = {Fire in California's ecosystems},
  title   = {Sierra Nevada bioregion},
  year    = {2018},
  pages   = {249--279},
}

@Article{Griffin1976,
  author  = {Griffin, James R and Critchfield, William B},
  journal = {USDA FOREST SERVICE RESEARCH PAPER PSW- 82 /1972},
  title   = {The distribution of forest trees in California.},
  year    = {1976},
}

@Article{Simoes2021,
  author  = {Rolf Simoes and Gilberto Camara and Gilberto Queiroz and Felipe Souza and Pedro Andrade and Lorena Santos and Alexandre Carvalho and Karine Ferreira},
  journal = {Remote Sensing},
  title   = {Satellite Image Time Series Analysis for Big Earth Observation Data},
  year    = {2021},
  number  = {13},
  pages   = {2428},
  volume  = {13},
  doi     = {10.3390/rs13132428},
}

@Article{Wagner2020UnetID,
  author         = {Wagner, Fabien H. and Dalagnol, Ricardo and Tarabalka, Yuliya and Segantine, Tassiana Y. F. and Thomé, Rogério and Hirye, Mayumi C. M.},
  journal        = {Remote Sensing},
  title          = {U-Net-Id, an Instance Segmentation Model for Building Extraction from Satellite Images—Case Study in the Joanópolis City, Brazil},
  year           = {2020},
  issn           = {2072-4292},
  number         = {10},
  volume         = {12},
  article-number = {1544},
  doi            = {10.3390/rs12101544},
  url            = {https://www.mdpi.com/2072-4292/12/10/1544},
}

@Article{Verbesselt2010,
  author  = {Jan Verbesselt and Rob Hyndman and Glenn Newnham and Darius Culvenor},
  journal = {Remote Sensing of Environment},
  title   = {Detecting Trend and Seasonal Changes in Satellite Image Time Series},
  year    = {2010},
  number  = {1},
  pages   = {106--115},
  volume  = {114},
  doi     = {10.1016/j.rse.2009.08.014},
}

@Article{Verbesselt2010a,
  author  = {Jan Verbesselt and Rob Hyndman and Achim Zeileis and Darius Culvenor},
  journal = {Remote Sensing of Environment},
  title   = {Phenological Change Detection while Accounting for Abrupt and Gradual Trends in Satellite Image Time Series},
  year    = {2010},
  number  = {12},
  pages   = {2970--2980},
  volume  = {114},
  doi     = {10.1016/j.rse.2010.08.003},
}

@Article{Wagner2024,
  author    = {Wagner, Fabien H and Roberts, Sophia and Ritz, Alison L and Carter, Griffin and Dalagnol, Ricardo and Favrichon, Samuel and Hirye, Mayumi CM and Brandt, Martin and Ciais, Philippe and Saatchi, Sassan},
  journal   = {Remote Sensing of Environment},
  title     = {Sub-meter tree height mapping of California using aerial images and LiDAR-informed U-Net model},
  year      = {2024},
  pages     = {114099},
  volume    = {305},
  publisher = {Elsevier},
}

@TechReport{NASA2026,
  author      = {NASA},
  institution = {NASA},
  title       = {Shuttle Radar Topography Mission (SRTM)(2013). Shuttle Radar Topography Mission (SRTM) Global. Distributed by OpenTopography. https://doi.org/10.5069/G9445JDF. Accessed 2026-05-20},
  year        = {2026},
}

@Article{Daly2008,
  author    = {Daly, Christopher and Halbleib, Michael and Smith, Joseph I and Gibson, Wayne P and Doggett, Matthew K and Taylor, George H and Curtis, Jan and Pasteris, Phillip P},
  journal   = {International Journal of Climatology: a Journal of the Royal Meteorological Society},
  title     = {Physiographically sensitive mapping of climatological temperature and precipitation across the conterminous United States},
  year      = {2008},
  number    = {15},
  pages     = {2031--2064},
  volume    = {28},
  publisher = {Wiley Online Library},
}

@Misc{PRISM2026,
  author       = {{PRISM Climate Group}},
  howpublished = {\url{https://prism.oregonstate.edu}},
  note         = {Accessed: 2025-12-16},
  title        = {PRISM Climate Data},
  school       = {Oregon State University},
}

@Article{Soderberg2023,
  author  = {Soderberg, David N and Das, Adrian J},
  journal = {US Geological Survey (USGS) Data Release},
  title   = {Assessment of Giant Sequoia Mortality and Regeneration within Burned Groves in Sequoia and Kings Canyon National Parks (ver. 4.0, February 2026)},
  year    = {2023},
  pages   = {805},
}

@Misc{CDFFPF2025,
  author    = {{California Department of Forestry and Fire Protection (CAL FIRE)}},
  note      = {Version Firep24\_1; statewide wildfire perimeter database},
  title     = {Fire Perimeters (Firep24\_1 Geodatabase)},
  year      = {2025},
  address   = {Sacramento, California, USA},
  publisher = {Fire and Resource Assessment Program (FRAP)},
  url       = {https://www.fire.ca.gov/what-we-do/fire-resource-assessment-program/fire-perimeters},
  urldate   = {2026-06-01},
}

@Article{Favrichon2025,
  author   = {Favrichon, Samuel and Lee, Jake and Yang, Yan and Dalagnol, Ricardo and Wagner, Fabien and Sagang, Le Bienfaiteur and Saatchi, Sassan},
  journal  = {Frontiers in Remote Sensing},
  title    = {Monitoring changes of forest height in California},
  year     = {2025},
  issn     = {2673-6187},
  volume   = {Volume 5 - 2024},
  doi      = {10.3389/frsen.2024.1459524},
  url      = {https://www.frontiersin.org/journals/remote-sensing/articles/10.3389/frsen.2024.1459524},
}

@Article{Burge2021,
  author    = {Burge, Dylan O and Thorne, James H and Harrison, Susan P and O'Brien, Bart C and Rebman, Jon P and Shevock, James R and Alverson, Edward R and Hardison, Linda K and Delgadillo-Rodr{\'\i}guez, Jos{\'e} and Junak, Steven A and others},
  journal   = {V1},
  title     = {Data from: Plant diversity and endemism in the California Floristic Province},
  year      = {2021},
  doi       = {https://doi.org/10.5683/SP2/7DLO5Q},
  publisher = {Borealis},
  url       = {https://borealisdata.ca/file.xhtml?fileId=153128&version=1.0},
}

@Article{Carter2024,
  author    = {Carter, Griffin and Wagner, Fabien H and Dalagnol, Ricardo and Roberts, Sophia and Ritz, Alison L and Saatchi, Sassan},
  journal   = {Frontiers in Remote Sensing},
  title     = {Detection of forest disturbance across California using deep-learning on PlanetScope imagery},
  year      = {2024},
  pages     = {1409400},
  volume    = {5},
  publisher = {Frontiers Media SA},
}

@Article{Weinstein2019,
  author    = {Weinstein, Ben G and Marconi, Sergio and Bohlman, Stephanie and Zare, Alina and White, Ethan},
  journal   = {Remote Sensing},
  title     = {Individual tree-crown detection in RGB imagery using semi-supervised deep learning neural networks},
  year      = {2019},
  number    = {11},
  pages     = {1309},
  volume    = {11},
  publisher = {MDPI},
}

@Article{Cheng2024,
  author    = {Cheng, Yan and Oehmcke, Stefan and Brandt, Martin and Rosenthal, Lisa and Das, Adrian and Vrieling, Anton and Saatchi, Sassan and Wagner, Fabien and Mugabowindekwe, Maurice and Verbruggen, Wim and others},
  journal   = {Nature communications},
  title     = {Scattered tree death contributes to substantial forest loss in California},
  year      = {2024},
  number    = {1},
  pages     = {641},
  volume    = {15},
  publisher = {Nature Publishing Group UK London},
}

@Article{Shive2026,
  author    = {Shive, Kristen L and Baker, Brianna and Soderberg, David and Hardlund, Linnea J and Meyer, Marc D and Nagelson, P Bryant and Bisbing, Sarah M and Das, Adrian J and Stephenson, Nathan L},
  journal   = {Fire Ecology},
  title     = {The state of the giant sequoias: losses, risks, and opportunities},
  year      = {2026},
  number    = {1},
  pages     = {30},
  volume    = {22},
  publisher = {Springer},
}

@Book{Stephenson1996,
  author    = {Stephenson, N. L.},
  publisher = {Centers for Water and Wildland Resources, University of California},
  title     = {Ecology and Management of Giant Sequoia Groves},
  year      = {1996},
  address   = {Davis, CA},
  booktitle = {Sierra Nevada Ecosystem Project: Final Report to Congress, Volume II. Assessments and Scientific Basis for Management Options},
  pages     = {1431--1465},
}

@Article{Winsemius2024,
  author    = {Winsemius, Sara and Babcock, Chad and Kane, Van R and Bormann, Kat J and Safford, Hugh D and Jin, Yufang},
  journal   = {Carbon Balance and Management},
  title     = {Improved aboveground biomass estimation and regional assessment with aerial lidar in California’s subalpine forests},
  year      = {2024},
  number    = {1},
  pages     = {41},
  volume    = {19},
  publisher = {Springer},
}

@Book{Pebesma2023,
  author    = {Edzer Pebesma and Roger Bivand},
  publisher = {{Chapman and Hall/CRC}},
  title     = {{Spatial Data Science: With applications in R}},
  year      = {2023},
  doi       = {10.1201/9780429459016},
  url       = {https://r-spatial.org/book/},
}

@Manual{Hijmans2026,
  title  = {terra: Spatial Data Analysis},
  author = {Robert J. Hijmans},
  note   = {R package version 1.8-93},
  year   = {2026},
  url    = {https://CRAN.R-project.org/package=terra},
}

@Article{Stephenson2014,
  author    = {Stephenson, Nathan L and Das, AJ and Condit, R and Russo, SE and Baker, PJ and Beckman, Noelle G and Coomes, DA and Lines, ER and Morris, WK and R{\"u}ger, Nadja and others},
  journal   = {Nature},
  title     = {Rate of tree carbon accumulation increases continuously with tree size},
  year      = {2014},
  number    = {7490},
  pages     = {90--93},
  volume    = {507},
  publisher = {Nature Publishing Group UK London},
}

@Article{Chen2021,
  author    = {Chen, Na and Tsendbazar, Nandin-Erdene and Hamunyela, Eliakim and Verbesselt, Jan and Herold, Martin},
  journal   = {International Journal of Applied Earth Observation and Geoinformation},
  title     = {Sub-annual tropical forest disturbance monitoring using harmonized Landsat and Sentinel-2 data},
  year      = {2021},
  pages     = {102386},
  volume    = {102},
  publisher = {Elsevier},
}

@Article{Schiller2026,
  author    = {Christopher Schiller and Fabian Ewald Fassnacht},
  journal   = {European Journal of Remote Sensing},
  title     = {Comparative study of near real-time monitoring algorithms for early detection of bark beetle infestations in Germany with Sentinel-2},
  year      = {2026},
  number    = {1},
  pages     = {2662660},
  volume    = {59},
  doi       = {10.1080/22797254.2026.2662660},
  eprint    = {https://doi.org/10.1080/22797254.2026.2662660},
  publisher = {Taylor \& Francis},
  url       = {https://doi.org/10.1080/22797254.2026.2662660},
}

@Article{Kennedy2021,
  author    = {Kennedy, Maureen C and Bart, Ryan R and Tague, Christina L and Choate, Janet S},
  journal   = {Ecosphere},
  title     = {Does hot and dry equal more wildfire? Contrasting short-and long-term climate effects on fire in the Sierra Nevada, CA},
  year      = {2021},
  number    = {7},
  pages     = {e03657},
  volume    = {12},
  publisher = {Wiley Online Library},
}

@Article{Keeley2021,
  author    = {Keeley, Jon E and Syphard, Alexandra D},
  journal   = {Fire Ecology},
  title     = {Large California wildfires: 2020 fires in historical context},
  year      = {2021},
  number    = {1},
  pages     = {22},
  volume    = {17},
  publisher = {Springer},
}

@Article{Safford2022,
  author    = {Safford, Hugh D and Paulson, Alison K and Steel, Zachary L and Young, Derek JN and Wayman, Rebecca B},
  journal   = {Global Ecology and Biogeography},
  title     = {The 2020 California fire season: A year like no other, a return to the past or a harbinger of the future?},
  year      = {2022},
  number    = {10},
  pages     = {2005--2025},
  volume    = {31},
  publisher = {Wiley Online Library},
}

@Article{Miller2009,
  author    = {Miller, Jay D and Knapp, Eric E and Key, Carl H and Skinner, Carl N and Isbell, Clint J and Creasy, R Max and Sherlock, Joseph W},
  journal   = {Remote sensing of environment},
  title     = {Calibration and validation of the relative differenced Normalized Burn Ratio (RdNBR) to three measures of fire severity in the Sierra Nevada and Klamath Mountains, California, USA},
  year      = {2009},
  number    = {3},
  pages     = {645--656},
  volume    = {113},
  publisher = {Elsevier},
}

@Article{Miller2007,
  author    = {Miller, Jay D and Thode, Andrea E},
  journal   = {Remote sensing of Environment},
  title     = {Quantifying burn severity in a heterogeneous landscape with a relative version of the delta Normalized Burn Ratio (dNBR)},
  year      = {2007},
  number    = {1},
  pages     = {66--80},
  volume    = {109},
  publisher = {Elsevier},
}

@Article{Parks2018,
  author         = {Parks, Sean A. and Holsinger, Lisa M. and Voss, Morgan A. and Loehman, Rachel A. and Robinson, Nathaniel P.},
  journal        = {Remote Sensing},
  title          = {Mean Composite Fire Severity Metrics Computed with Google Earth Engine Offer Improved Accuracy and Expanded Mapping Potential},
  year           = {2018},
  issn           = {2072-4292},
  number         = {6},
  volume         = {10},
  article-number = {879},
  doi            = {10.3390/rs10060879},
  url            = {https://www.mdpi.com/2072-4292/10/6/879},
}

@Article{Ritz2026,
  author    = {Alison L. Ritz and Randolph H. Wynne and Valerie A. Thomas and Fabien H. Wagner and P. Corey Green and Todd A. Schroeder and Sassan Saatchi},
  journal   = {International Journal of Remote Sensing},
  title     = {Applying a convolutional neural network (CNN) to Virginia’s forests: how forest type and age can impact individual tree segmentation},
  year      = {2026},
  number    = {1},
  pages     = {218--245},
  volume    = {47},
  doi       = {10.1080/01431161.2025.2598075},
  eprint    = {https://doi.org/10.1080/01431161.2025.2598075},
  publisher = {Taylor \& Francis},
  url       = {https://doi.org/10.1080/01431161.2025.2598075},
}

@Article{Kane2023,
  author    = {Kane, Van R and Bartl-Geller, Bryce N and Cova, Gina R and Chamberlain, Caden P and van Wagtendonk, Liz and North, Malcolm P},
  journal   = {Forest Ecology and Management},
  title     = {Where are the large trees? A census of Sierra Nevada large trees to determine their frequency and spatial distribution across three large landscapes},
  year      = {2023},
  pages     = {121351},
  volume    = {546},
  publisher = {Elsevier},
}

@Article{Williams2023,
  author    = {Williams, JN and Safford, HD and Enstice, N and Steel, ZL and Paulson, AK},
  journal   = {Ecosphere},
  title     = {High-severity burned area and proportion exceed historic conditions in Sierra Nevada, California, and adjacent ranges},
  year      = {2023},
  number    = {1},
  pages     = {e4397},
  volume    = {14},
  publisher = {Wiley Online Library},
}

@Article{Blomdahl2019,
  author    = {Blomdahl, Erika M and Kolden, Crystal A and Meddens, Arjan JH and Lutz, James A},
  journal   = {Forest Ecology and Management},
  title     = {The importance of small fire refugia in the central Sierra Nevada, California, USA},
  year      = {2019},
  pages     = {1041--1052},
  volume    = {432},
  publisher = {Elsevier},
}

@Article{Paudel2023,
  author    = {Paudel, Asha and Markwith, Scott H},
  journal   = {Journal of Vegetation Science},
  title     = {From the severity patch to the landscape: Wildfire and spatial heterogeneity in northern Sierra Nevada conifer forests},
  year      = {2023},
  number    = {5},
  pages     = {e13207},
  volume    = {34},
  publisher = {Wiley Online Library},
}

@Article{Jeronimo2020,
  author    = {Jeronimo, Sean MA and Lutz, James A and R. Kane, Van and Larson, Andrew J and Franklin, Jerry F},
  journal   = {Landscape Ecology},
  title     = {Burn weather and three-dimensional fuel structure determine post-fire tree mortality},
  year      = {2020},
  number    = {4},
  pages     = {859--878},
  volume    = {35},
  publisher = {Springer},
}

@Article{Reilly2023,
  author    = {Reilly, Matthew J and Zuspan, Aaron and Yang, Zhiqiang},
  journal   = {Fire Ecology},
  title     = {Characterizing post-fire delayed tree mortality with remote sensing: sizing up the elephant in the room},
  year      = {2023},
  number    = {1},
  pages     = {64},
  volume    = {19},
  publisher = {Springer},
}

@Article{Hanson2009,
  author    = {Hanson, Chad T and North, Malcolm P},
  journal   = {International Journal of Wildland Fire},
  title     = {Post-fire survival and flushing in three Sierra Nevada conifers with high initial crown scorch},
  year      = {2009},
  number    = {7},
  pages     = {857--864},
  volume    = {18},
  publisher = {CSIRO Publishing},
}

@Article{Dixon2023,
  author    = {Dixon, Dan J and Zhu, Yunzhe and Brown, Christopher F and Jin, Yufang},
  journal   = {Remote Sensing of Environment},
  title     = {Satellite detection of canopy-scale tree mortality and survival from California wildfires with spatio-temporal deep learning},
  year      = {2023},
  pages     = {113842},
  volume    = {298},
  publisher = {Elsevier},
}

@Article{Hung2026,
  author    = {Hung, Mitchell J and Williams, A Park},
  journal   = {Proceedings of the National Academy of Sciences},
  title     = {High-severity fire now dominant in California forests},
  year      = {2026},
  number    = {26},
  pages     = {e2532829123},
  volume    = {123},
  publisher = {National Academy of Sciences},
}

@Article{Sillett2019,
  author    = {Sillett, Stephen C and Van Pelt, Robert and Carroll, Allyson L and Campbell-Spickler, Jim and Antoine, Marie E},
  journal   = {Forest Ecology and Management},
  title     = {Structure and dynamics of forests dominated by Sequoiadendron giganteum},
  year      = {2019},
  pages     = {218--239},
  volume    = {448},
  publisher = {Elsevier},
}

@Article{Dixon2026,
  author    = {Dixon, Dan J and Das, Adrian J and Dong, Xiaoli and Latimer, Andrew M and Soderberg, David N and Stephenson, Nathan L and Caprio, Anthony C and Jin, Yufang},
  journal   = {Nature Communications},
  title     = {Previous prescribed burns saved thousands of ancient sequoias during historically unprecedented wildfires},
  year      = {2026},
  publisher = {Nature Publishing Group UK London},
}

@Article{Hardlund2026,
  author    = {Hardlund, Linnea J and Collins, Brandon M and Bernal, Alexis A and Stephens, Scott L and York, Robert A and Shive, Kristen L},
  journal   = {Fire Ecology},
  title     = {How big is big enough? Exploring drivers of fire-induced giant sequoia mortality from the individual to the stand level},
  year      = {2026},
  publisher = {Springer},
}

@misc{NPS1997,
  author       = {{National Park Service}},
  title        = {The {General Sherman Tree}},
  year         = {1997},
  howpublished = {Sequoia and Kings Canyon National Parks, U.S. National Park Service},
  url          = {https://www.nps.gov/seki/learn/nature/sherman.htm},
  note         = {Accessed 5 November 2026}
}

@Article{Dalponte2016,
  author  = {Dalponte, Michele and Coomes, David A.},
  title   = {Tree-centric mapping of forest carbon density from airborne laser scanning and hyperspectral data},
  journal = {Methods in Ecology and Evolution},
  year    = {2016},
  volume  = {7},
  pages   = {1236--1245},
  doi     = {10.1111/2041-210X.12575}
}

@Article{Piponiot2022,
  author  = {Piponiot, Camille and Anderson-Teixeira, Kristina J. and Davies, Stuart J. and Allen, David and Bourg, Norman A. and Burslem, David F. R. P. and C{\'a}rdenas, Dairon and Chang-Yang, Chia-Hao and Chuyong, George and Cordell, Susan and Dattaraja, Handanakere Shivaramaiah and Duque, {\'A}lvaro and Ediriweera, Sisira and Ewango, Corneille and Ezedin, Zacky and Filip, Jonah and Giardina, Christian P. and Howe, Robert and Hsieh, Chang-Fu and Hubbell, Stephen P. and Inman-Narahari, Faith M. and Itoh, Akira and Jan{\'i}k, David and Kenfack, David and Kr{\'a}l, Kamil and Lutz, James A. and Makana, Jean-Remy and McMahon, Sean M. and McShea, William and Mi, Xiangcheng and Bt. Mohamad, Mohizah and Novotn{\'y}, Vojt{\v{e}}ch and O'Brien, Michael J. and Ostertag, Rebecca and Parker, Geoffrey and P{\'e}rez, Rolando and Ren, Haibao and Reynolds, Glen and Md Sabri, Mohamad Danial and Sack, Lawren and Shringi, Ankur and Su, Sheng-Hsin and Sukumar, Raman and Sun, I-Fang and Suresh, Hebbalalu S. and Thomas, Duncan W. and Thompson, Jill and Uriarte, Maria and Vandermeer, John and Wang, Yunquan and Ware, Ian M. and Weiblen, George D. and Whitfeld, Timothy J. S. and Wolf, Amy and Yao, Tze Leong and Yu, Mingjian and Yuan, Zuoqiang and Zimmerman, Jess K. and Zuleta, Daniel and Muller-Landau, Helene C.},
  title   = {Distribution of biomass dynamics in relation to tree size in forests across the world},
  journal = {New Phytologist},
  year    = {2022},
  volume  = {234},
  number  = {5},
  pages   = {1664--1677},
  doi     = {10.1111/nph.17995},
}

@misc{USGSNAIP2026,
  author = {{U.S. Geological Survey}},
  title = {{USGS EROS Archive -- Aerial Photography -- National Agriculture Imagery Program (NAIP)}},
  year = {2026},
  doi = {10.5066/F7QN651G},
  howpublished = {\url{https://www.usgs.gov/centers/eros/science/usgs-eros-archive-aerial-photography-national-agriculture-imagery-program-naip}},
  note = {Accessed: 19 November 2026}
}

@misc{ESAS2L2A2026,
  author = {{European Space Agency}},
  title = {{Copernicus Sentinel-2 Collection 1 Level-2A: product characteristics}},
  year = {2026},
  howpublished = {\url{https://sentinels.copernicus.eu/-/collection-1-level-2a}},
  note = {Accessed: 19 November 2026}
}
\clearpage
\begin{appendices}
\section{Change of the NDVI and red reflectance by class and field status}\label{secA_bands}

\begin{table}[ht]
\caption{Change of the mean NDVI and of the mean red reflectance of the crown after the breakpoint (mean $\pm$ SD) for the giant sequoias alive and dead on the ground, by class (field validation, Table \ref{tab_valid_classes}). The permanent alteration (5) class, with a single dead tree, is not shown.}
\centering
\footnotesize
\setlength{\tabcolsep}{4pt}
\begin{tabular}{lrrrrr}
\hline
Class & n (alive / dead) & \multicolumn{2}{c}{NDVI change} & \multicolumn{2}{c}{Red change} \\
 & & alive & dead & alive & dead \\
\hline
alive with breakpoint (2) & 142 / 7 & $-$0.086 $\pm$ 0.042 & $-$0.124 $\pm$ 0.040 & $+$0.014 $\pm$ 0.006 & $+$0.018 $\pm$ 0.008 \\
\quad under a closed canopy (3a) & 28 / 7 & $-$0.137 $\pm$ 0.042 & $-$0.200 $\pm$ 0.041 & $+$0.017 $\pm$ 0.004 & $+$0.018 $\pm$ 0.004 \\
\quad with canopy opening (3b) & 29 / 28 & $-$0.183 $\pm$ 0.053 & $-$0.317 $\pm$ 0.075 & $+$0.025 $\pm$ 0.007 & $+$0.032 $\pm$ 0.007 \\
breakpoint with $<$80\% NDVI recovery (4) & 142 / 365 & $-$0.272 $\pm$ 0.096 & $-$0.428 $\pm$ 0.079 & $+$0.031 $\pm$ 0.009 & $+$0.041 $\pm$ 0.009 \\
dead (6) & 14 / 31 & $-$0.345 $\pm$ 0.099 & $-$0.439 $\pm$ 0.097 & $+$0.034 $\pm$ 0.009 & $+$0.042 $\pm$ 0.009 \\
\hline
\end{tabular}
\label{tab_bands_field_status}
\end{table}

\clearpage 
\section{Disturbed large trees by crown area, fire and year of fire}\label{secA_tables}
\setcounter{table}{0}\setcounter{figure}{0}

\begin{table}[ht]
\caption{Proportion of disturbed large crowns (breakpoint with $<$80\% NDVI recovery, dead, or breakpoint with $\geq$80\% NDVI recovery and canopy opening (3b)) by crown area, inside and outside the 2020--2024 fire perimeters. N is the number of large crowns.}
\centering
\begin{tabular}{lrrrr}
\hline
Crown area (m$^2$) & \multicolumn{2}{c}{Inside fire perimeters} & \multicolumn{2}{c}{Outside fire perimeters} \\
 & N & Disturbed (\%) & N & Disturbed (\%) \\
\hline
100--200 & 1,488,951 & 61.6 & 4,830,251 & 2.3 \\
200--400 & 51,903 & 58.0 & 143,799 & 2.0 \\
$\geq$400 & 170 & 47.1 & 631 & 3.3 \\
\hline
\end{tabular}
\label{tab_crown_size}
\end{table}

\begin{table}[ht]
\caption{Proportion of disturbed large crowns (breakpoint with $<$80\% NDVI recovery, dead, or breakpoint with $\geq$80\% NDVI recovery and canopy opening (3b)) inside the eight largest fires of 2020--2024 and in a 2--5 km band around each fire that excludes all the 2020--2024 fire perimeters. N is the number of large crowns.}
\centering
\begin{tabular}{lrrrrr}
\hline
Fire & Year & \multicolumn{2}{c}{Inside the fire perimeter} & \multicolumn{2}{c}{2--5 km band outside} \\
 & & N & Disturbed (\%) & N & Disturbed (\%) \\
\hline
Castle & 2020 & 90,411 & 52.6 & 33,187 & 2.6 \\
Creek & 2020 & 193,701 & 67.9 & 72,017 & 4.3 \\
North Complex & 2020 & 394,005 & 68.2 & 111,723 & 1.5 \\
Caldor & 2021 & 240,339 & 65.9 & 95,859 & 2.2 \\
Dixie & 2021 & 240,613 & 67.3 & 53,161 & 1.8 \\
KNP Complex & 2021 & 71,954 & 46.5 & 44,903 & 3.0 \\
Sugar & 2021 & 22,083 & 74.9 & 2,247 & 3.8 \\
Windy & 2021 & 46,685 & 64.3 & 12,943 & 3.5 \\
\hline
\end{tabular}
\label{tab_largest_fires}
\end{table}

\begin{table}[ht]
\caption{Classes of the large crowns (\% of the crowns) by year of the first fire that burned them, and outside the fire perimeters. N is the number of large crowns. The crowns burned in 2023 and 2024 have less than two years of Sentinel-2 observations after the fire, and the subclasses 3a and 3b are not informative for 2024.}
\centering
\footnotesize
\setlength{\tabcolsep}{3pt}
\begin{tabular}{lrrrrrr}
\hline
Class & 2020 & 2021 & 2022 & 2023 & 2024 & Outside fire perimeters \\
\hline
N & 704,140 & 650,879 & 125,840 & 10,215 & 49,950 & 4,974,681 \\
\hline
alive (1) & 13.8 & 15.5 & 33.7 & 60.5 & 50.0 & 70.1 \\
alive with breakpoint (2) & 10.0 & 10.7 & 6.7 & 8.4 & 4.6 & 13.3 \\
breakpoint with $\geq$80\% NDVI recovery (3) & 14.5 & 5.9 & 5.3 & 1.9 & 10.3 & 2.1 \\
\quad under a closed canopy (3a) & 3.1 & 2.5 & 2.2 & 1.7 & 9.7 & 1.7 \\
\quad with canopy opening (3b) & 11.4 & 3.4 & 3.1 & 0.2 & 0.6 & 0.3 \\
breakpoint with $<$80\% NDVI recovery (4) & 45.3 & 49.9 & 38.1 & 4.3 & 1.0 & 1.1 \\
permanent alteration (5) & 1.4 & 1.4 & 1.4 & 0.8 & 0.4 & 0.7 \\
dead (6) & 8.8 & 10.8 & 10.9 & 4.6 & 0.7 & 0.8 \\
already dead or outside the reference colour range (0) & 6.1 & 5.7 & 3.8 & 19.4 & 33.0 & 11.8 \\
not enough data (255) & 0.1 & 0.0 & 0.0 & 0.2 & 0.1 & 0.1 \\
\hline
disturbed (4 + 6 + 3b) & 65.5 & 64.2 & 52.0 & 9.1 & 2.2 & 2.2 \\
\hline
\end{tabular}
\label{tab_fire_year}
\end{table}

\clearpage 
\section{Generation of synthetic canopy height models}\label{secA_synthetic}
\setcounter{table}{0}\setcounter{figure}{0}

We generate a fully synthetic canopy height model (CHM) by simulating 3D individual tree crowns and merging them into LiDAR--like canopy height model. The goal is to obtain realistic CHM with fully controlled samples, with known tree location, size, and geometry. Simulations are done in R using \texttt{terra}, \texttt{sf}, and \texttt{tensorflow}. Ground is flat (0~m), so CHM directly gives tree height and the resolution is around 0.6 m resolution to be close to the real resolution of our CHM of California \citep{Wagner2024}. As our CHM for California, simulated height is scaled by $\times$2.5 to be stored in integer 8 bits while keeping some sub-meter definition.

Each simulated forest \replaced[id=USGS]{consists of}{consist in} a squared grid of 192$\times$192  pixels with a central 160$\times$ 160 \replaced[id=USGS]{pixel}{pixels} tree placement window. Final CHM and labels are cropped to 128 $\times$ 128 pixels. 
Trees are randomly placed in the 160~\replaced[id=USGS]{pixel}{pixels} window.

To avoid dense clustering of stems and excessive crown overlap in the CHM, we defined a minimum–distance constraint between trees. A candidate tree $i$ with crown radius $R_i$ is accepted only if the Euclidean distance to every previously accepted tree $j$ with radius $R_j$ satisfies

 \begin{eqnarray}
 \| X_i - X_j \| > (R_i + R_j) f_{\min}   
 \end{eqnarray}   

where $X_i = (x_i, y_i)$, $X_j = (x_j, y_j)$ and $f_{\min} = 1/3$ is a constant distance factor. Candidate trees that do not respect this condition are rejected and resampled until a valid position is found or the defined maximum number of attempts is reached. Distance computations are carried out using tensors for improved efficiency.

The number of trees and their functional types is defined by scenarios of forest with different functional compositions and densities, Table \ref{paramscenario}. For example, large-tree forests mix conifers and deciduous with some giant \replaced[id=USGS]{sequoia}{Sequoia} trees. Sequoia scenarios are dominated by giant \replaced[id=USGS]{sequoia}{Sequoia} trees (20--50\%). The number of trees varies by scenario (e.g., 15 to 40 for \replaced[id=USGS]{sequoia}{Sequoia} scenario) to avoid saturation while keeping a dense canopy. Scenarios generated using an automated agent were used to simulate the different forests observed in the Sierra Nevada, producing mixed or homogeneous stands from shrubs to giant \replaced[id=USGS]{sequoia}{Sequoias} forest. 

\begin{table}[ht]
  \centering
  \setlength{\tabcolsep}{4pt}
  \footnotesize
\caption{Example of giant sequoia forest scenarios. For each scenario, the number of trees per simulation $N$ is drawn uniformly from the integer range in the second column. The remaining columns are sampling weights for functional types (row sums to 1). Abbreviations: con., coniferous; t.d., tall deciduous; l.c.d., large-crown deciduous; m.d., medium deciduous; s.c., sapling coniferous; s.d., sapling deciduous; sh.r., shrub rounded; sh.s., shrub spreading; g.d., giant deciduous; g.c., giant coniferous; g.r., giant sequoia (``redwood'' in the scenario names).}
  \begin{tabular}{lcccccccccccc}
    \hline
    Scenario (short name) & $N$ range &    con. & t.d. & l.c.d. & m.d. & s.c. & s.d. & sh.r. & sh.s. & g.d. & g.c. & g.r. \\
    \hline
    \texttt{dense\_redwood}      & 20--30 &
      .05 & .05 & .05 & .05 & .05 & .05 & .05 & .05 & .10 & .10 & .40 \\
    \texttt{redwood\_conif\_mix} & 25--35 &
      .20 & .05 & .05 & .05 & .10 & .05 & .05 & .05 & .05 & .15 & .20 \\
    \texttt{old\_growth\_redwood}& 20--30 &
      .10 & .10 & .10 & .10 & .05 & .05 & .05 & .05 & .10 & .10 & .20 \\
    \texttt{redwood\_giant\_mix} & 25--35 &
      .10 & .10 & .10 & .10 & .05 & .05 & .05 & .05 & .15 & .15 & .15 \\
    \texttt{pure\_redwood}       & 15--25 &
      .05 & .05 & .05 & .05 & .05 & .05 & .05 & .05 & .05 & .05 & .50 \\
    \texttt{redwood\_tall}       & 25--35 &
      .10 & .20 & .10 & .10 & .05 & .05 & .05 & .05 & .10 & .05 & .15 \\
    \texttt{redwood\_conif\_dom} & 25--35 &
      .25 & .05 & .05 & .05 & .10 & .05 & .05 & .05 & .05 & .15 & .20 \\
    \texttt{mixed\_redwood}      & 30--40 &
      .12 & .12 & .12 & .12 & .08 & .08 & .08 & .08 & .10 & .10 & .10 \\
    \hline
  \end{tabular}
  \label{paramscenario}
\end{table}

Trees are defined by functional types with ranges of height and crown radius, Table \ref{paramsim}. This includes conifers, deciduous, shrubs, and giant conifers and giant \replaced[id=USGS]{sequoia}{Sequoia} types. For example, for \replaced[id=USGS]{sequoia}{Sequoia} tree, ranges of height (H) and radius (R) are given by :
\begin{eqnarray}
H \sim \mathcal{U}(80, 250), \quad R \sim \mathcal{U}(13, 27)
\end{eqnarray}
For each simulated tree we sample $H$ and $R$ from its type range.

\begin{table}[ht]
  \centering
  \setlength{\tabcolsep}{3pt}
  \caption{Tree functional types and their height and crown parameters used in the CHM simulation. Height values in the simulator are stored in scaled units; physical height (m) is obtained by dividing by $2.5$ (heights are stored as 8-bit integers scaled by 2.5 to limit data storage, that is, with a 0.4 m step, as in our CHM of California \citep{Wagner2024}). Crown radius is given in pixels in the simulator; crown radius (m), where one pixel corresponds to $0.6$\,m on the ground. The column $H/R$ corresponds to the \texttt{height\_radius\_ratio} field in the same list.}
  \begin{tabular}{lccccc}
    \hline
    Type (code name) &     Height (scaled) &     Height (m) &     Crown radius (px) &
    Crown radius (m) &     $H/R$ \\
    \hline
    coniferous              & 50--100  & 20.0--40.0  & 4--6   & 2.4--3.6   & 12 \\
    tall\_deciduous         & 60--100  & 24.0--40.0  & 3--5   & 1.8--3.0   & 15 \\
    large\_crown\_deciduous & 35--80   & 14.0--32.0  & 6--8   & 3.6--4.8   & 6 \\
    medium\_deciduous       & 40--70   & 16.0--28.0  & 4--6   & 2.4--3.6   & 9 \\
    sapling\_coniferous     & 10--25   & 4.0--10.0   & 1--2   & 0.6--1.2   & 10 \\
    sapling\_deciduous      & 8--20    & 3.2--8.0    & 1--3   & 0.6--1.8   & 6 \\
    shrub\_rounded          & 7--15    & 2.8--6.0    & 2--4   & 1.2--2.4   & 3 \\
    shrub\_spreading        & 7--12    & 2.8--4.8    & 3--5   & 1.8--3.0   & 2 \\
    giant\_deciduous        & 80--120  & 32.0--48.0  & 10--15 & 6.0--9.0   & 7 \\
    giant\_coniferous       & 90--130  & 36.0--52.0  & 8--12  & 4.8--7.2   & 10 \\
    gigantic\_redwood       & 150--250 & 60.0--100.0 & 8--20  & 4.8--12.0  & 9 \\
    \hline
  \end{tabular}
  \label{paramsim}
\end{table}

Then tree attributes are converted into rasterized canopy surfaces using shape templates. 

For conifers and giant trees, narrow forms, conical crowns are used. For a tree with total height $H$ and crown radius $R$, the idealized crown surface as a function of horizontal distance $r$ from the stem is defined as 
\[
z(r) = H \left(1 - \frac{r}{R}\right)^{s},
\]
where $s$ is a cone–shape parameter. Larger values of $s$ produce more pointed crowns, whereas values close to one approximate straight–sided cones.

For broadleaf trees and shrubs, we use rounded crowns. In this case, the canopy surface is defined as
\[
z(r) = H \cdot f\!\left(\frac{r}{R}; q\right),
\]
where $f$ is a smooth decreasing function, such as a cosine–based profile, that is equal to one at $r = 0$ and zero at $r = R$, and $q$ is a roundness parameter that controls whether the crowns are relatively flat or dome–shaped.

To mimic a LiDAR CHM, only the highest point per pixel is kept and the id of the tree that has this highest point updated accordingly in a separate tree ID raster. Final outputs are cropped to 128 $\times$ 128 pixels. We further randomly add a blur to the simulated canopy height as a data augmentation technique, rescale the maximum height to the maximum initial height and remove all height below 1.60 m. 

Finally we saved the simulated canopy height model without data augmentation, with random blur augmentation, a tree-ID map in raster which associates each height in the canopy height model to its original tree and the individual polygons of the trees.

Our generated dataset consists of 59,134 synthetic CHM images of 128 $\times$ 128 pixels, containing a total of 3,356,877 simulated trees.

\begin{figure}[ht]
 \centering
 \includegraphics[width=0.85\linewidth]{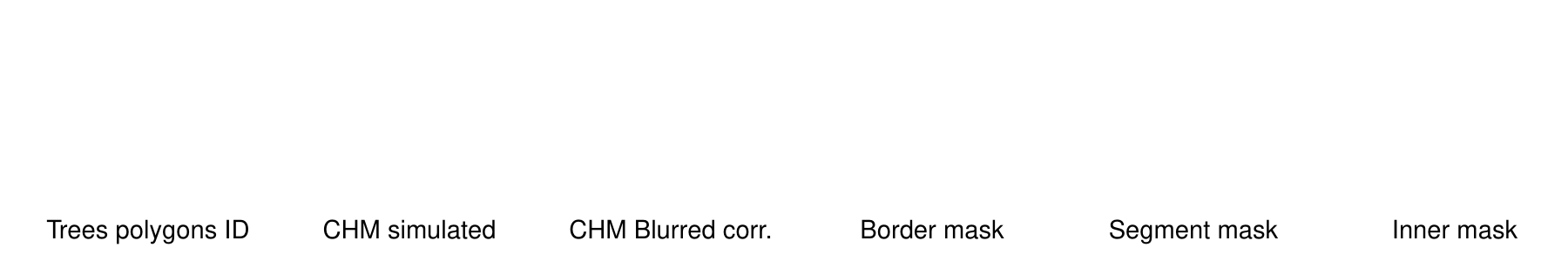}
 \includegraphics[width=0.85\linewidth]{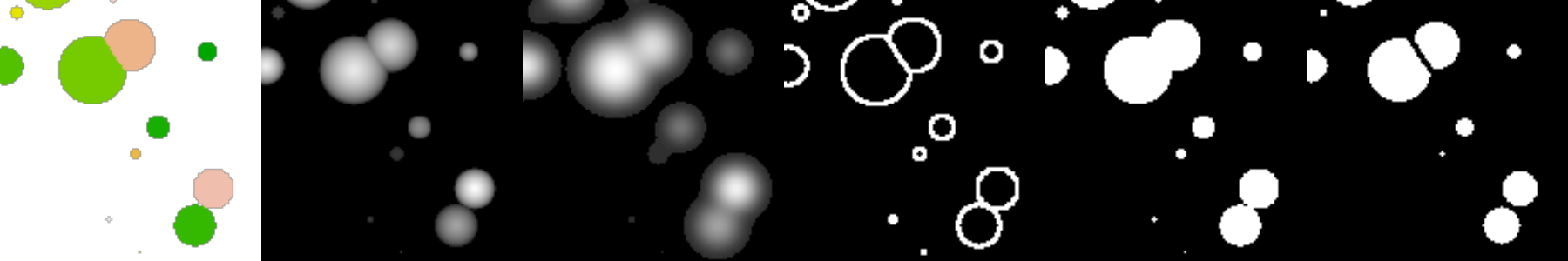}
 \includegraphics[width=0.85\linewidth]{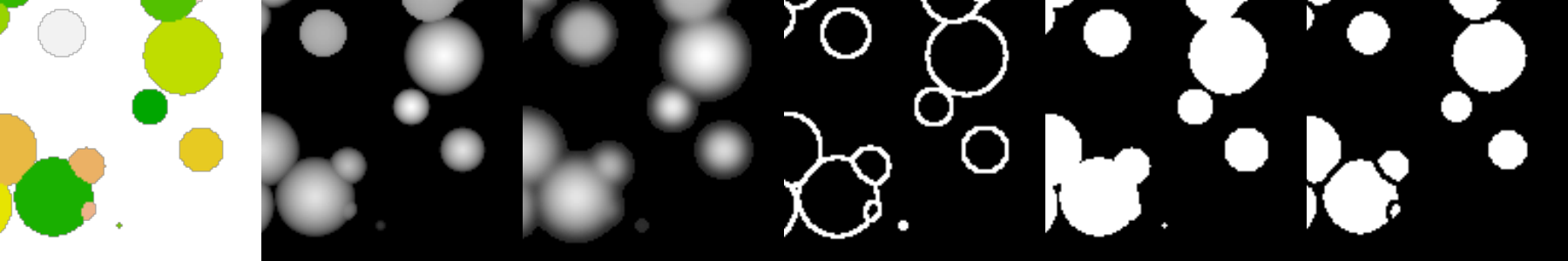}
  \includegraphics[width=0.85\linewidth]{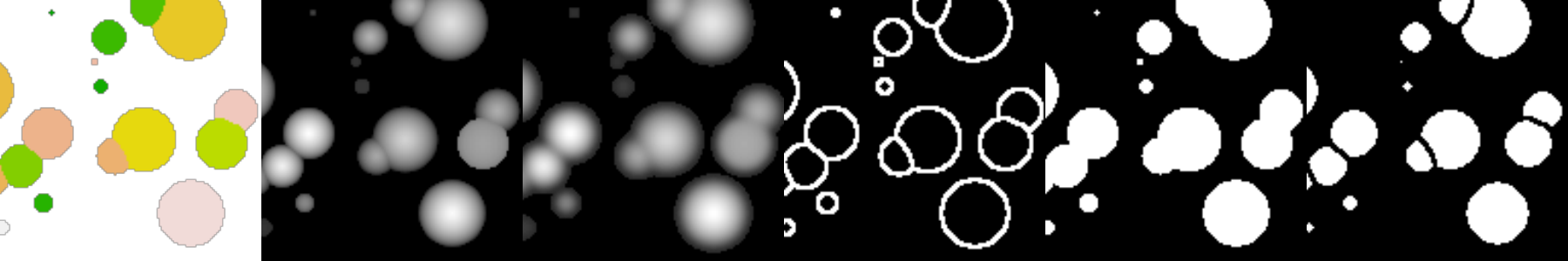}
  \includegraphics[width=0.85\linewidth]{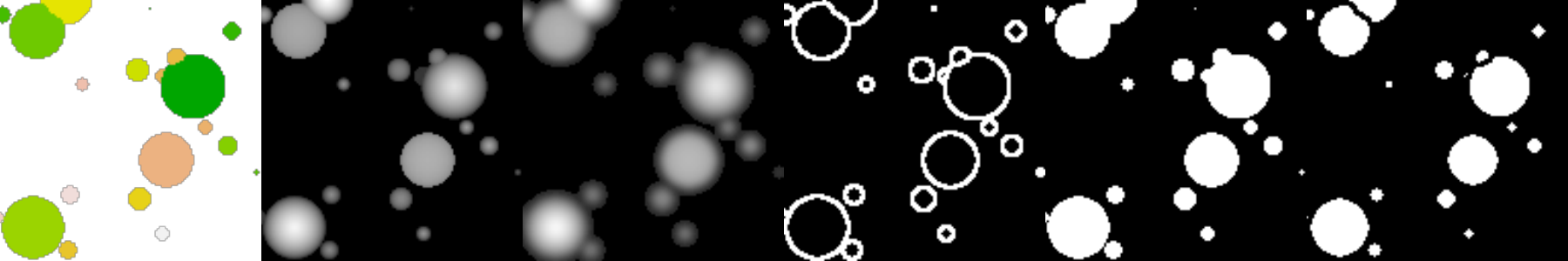}
\caption{Example of simulated individual tree polygons and the associated simulated CHM, simulated blurred CHM and masks in raster format.} 
  \label{Fig3}
  \end{figure}

\clearpage 
\section{Download of the Sentinel-2 images}\label{secA_sits}
\setcounter{table}{0}\setcounter{figure}{0}

We prepared a Sentinel--2 surface reflectance data cube for the Sierra Nevada region using the \texttt{sits} package \citep{Simoes2021}. As a first step, we delineated the region of interest using the California Floristic Province shapefile and selected only the polygons corresponding to the Sierra Nevada subsection. This polygon mask was reprojected to WGS84 and used to intersect the global Sentinel--2 tiling grid, which yielded 22 Sentinel--2 tiles overlapping the Sierra Nevada. The tile identifiers were stored in a list of tile names, including the tile that contains our sequoia field plots (11SLA). For each tile in this list, we used \texttt{sits\_cube} to query the Sentinel--2 L2A collection hosted on Amazon Web Services. We requested the blue (B02), green (B03), red (B04), near–infrared (B08), shortwave infrared (B11 and B12) and cloud mask (CLOUD) bands, and restricted the temporal coverage to the period from December~2017 to February~2025. To reduce the influence of clouds at the scene level, we exploited the \texttt{file\_info} metadata returned by \texttt{sits\_cube} and retained only images whose reported cloud cover was less than or equal to five percent over the tile, discarding more contaminated acquisitions.

The selected images for each tile were then copied from the cloud provider to local storage using \texttt{sits\_cube\_copy}, with multi–core processing activated to accelerate data transfer. We used a simple consistency check to ensure that all requested bands were present for every acquisition date, repeatedly calling \texttt{sits\_cube\_copy} until the local directory contained the same number of images for each band and for the cloud mask. This procedure produced, for each Sentinel--2 tile intersecting the Sierra Nevada, a locally stored, band–complete archive of Sentinel--2 L2A scenes with low overall cloud cover, which was subsequently used as input to the regularization and time–series analysis steps described in the following sections.

\clearpage 
\section{NAIP 2020 and 2024 images of large crowns of each class}\label{secA_naip}
\setcounter{table}{0}\setcounter{figure}{0}

\begin{figure}[ht]
\centering
\includegraphics[width=1.0\linewidth]{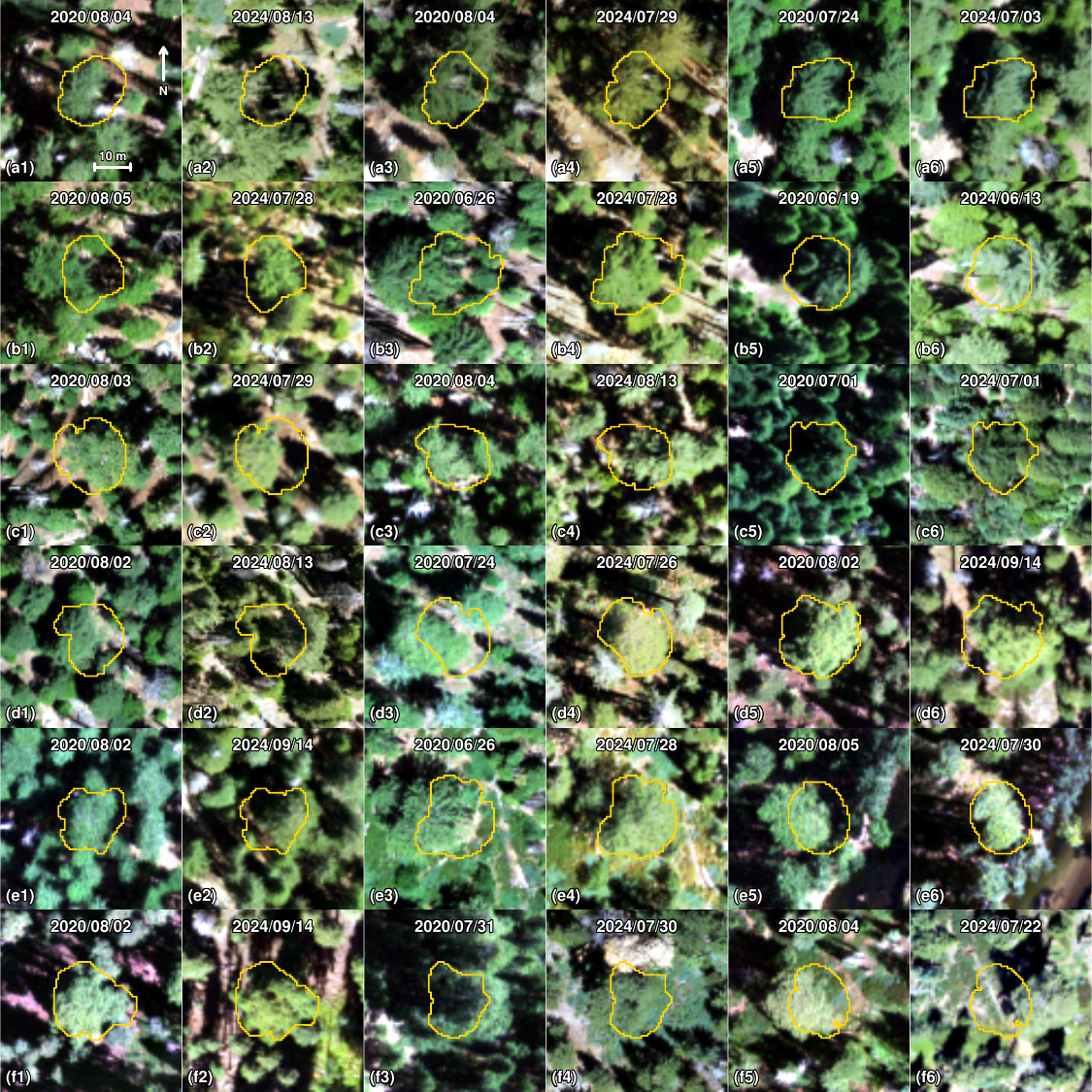}
\caption{NAIP images of 18 large crowns of the class alive (1), before (2020, left image of each pair) and after the fires (2024, right image of each pair). Each image covers 50 $\times$ 50 m around the crown (polygon in gold) and was stretched to improve visualisation; the acquisition date is given at the top.}
\label{fig_naip_class1}
\end{figure}
\newpage

\begin{figure}[ht]
\centering
\includegraphics[width=1.0\linewidth]{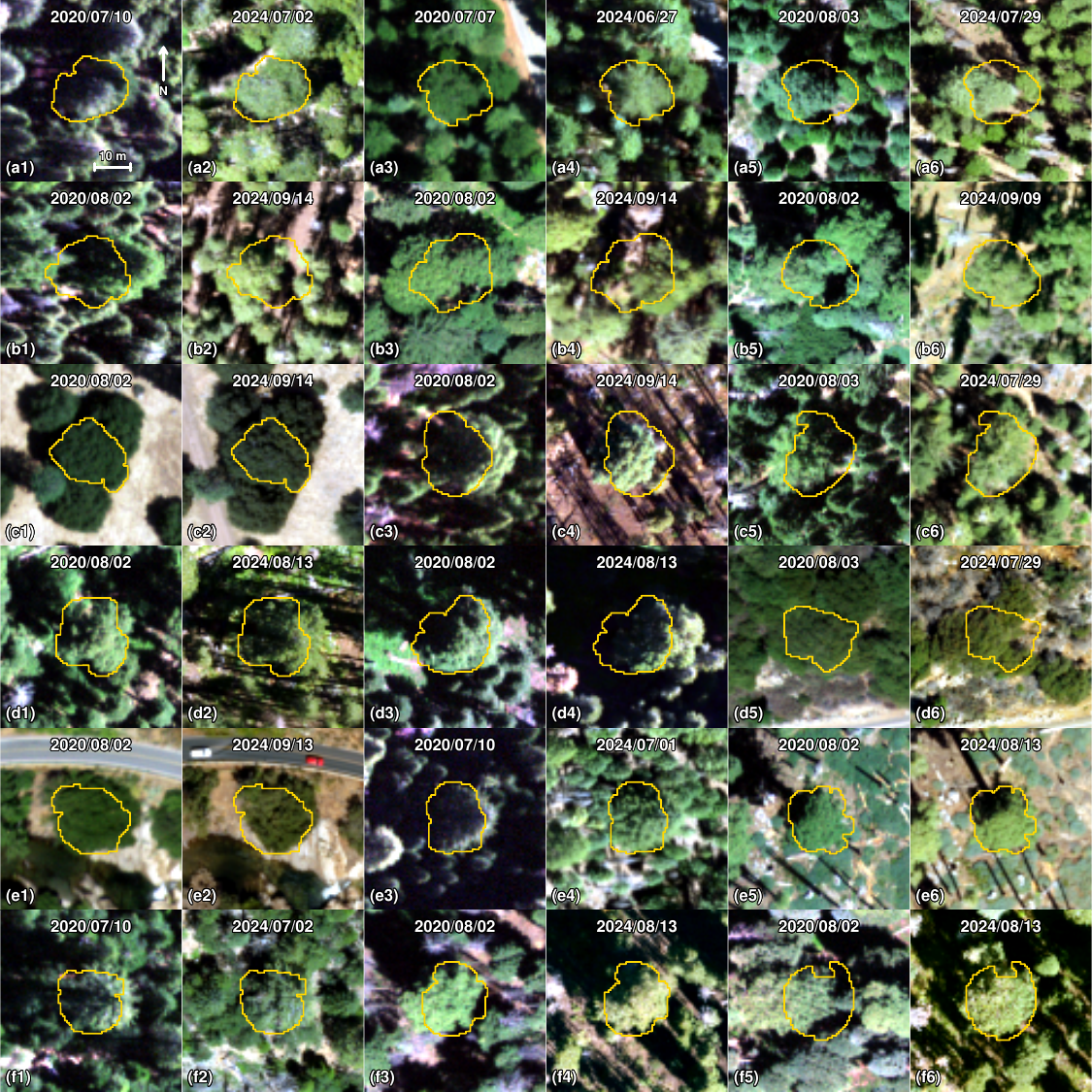}
\caption{NAIP images of 18 large crowns of the class alive with breakpoint (2), before (2020, left image of each pair) and after the fires (2024, right image of each pair). Each image covers 50 $\times$ 50 m around the crown (polygon in gold) and was stretched to improve visualisation; the acquisition date is given at the top.}
\label{fig_naip_class2}
\end{figure}
\newpage

\begin{figure}[ht]
\centering
\includegraphics[width=1.0\linewidth]{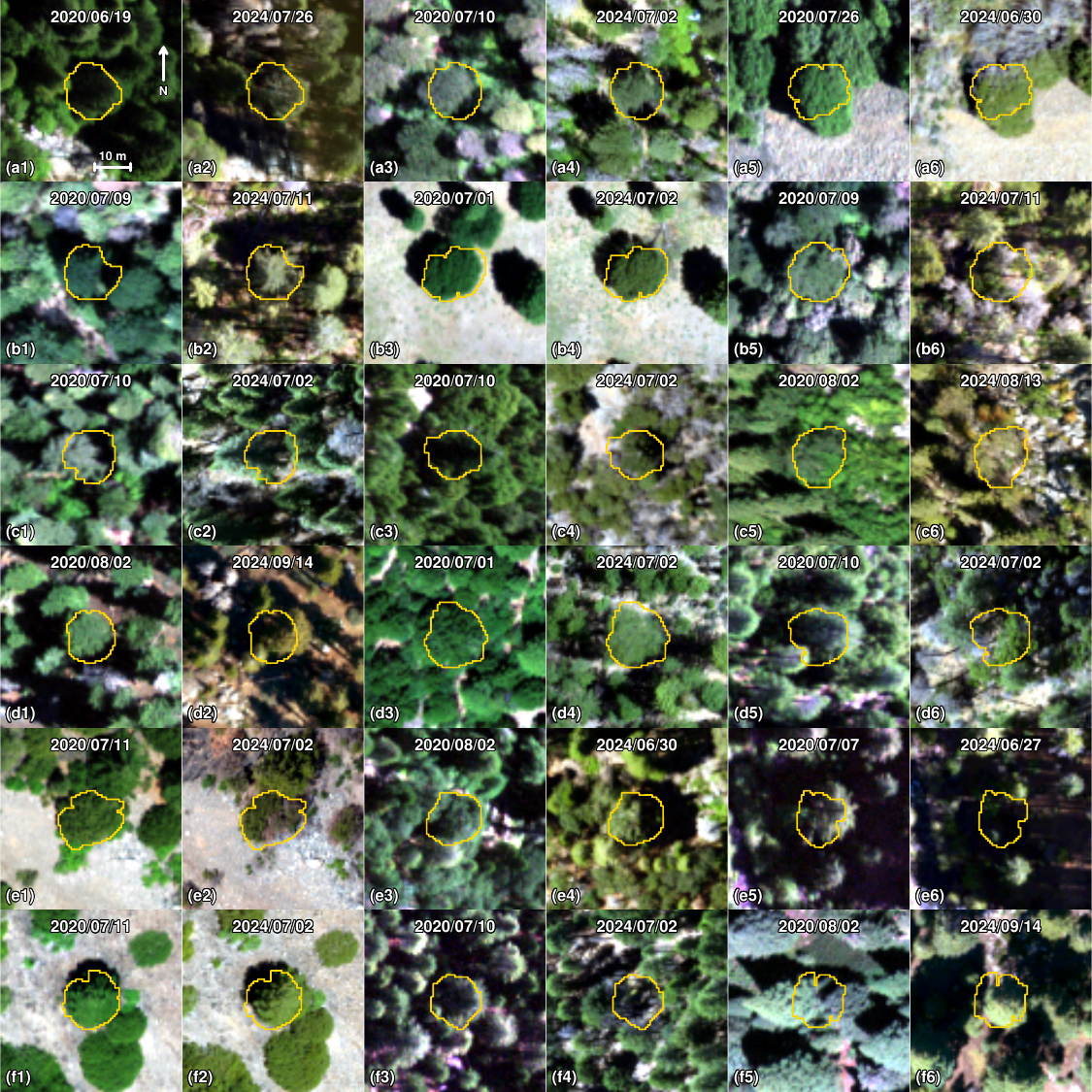}
\caption{NAIP images of 18 large crowns of the class breakpoint with $\geq$80\% NDVI recovery under a closed canopy (3a), before (2020, left image of each pair) and after the fires (2024, right image of each pair). Each image covers 50 $\times$ 50 m around the crown (polygon in gold) and was stretched to improve visualisation; the acquisition date is given at the top. The crown is still present in 2024.}
\label{fig_naip_class3a}
\end{figure}
\newpage

\begin{figure}[ht]
\centering
\includegraphics[width=1.0\linewidth]{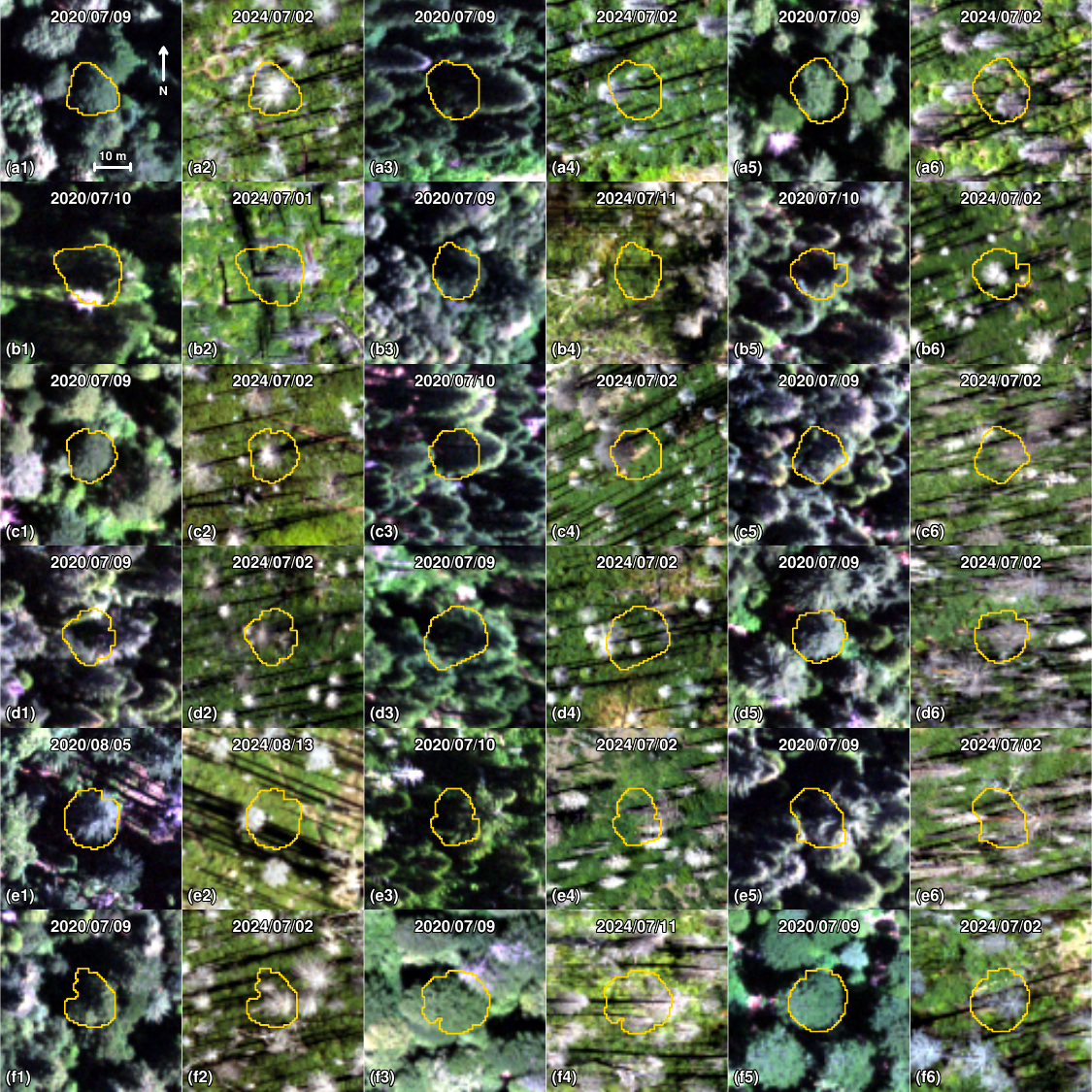}
\caption{NAIP images of 18 large crowns of the class breakpoint with $\geq$80\% NDVI recovery and canopy opening (3b), before (2020, left image of each pair) and after the fires (2024, right image of each pair). Each image covers 50 $\times$ 50 m around the crown (polygon in gold) and was stretched to improve visualisation; the acquisition date is given at the top. The crown is lost in 2024 and the green signal comes from the understory between the snags.}
\label{fig_naip_class3b}
\end{figure}
\newpage

\begin{figure}[ht]
\centering
\includegraphics[width=1.0\linewidth]{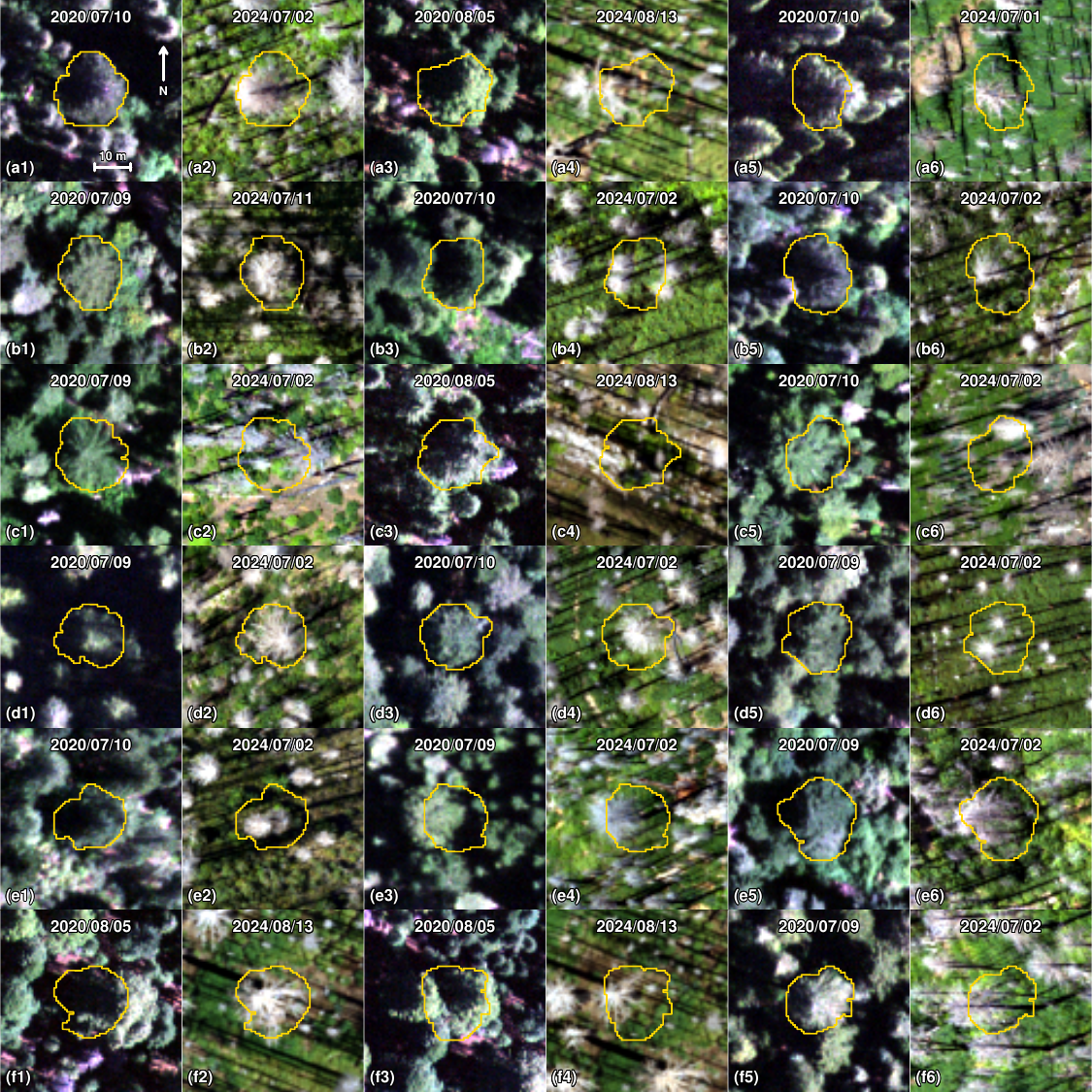}
\caption{NAIP images of 18 large crowns of the class breakpoint with $<$80\% NDVI recovery (4), before (2020, left image of each pair) and after the fires (2024, right image of each pair). Each image covers 50 $\times$ 50 m around the crown (polygon in gold) and was stretched to improve visualisation; the acquisition date is given at the top.}
\label{fig_naip_class4}
\end{figure}
\newpage

\begin{figure}[ht]
\centering
\includegraphics[width=1.0\linewidth]{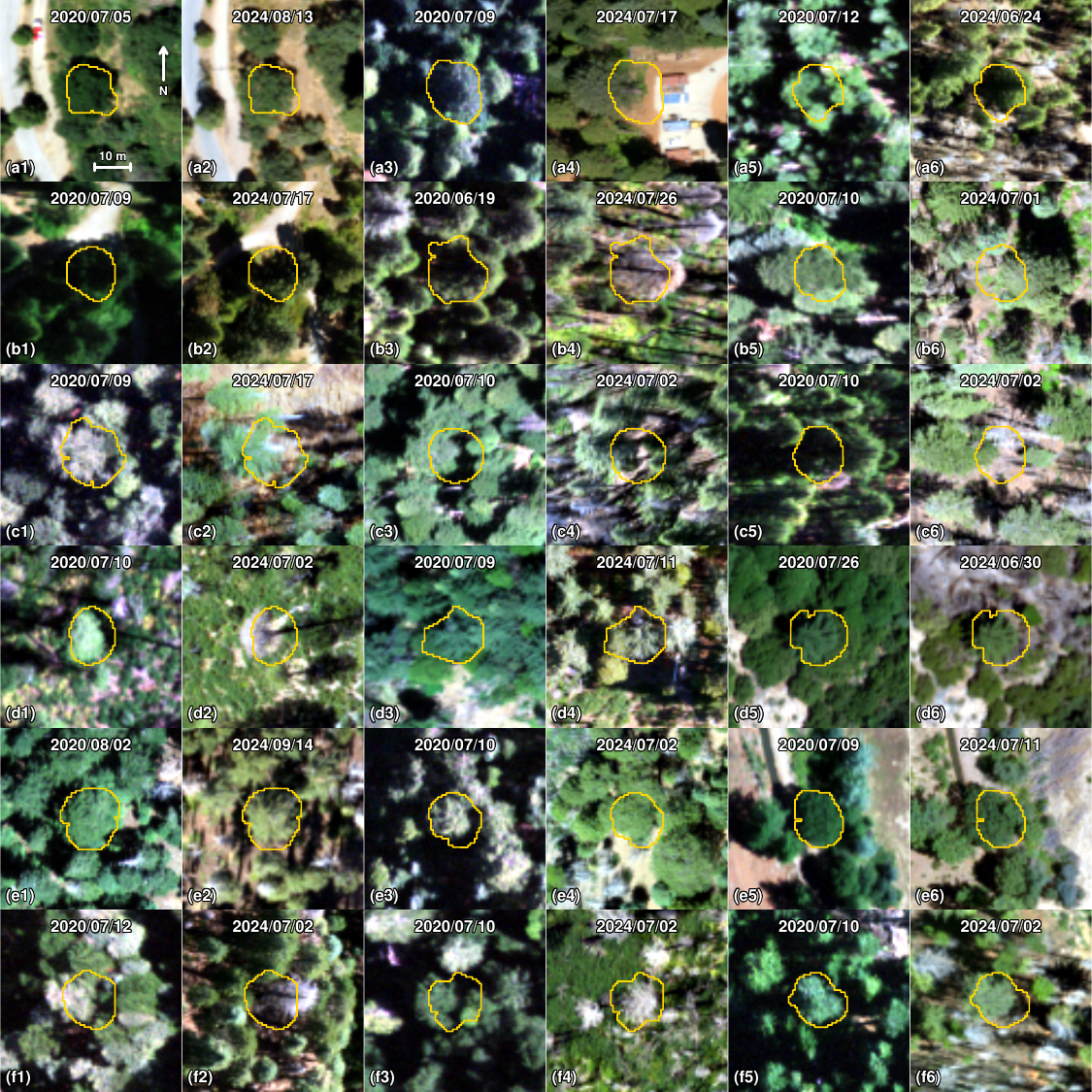}
\caption{NAIP images of 18 large crowns of the class permanent alteration (5), before (2020, left image of each pair) and after the fires (2024, right image of each pair). Each image covers 50 $\times$ 50 m around the crown (polygon in gold) and was stretched to improve visualisation; the acquisition date is given at the top. This class is not visually distinct: some crowns look unchanged and others lost.}
\label{fig_naip_class5}
\end{figure}
\newpage

\begin{figure}[ht]
\centering
\includegraphics[width=1.0\linewidth]{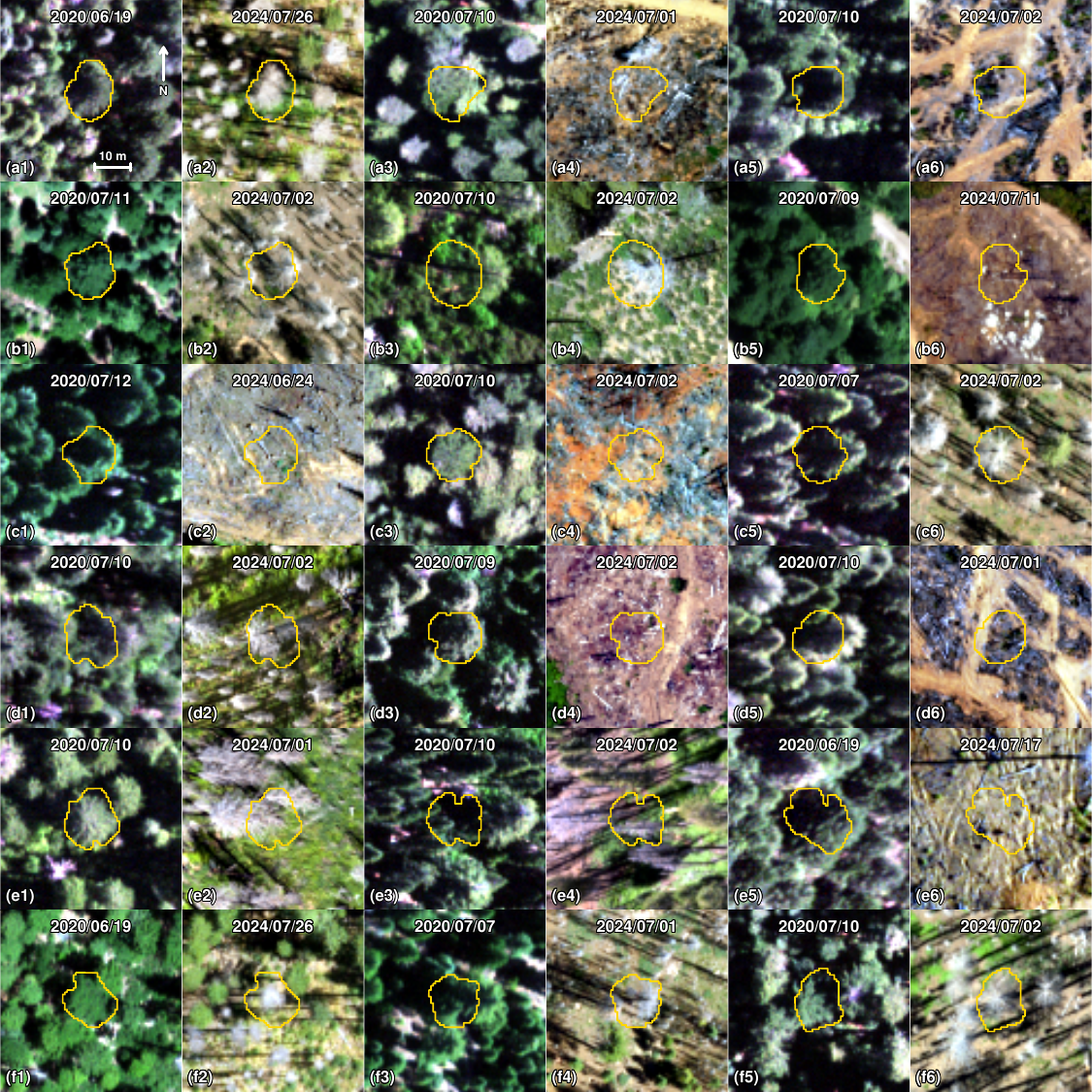}
\caption{NAIP images of 18 large crowns of the class dead (6), before (2020, left image of each pair) and after the fires (2024, right image of each pair). Each image covers 50 $\times$ 50 m around the crown (polygon in gold) and was stretched to improve visualisation; the acquisition date is given at the top.}
\label{fig_naip_class6}
\end{figure}
\newpage

\begin{figure}[ht]
\centering
\includegraphics[width=1.0\linewidth]{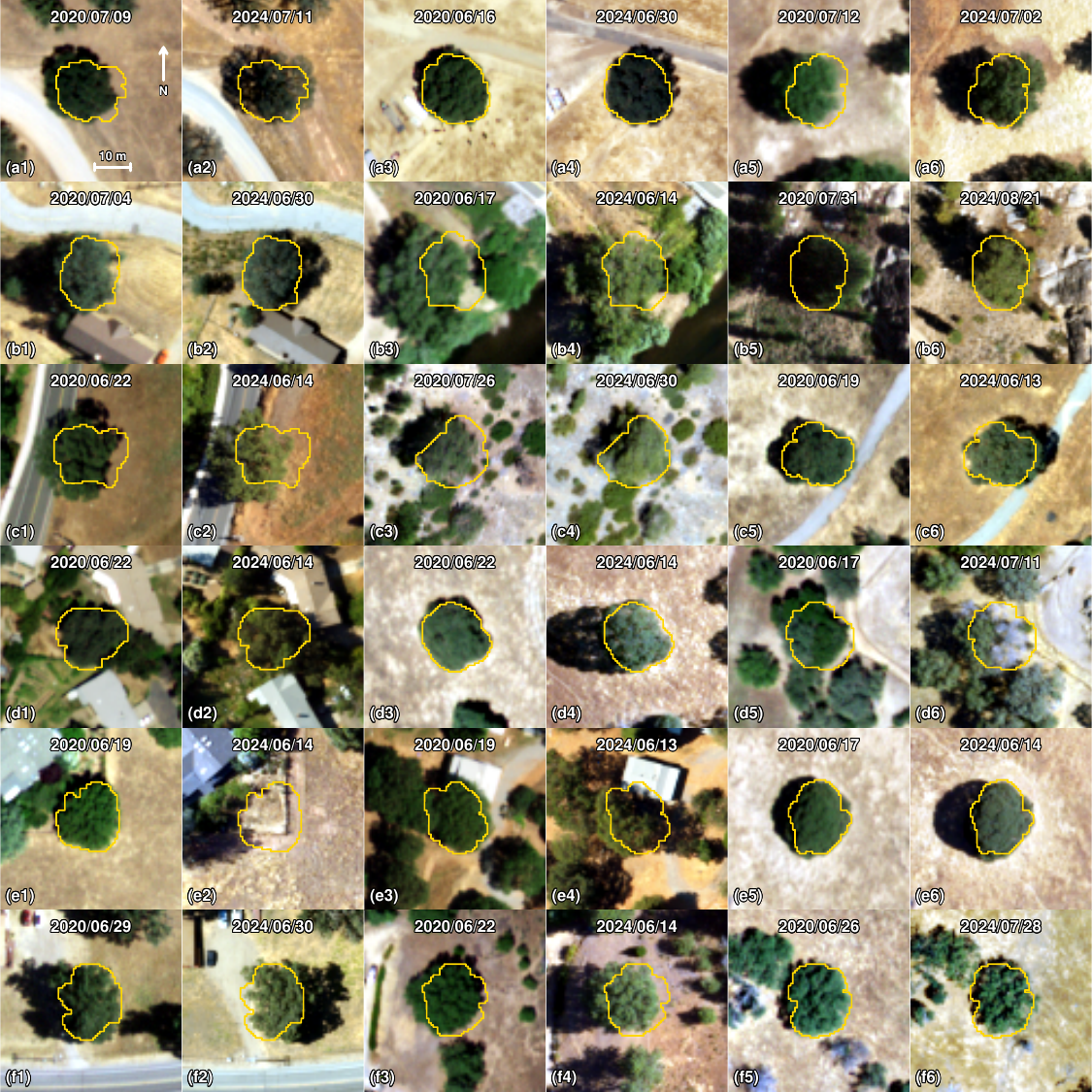}
\caption{NAIP images of 18 large crowns of the class already dead or outside the reference colour range (0), before (2020, left image of each pair) and after the fires (2024, right image of each pair). Each image covers 50 $\times$ 50 m around the crown (polygon in gold) and was stretched to improve visualisation; the acquisition date is given at the top. Most crowns of this class are isolated trees on dry grassland.}
\label{fig_naip_class0}
\end{figure}
\end{appendices}

\end{document}